\documentclass{article}
\usepackage{colortbl}

\PassOptionsToPackage{numbers,sort&compress}{natbib}
    \usepackage[preprint]{neurips_2025}

\usepackage[utf8]{inputenc} % allow utf-8 input
\usepackage[T1]{fontenc}    % use 8-bit T1 fonts
\usepackage[pagebackref=true,breaklinks=true,letterpaper=true,colorlinks,bookmarks=false]{hyperref}       % hyperlinks
\usepackage{url}            % simple URL typesetting

\usepackage{booktabs}       % professional-quality tables
\usepackage{amsfonts}       % blackboard math symbols
\usepackage{xspace}         % for \xspace command
\usepackage{caption} % for \captionof outside floats
\usepackage{nicefrac}       % compact symbols for 1/2, etc.
\usepackage{microtype}      % microtypography
\usepackage{xcolor}         % colors

\usepackage{times}
\usepackage{epsfig}
\usepackage{graphicx}
\usepackage{amsmath}
\usepackage{amssymb}

\usepackage[export]{adjustbox}

\usepackage{booktabs}

\usepackage{float}

\usepackage{stfloats} % added by author2

\usepackage{dsfont}
\usepackage{subcaption}
\usepackage{caption}
\usepackage{nicefrac}
\usepackage{mathrsfs}
\usepackage{xcolor}
\usepackage{enumitem}
\usepackage{tcolorbox}
\usepackage{colortbl}
\usepackage{multirow}
\usepackage{comment}
\usepackage{wrapfig}
\usepackage{marvosym}
\usepackage[numbers,sort&compress]{natbib} 
\usepackage[capitalize]{cleveref}
\crefname{section}{Sec.}{Secs.}
\Crefname{section}{Section}{Sections}
\Crefname{table}{Table}{Tables}
\crefname{table}{Tab.}{Tabs.}

\definecolor{maroon}{cmyk}{0,0.87,0.68,0.32}
\definecolor{myyellow}{RGB}{218, 160, 109}
\definecolor{brickred}{rgb}{0.8, 0.25, 0.33}
\definecolor{brandeisblue}{rgb}{0.0, 0.44, 1.0}
\definecolor{applegreen}{rgb}{0.55, 0.71, 0.0}
\definecolor{aogreen}{rgb}{0.0, 0.5, 0.0}

\definecolor{gdmb}{RGB}{47, 114, 173}  % "alpha = 0.25" with white, else too aggressive
\definecolor{gdmr}{RGB}{199, 100,  38}
\definecolor{gdmg}{RGB}{70, 155, 118}
\definecolor{gdmm}{RGB}{193, 126, 165}
\definecolor{gdmy}{RGB}{239, 227,  98}
\definecolor{gdmc}{RGB}{110, 179, 228}
\definecolor{gdmk}{RGB}{20, 20, 20}
\definecolor{turquoise}{cmyk}{0.65,0,0.1,0.3}
\definecolor{purple}{rgb}{0.65,0,0.65}
\definecolor{dark_green}{rgb}{0, 0.5, 0}
\definecolor{orange}{rgb}{0.8, 0.6, 0.2}
\definecolor{red}{rgb}{0.8, 0.2, 0.2}
\definecolor{darkred}{rgb}{0.6, 0.1, 0.05}
\definecolor{blueish}{rgb}{0.0, 0.3, .6}
\definecolor{light_gray}{rgb}{0.7, 0.7, .7}
\definecolor{pink}{rgb}{1, 0, 1}
\definecolor{greyblue}{rgb}{0.25, 0.25, 1}
\definecolor{orgred}{rgb}{1.0, 0, 0}

\usepackage{pifont}

\definecolor{sh_gray}{rgb}{0.84,0.84,0.84}
\definecolor{sh_gray2}{rgb}{1,0.89,0.75}
\definecolor{color3}{rgb}{0.95,0.95,0.95}
\definecolor{color4}{rgb}{0.94,0.94,1}
\definecolor{color5}{rgb}{1,0.96,0.88}
\definecolor{LightGray}{gray}{0.9}

\newif\ifdraft
\drafttrue
\ifdraft
\newcommand{\jwc}[1]{{\color{red}[\textbf{jw} #1]}}

\newcommand{\todo}[1]{{\color{blue}[TODO: #1]}}

\else
\newcommand{\jwc}[1]{}
\newcommand{\smc}[1]{}
\newcommand{\todo}[1]{}

\fi
\newcommand{\gcell}[1]{\cellcolor{LightGray!70}#1}
\newcommand\cellzero{\texttt{CELL0}\xspace}
\newcommand\cellone{\texttt{CELL1}\xspace}
\newcommand\celltwo{\texttt{CELL2}\xspace}

\newcommand\xtwo{\texttt{X2}\xspace}
\newcommand\xfour{\texttt{X4}\xspace}
\newcommand\xeight{\texttt{X8}\xspace}

\def\namecolor{\textcolor{brickred}{4}%
\textcolor[RGB]{255,140,0}{K}%
\textcolor[RGB]{40,116,210}{A}%
\textcolor[RGB]{55,126,215}{g}%
\textcolor[RGB]{70,136,220}{e}%
\textcolor[RGB]{85,146,225}{n}%
\textcolor[RGB]{100,156,230}{t}\xspace}
\def\namenocolor{4KAgent\xspace}
\def\namenocolorbf{\textbf{4KAgent}\xspace}
\def\ie{\textit{i.e.}}
\def\eg{\textit{e.g.}}

\def\etc{\textit{etc.}}

\usepackage[object=vectorian]{pgfornament}
\newcommand{\mysectionline}[3]{%
    \nointerlineskip \vspace{.5\baselineskip}\hspace{\fill}{
        \resizebox{0.5\linewidth}{#3}{
            \pgfornament[color = #1]{#2}
        }
    }
    \hspace{\fill}
    \par\nointerlineskip \vspace{.5\baselineskip}
}

\newcommand{\agentlogo}{\raisebox{-0.5em}{\includegraphics[height=2.0em]{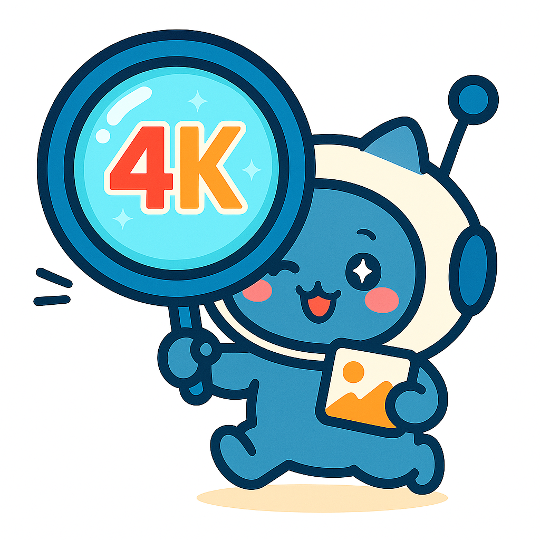}}}
\usepackage{hyperref}
\hypersetup{
colorlinks = true,
breaklinks=true,
colorlinks=true,
citecolor=blueish,
linkcolor=blueish,
}

\definecolor{PastelRed}{HTML}{F8C8C8}
\definecolor{PastelRedFrame}{HTML}{D96A6A}

\definecolor{PastelOrange}{HTML}{F8E0C8}
\definecolor{PastelOrangeFrame}{HTML}{D99A6A}

\definecolor{PastelYellow}{HTML}{F8F0C8}
\definecolor{PastelYellowFrame}{HTML}{D9C96A}

\definecolor{PastelGreen}{HTML}{D8F8C8}
\definecolor{PastelGreenFrame}{HTML}{6AD96A}

\definecolor{PastelTeal}{HTML}{C8F8E8}
\definecolor{PastelTealFrame}{HTML}{6AD9A3}

\definecolor{PastelBlue}{HTML}{C8E0F8}
\definecolor{PastelBlueFrame}{HTML}{6A8DD9}

\definecolor{PastelPurple}{HTML}{E8C8F8}
\definecolor{PastelPurpleFrame}{HTML}{9A6AD9}

\definecolor{PastelPink}{HTML}{F8C8E8}
\definecolor{PastelPinkFrame}{HTML}{D96AB9}

\definecolor{PastelBrown}{HTML}{F0E0D8}
\definecolor{PastelBrownFrame}{HTML}{B7966D}

\newtcolorbox{taskblock}[2][]{%
  center title,
  fonttitle=\bfseries\footnotesize,
  fontupper=\scriptsize,
  boxrule=0.8pt,
  arc=3pt,
  left=1pt,right=1pt,top=1pt,bottom=1pt,
  title=#2,
  halign=center,
  before skip=4pt,   
  after skip=4pt,    
  colback=blue!5,
  colframe=blue!50!black,
  #1
}

\title{
\agentlogo\ %
\namecolor:\\ Agentic Any Image to 4K Super-Resolution}

\author{%
Yushen Zuo$^1$, Qi Zheng$^{1\dagger}$, Mingyang Wu$^{1\dagger}$, Xinrui Jiang$^{2\dagger}$, Renjie Li$^1$,\\ 
\bf Jian Wang$^3$, Yide Zhang$^4$, Gengchen Mai$^5$, Lihong V. Wang$^6$, James Zou$^2$,\\
\bf Xiaoyu Wang$^7$, Ming-Hsuan Yang$^8$, Zhengzhong Tu$^{1\star}$
\\[2pt]
$^1$Texas A\&M University\quad $^2$Stanford University\quad $^3$Snap Inc.\quad $^4$CU Boulder\\
$^5$UT Austin\quad
$^6$California Institute of Technology
\quad$^7$Topaz Labs\quad $^8$UC Merced
\\[2pt]
\small $^\star$Corresponding Author: \texttt{tzz@tamu.edu}. $^{\dagger}$Equal contributions.\\[2pt]
\textbf{\textcolor{magenta}{Project Website}}: \href{https://4kagent.github.io}{\color{black}{\texttt{4kagent.github.io}}}
}

\begin{document}

\maketitle
\vspace{-8mm}

\noindent\makebox[\textwidth][c]{%
    \begin{minipage}{1.1\textwidth}
        \includegraphics[width=\textwidth]{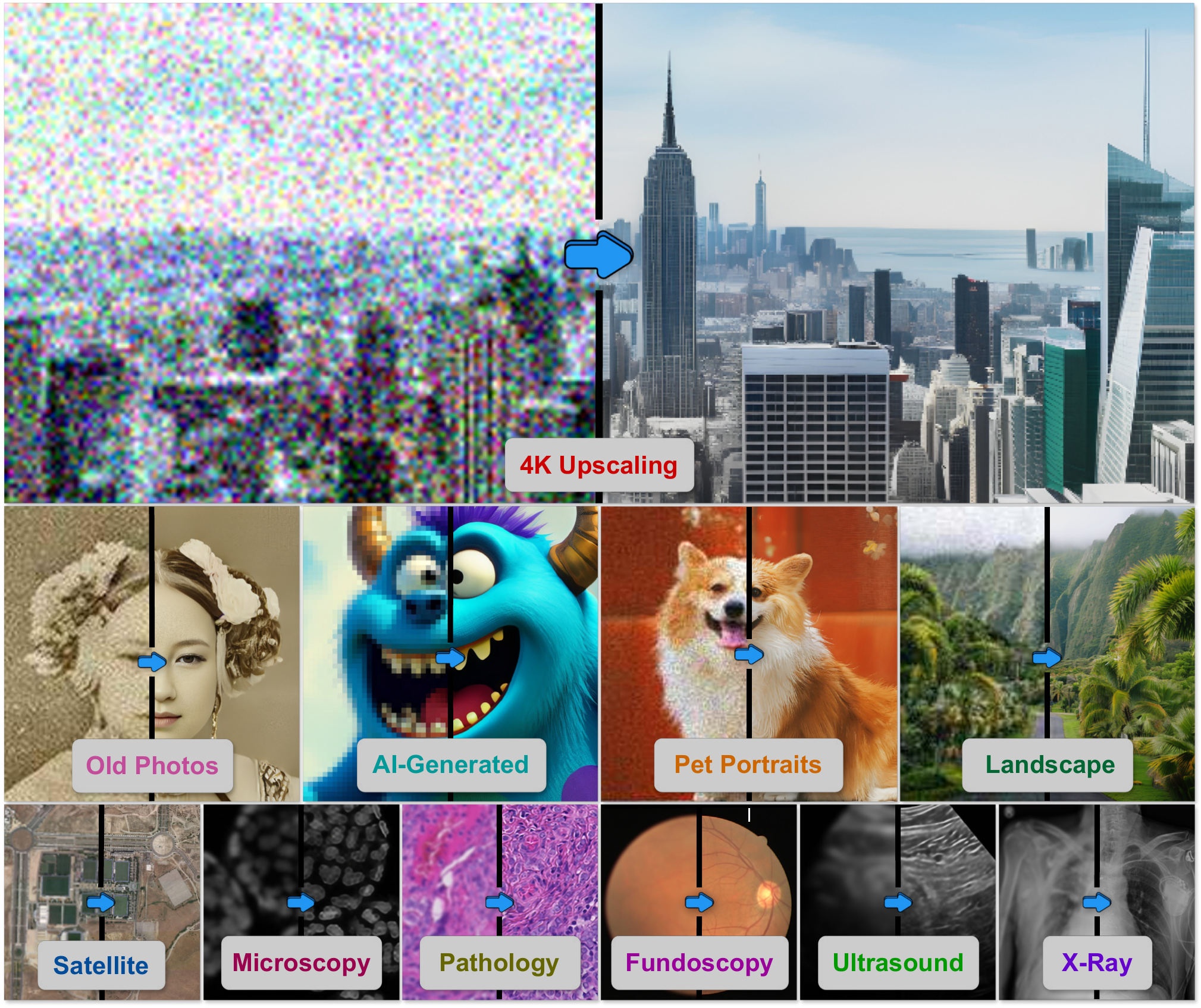}
        \vspace{-5mm}
        \captionof{figure}{We present
        \textbf{\namecolor}, an agentic image super-resolution generalist designed to universally upscale \textbf{\underline{any} image to 4K}, regardless of input \emph{type}, \emph{degradation level}, or \emph{domain}. 
        That is, \textbf{4KAgent} effectively restores diverse imagery, spanning from natural scenes, severely degraded captures (\eg, old photos), human/pet portraits, AI-generated content (AIGC), as well as \emph{specialized scientific imaging domains}, such as remote sensing, fluorescence microscopy, pathology, and various medical modalities like X-ray, ultrasound, and funduscopy---all \textbf{without the need} for any re-training or domain-specific adaptation.}
        \label{fig:teaser}
    \end{minipage}
}

\begin{abstract}
We present \textbf{\namenocolor}, a unified agentic super-resolution generalist system designed to universally upscale any image to 4K resolution (and even higher, if applied iteratively). 
Our system can transform images from extremely low resolutions with severe degradations, for example, highly distorted inputs at $256\times 256$, into crystal-clear, photorealistic 4K outputs.
4KAgent comprises three core components: \textbf{(1)} \textit{Profiling}, a module that customizes the 4KAgent pipeline based on bespoke use cases; \textbf{(2)} A \textit{Perception Agent}, which leverages vision-language models alongside image quality assessment experts to analyze the input image and make a tailored restoration plan; and \textbf{(3)} A \textit{Restoration Agent}, which executes the plan, following a recursive execution-reflection paradigm, guided by a quality-driven mixture-of-expert policy to select the optimal output for each step.
Additionally, 4KAgent embeds a specialized face restoration pipeline, significantly enhancing facial details in portrait and selfie photos.
We rigorously evaluate our 4KAgent across \textbf{11} distinct task categories encompassing a total of \textbf{26} diverse benchmarks, setting new state-of-the-art on a broad spectrum of imaging domains.
Our evaluations cover natural images, portrait photos, AI-generated content, satellite imagery, fluorescence microscopy, and medical imaging like fundoscopy, ultrasound, and X-ray, demonstrating superior performance in terms of both perceptual (\eg, NIQE, MUSIQ) and fidelity (\eg, PSNR) metrics. 
By establishing a novel agentic paradigm for low-level vision tasks, we aim to catalyze broader interest and innovation within vision-centric autonomous agents across diverse research communities.
We will release all the code, models, and results at: \url{https://4kagent.github.io}.
\end{abstract}

\mysectionline{gdmk}{88}{1ex}

{
  \hypersetup{linkcolor=black}
  \tableofcontents
}
\clearpage

\section{Introduction}
Image super-resolution (SR) is a fundamental task in computer vision that aims to reconstruct high-resolution (HR) images from their low-resolution (LR) counterparts~\cite{yang2010image,dong2014learning,dong2015image,ledig2017photo,wang2020deep,zhang2018residual,wang2018esrgan,wang2021real,saharia2022image,wang2024exploiting}. 
It serves as a bedrock for various low-level vision tasks~\cite{zhang2017beyond,liang2021swinir,zamir2022restormer,tu2022maxim}, including deblurring~\cite{tao2018scale,cho2021rethinking}, dehazing~\cite{he2010single,li2018benchmarking}, deraining~\cite{ren2019progressive,jiang2020multi}, and low-light enhancement~\cite{wei2018deep,guo2016lime}.
Beyond its classical role in computational photography and imaging, SR techniques significantly influence numerous domains, such as biomedical imaging~\cite{greenspan2009super,schermelleh2019super}, remote sensing~\cite{shermeyer2019effects,he2021spatial,kowaleczko2023real}, surveillance~\cite{zhang2010super}, and embodied artificial intelligence applications~\cite{haris2021task,shan2020simulation,islam2020underwater}.

Traditional SR methods~\cite{dong2014learning,wang2018esrgan} typically assume known synthetic degradation during training, which limits their generalization to real-world captures that suffer from complex, heterogeneous, and unpredictable degradations~\cite{wang2021real}.
Recent research has increasingly shifted to a more practical real-world super-resolution (RealSR) task~\cite{chen2022real,wu2024one}, which attempts to explicitly address diverse and unknown degradations in naturally captured photo- and video-graphs. 
RealSR requires models not only to handle multiple combined degradations effectively but also to exhibit strong adaptability and generalization across varied scenarios~\cite{ying2020patches,tu2021ugc}.
Many effective solutions have been proposed to solve the RealSR problem, via simulating complex real-world degradations~\cite{zhang2021designing,wang2021real}, leveraging the powerful generative prior of pre-trained diffusion models~\cite{jiang2024autodir,wu2024one,wu2024seesr,sun2024pixel,mei2025power}, enabling robust restoration under unknown conditions. 
Inspired by the advanced planning and reasoning capabilities of large language models (LLMs)~\cite{wei2022chain, huang2024understanding,yao2023react,durante2024agent}, agentic restoration frameworks~\cite{chen2024restoreagent,zhu2024intelligent} have emerged as a powerful tool that can adaptively handle multiple degradations through sequential planning and dynamic restoration strategies.

Despite their successes in certain scenarios, existing performant generative approaches~\cite{wu2024one,sun2024pixel} can only handle limited degradation ranges, \eg, up to 4$\times$ upscaling, failing to recover extremely low-quality images with highly complex and diverse degradations in the wild.
Moreover, SR specialist models are known to generalize poorly to out-of-distribution domains~\cite{chen2024low}, let alone when applied on a different scaling factor. 
This is mainly due to heavy reliance on supervised learning on synthetic image pairs that cannot fully simulate the complex real-world image degradations, not to mention other domains, ranging from AI-generated imagery, scientific computing, to biomedical images.
Last but not least, practically, users often demand highly specific workflows, \eg, only denoising, 4K upscaling, or prioritizing high fidelity over perceptual quality, and a one-size-fits-all system that can flexibly adapt to satisfy diverse requirements and application scenarios is in pressing need.

To fill this gap, we present \textbf{\namecolor}, the first-of-its-kind agentic framework for generic, flexible, and interpretable super-resolution of \textbf{any image to 4K}.
As illustrated in~\cref{fig:teaser}, 4KAgent is capable of upscaling any low resolution image (\eg, 0.065 megapixels) to 4K$\times$4K, (\ie, 16.7 megapixels) by 16$\times$ upscaling factors\footnote{\namenocolor\ can actually achieve arbitrarily large-scale super-resolution if applied recursively~\cite{saharia2022image}.} (\S\ref{ssec:16x-sr}).
It also sets new state-of-the-art (SoTA) on classical image super-resolution (\S\ref{ssec:classical-sr}), real-world image super-resolution (\S\ref{ssec:real-sr}), face restoration (\S\ref{ssec:face-sr}), and multiple-degradation image restoration (\S\ref{ssec:multiple-sr}) benchmarks, in terms of perceptual quality.
We also show that \namenocolor enjoys broader applications in broad low-level tasks, such as joint restoration \& 4K upscaling (\S\ref{ssec:div4k-sr}), and AI-generated content 4K upscaling (\S\ref{ssec:aigc-sr}).
Lastly, thanks to the mixture-of-experts and profile design, \namenocolor demonstrates larger impact on interdisciplinary areas such as scientific super-resolution (\S\ref{sec:experiment-ii}), including \ding{182} Satellite image super-resolution (\S\ref{ssec:sci-sr-remote-sensing}), \ding{183} fluorescence microscopy super-resolution (\S\ref{ssec:sci-sr-ii-micro}), and \ding{184} medical image super-resolution ((\S\ref{ssec:sci-sr-iii-patho},\ref{ssec:sci-sr-iiii-med})).

\textbf{Our contributions are as follows}:
\begin{itemize}[leftmargin=1em,nosep]
    \item \textbf{[\textcolor{brickred}{Framework}]} We present \namenocolorbf, the first AI agent framework for universal any-image-to-4K upscaling, capable of handling \textbf{all image categories}, ranging from classical and realistic degradations, extreme low-quality inputs, AI-generated imagery, to scientific imaging tasks such as remote sensing, microscopy, and biomedical inputs.
        \item \textbf{[\textcolor{gdmb}{System Design}]}
    We design a multi-agent system in 4KAgent, the \textbf{Perception Agent} employs large vision-language models (VLMs) to analyze the content and distortion in the image and provide the restoration plan for the restoration agent to execute. The \textbf{Restoration Agent}, 
    which sets up an execution---reflection---rollback procedure for recursive restoration and upscaling.
    \item \textbf{[\textcolor{gdmg}{Q-MoE \& Face Restoration pipeline}]}
    In each restoration step of the restoration plan, we propose a Quality-Driven Mixture-of-Expert (\textbf{Q-MoE}) policy in execution and reflection to select the optimal image. We further develop a face restoration pipeline to enhance faces in images.
    \item \textbf{[\textcolor{orange}{Profile Module}]}
    To expand the applicability of 4KAgent, we propose a \textbf{Profile Module} to bring the availability to customize the system for different restoration tasks. 4KAgent can adapt to different restoration tasks without extra training.
        \item \textbf{[\textcolor{purple}{DIV4K-50 Dataset}]}
        To evaluate 4K super-resolution performances, we build the \textbf{DIV4K-50} dataset as a challenging testset to upscale a low-quality (LQ) image in $256\times 256$ resolution with multiple degradations to a high-quality (HQ) 4K image in $ 4096\times 4096$ resolution.
    \item \textbf{[\textcolor{violet}{Experiments}]}
    Extensive experimental results demonstrate the superiority of 4KAgent as a \textbf{generalist 4K upscaling agent}: 4KAgent sets new state-of-the-art on a variety of real-world image super-resolution benchmarks, multiple-degradation restoration benchmarks, face restoration, 4K upscaling task, and various scientific imaging tasks, including satellite image super-resolution, fluorescence microscopic imaging, X-ray radiography, and bio-medical imaging super-resolution. 

\end{itemize}

\section{Method}
\label{method}

\subsection{System Overview}
\begin{figure*}[t]
\centering
\includegraphics[width=\textwidth]{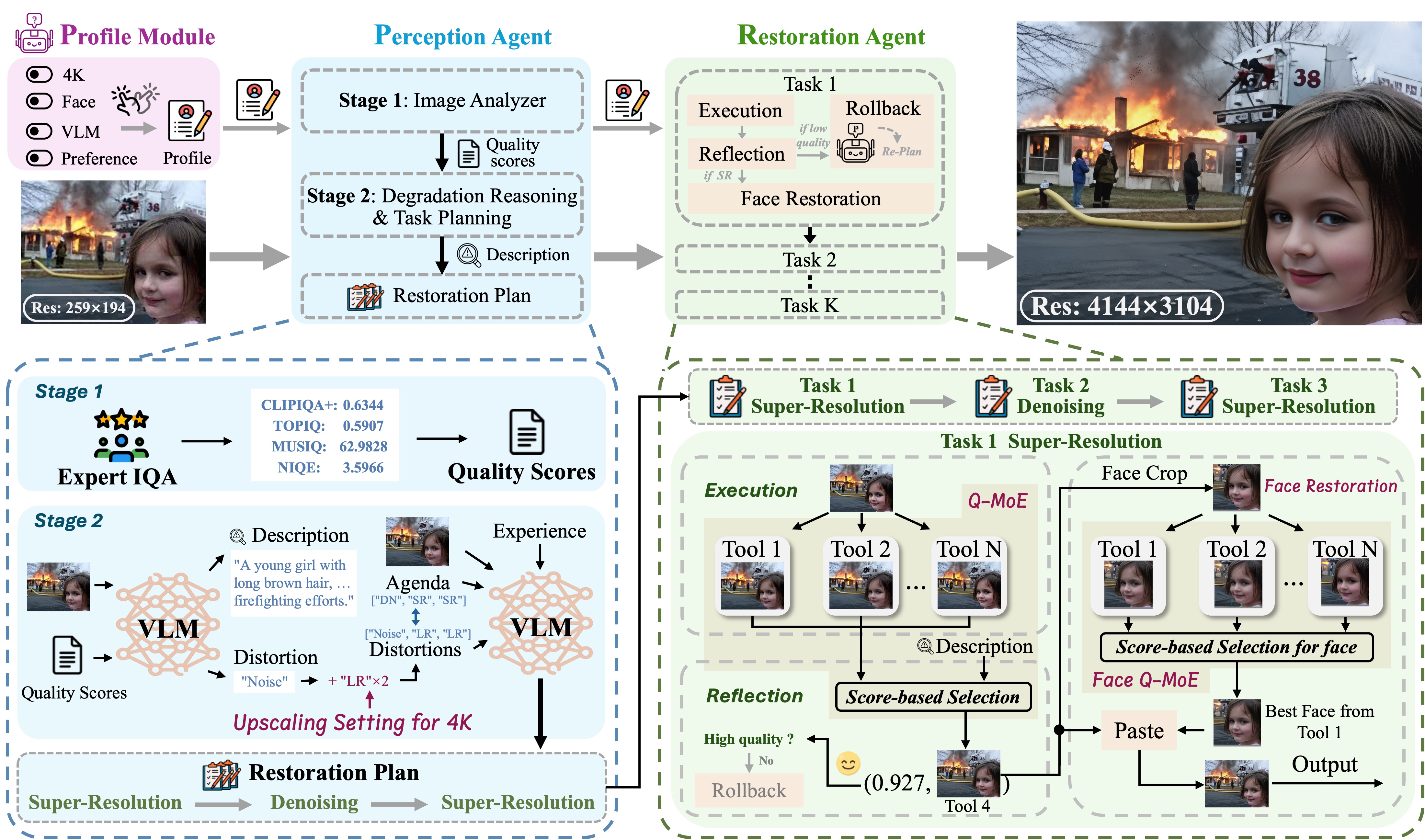}
\caption{\textbf{4KAgent system overview.}} 
\vspace{-1em}
\label{fig:agent_overview}
\end{figure*}

We introduce \textbf{4KAgent}, a multi-agent framework designed to upscale any real-world image to 4K resolution. \cref{fig:agent_overview} illustrates the overall workflow of our proposed 4KAgent, which decomposes the restoration pipeline into a collection of specialized agents. The \textbf{Perception Agent} analyzes degradations (noise, blur, \etc), extracts semantic/structural cues, and schedules a restoration plan containing a sequence of operators (denoising, deblurring, super-resolution, \etc). The \textbf{Restoration Agent} follows the restoration plan using our proposed Quality-driven Mixture-of-Experts (Q-MoE) to pick the best output from multiple restoration tools. The Rollback mechanism will be activated if the quality of the restored image falls below a threshold. Additionally, a dedicated \textbf{Face Restoration Pipeline} further enhances facial regions by triggering expert face restoration models. A user-configurable \textbf{Profile Module} allows users to customize the system (\eg, prioritize fidelity or perceptual quality), enabling robust, high-quality 4K SR across diverse content and degradation types.

\subsection{Perception Agent}
\label{ssec:perception_agent}
The \textbf{Perception Agent} is designed as a four-stage analytical module that bridges low-level image quality assessment with high-level reasoning. Its core function is to extract a robust, holistic understanding of the input image in terms of both semantic content and low-level degradations, and to create a restoration plan that will guide the subsequent restoration process. 

\textbf{Image Analyzer.} Perception agent invokes a suite of expert Image Quality Assessment (IQA) tools that evaluate the input image $I$ across multiple quality dimensions $Q_{I} = (Q_{1}, Q_{2}, ...)$. Specifically, we adopt CLIPIQA \cite{wang2023exploring}, TOPIQ \cite{chen2024topiq}, MUSIQ \cite{ke2021musiq}, and NIQE \cite{zhang2015feature} as the IQA metrics. 
These metrics represent perceptual quality from diverse aspects (due to their different model designs and training data), which will be employed as \emph{context} for the next step of degradation reasoning.

\textbf{Degradation Reasoning.} Perception agent leverages a VLM $M_{R}$ to reason over the obtained IQA metrics. Specifically, by incroporating the input image $I$, IQA metrics $Q_{I}$, the VLM $M_{R}$ will predict the degradation list $D_{I}$ from the input image, which will correspond to an initial restoration agenda $A'_{I}$. Meanwhile, $M_{D}$ also analyzes the content in the image and outputs the corresponding image descriptions $C_{I}$ (\ie, captioning). The whole process can be express as $C_{I}, D_{I}, A'_{I} = M_{R}(I, Q_{I})$.

\textbf{Upscaling Factor Configuration.}
4KAgent is able to automatically determine and apply an appropriate super-resolution scale to reach 4K resolution. Given an input image $I$ with height $H_{I}$ and width $W_{I}$, the scale factor $s$ is calculated by
\begin{equation}
\label{equ:scale_factor}
    s = \min\Bigl(\{\,s \in \{2,4,8,16\} \mid max(H_{I}, W_{I})*s \ge 4000\}\;\cup\;\{16\}\Bigr),
\end{equation}
After obtaining the initialized agenda $ A'_{I}$ from $M_{D}$, 4KAgent will calculate the scale factor $s$ and append super-resolution task(s) (\eg, when $s=16$, append two `super-resolution (4$\times$)' task) into $ A'_{I}$ to obtain the final agenda $A_{I}$ and update $D_{I}$ correspondingly.  
Under this setting, 4KAgent is able to upscale any image (resolution larger than $250\times 250$) to 4K resolution in a single process.

\textbf{Task Planning.}
After obtaining the degradation list $D_{I}$ present in the input image and the restoration agenda $A_{I}$, the perception agent employs an LLM / VLM $M_{P}$ to provide the restoration plan. Specifically, by coporating image descriptions $C_{I}$, degradation list $D_{I}$, restoration experience $E$, and input image itself $I$ (available when using VLM as $M_{P}$), $M_{P}$ outputs an initial restoration plan $P_{I} = M_{P}(C_{I}, D_{I}, A_{I}, E, I)$, which contains a sequence of restoration tasks.

\subsection{Restoration Agent}
Building upon the task plan $P_{I}$ provided by the Perception Agent, the Restoration Agent executes an iterative process, each stage of which tightly couples restoration and evaluation using an execution-reflection-rollback triplet. 
Within this agent, we propose a quality-driven mixture-of-experts (\textbf{Q-MoE}) policy, both in execution and reflection, to select the optimal image for each restoration step. 
We also employ a rollback mechanism to adjust the restoration plan if necessary.

\textbf{Execution.} 
Guided by the task plan $P_{I}$, this agent executes the restoration step by step. In each restoration step, the input image will go through all tools in the toolbox, which contains a number of advanced restoration models (detailed in~\cref{ssec:model-zoo}) corresponding to each individual restoration task. 
In 4KAgent, we have curated \textbf{9} different restoration tasks that are useful to enhance picture quality: \textbf{Brightening}, \textbf{Defocus Deblurring}, \textbf{Motion Deblurring}, \textbf{Dehazing}, \textbf{Denoising}, \textbf{Deraining}, \textbf{JPEG Compression Artifact Removal},  \textbf{Super Resolution}, and \textbf{Face Restoration}. 
Specifically, for step $k$ in the restoration plan, it produces multiple restoration results $\{T_{i}(I_{k-1}), i=1\sim N\}$ ($T_{i}$ is $i$-th tool in the toolbox, $N$ is the number of tools in the toolbox) based on the input image $I_{k-1}$.

\textbf{Reflection.} 
After obtaining restoration results $\{T_{i}(I_{k-1}), i=1\sim N\}$, the Restoration Agent will select the optimal image based on their quality. To evaluate the quality of image $T_{i}(I_{k-1})$, we compute the image quality score by combining the preference model HPSv2 \cite{wu2023human} and no-reference IQA metrics. Specifically, we use HPSv2 to assess the human preference of the resulting image $T_{i}(I_{k-1})$ based on the image content description $C_{I}$. For no-reference IQA metrics, we employ NIQE \cite{mittal2012making}, MANIQA \cite{yang2022maniqa}, MUSIQ \cite{ke2021musiq}, and CLIPIQA \cite{wang2023exploring} to calculate a weighted sum as its no-reference quality score:
\begin{equation}
 \label{equ:reflection}
    Q_{s}(T_{i}(I_{k-1})) = H(T_{i}(I_{k-1}, C_{I})) + Q_{nr}(T_{i}(I_{k-1})) / 4,
\end{equation}
where $H$ denotes HPSv2, 
\begin{equation}
    Q_{nr}(T_{i}(I_{k-1})) =  w_\text{NIQE}*(1 - Q_\text{NIQE}/10) + \sum_{j\in\Omega} w_{j}*Q_{j}.
\end{equation}
$\Omega = \{\text{MUSIQ}, \text{MANIQA}, \text{CLIPIQA}\}$. After obtaining the quality score of each result image, the final result of this restoration step is obtained by the highest quality score:
\begin{equation}
    I_{k} = \mathop{\arg\max}\limits_{Q_{s}}(T_{1}(I_{k-1}), T_{2}(I_{k-1}), ..., T_{N}(I_{k-1})).
\end{equation}
   The combination of execution and reflection can be viewed as a Mixture-of-Expert (MoE) system, which we refer to as a quality-driven mixture-of-expert (\textbf{Q-MoE}): the input image is processed through each expert (execution), and the Reflection function selects the optimal image among all.

\textbf{Rollback.}
Following previous AI Agent systems \cite{patil2024goex,zhu2024intelligent,zhang2025enhancing,li2025generator,hu2025webcot}, we also design a rollback mechanism in the 4KAgent system. Specifically, if the quality score of $I_{k}$ after step $k$ in the initial restoration plan $P_{I}$ is lower than a threshold $\eta$, \ie, $Q_{s}(I_{k}) \leq \eta$, the restoration step will be seen as a failure step and 4KAgent will generate a failure message $S_{I}$. Then the system will employ the Perception Agent to adjust the subsequent plan based on the degradation list $D_{I}$, the remaining restoration tasks $A_{I}^{R}$ of the restoration agenda $A_{I}$, restoration experience $E$, and failure message $F_{I}$: $P_{I}^{adj} = M_{P}(D_{I}, A_{I}^{R}, E, S_{I})$. After that, the system will assign a different restoration task in this step. If all subsequent restoration tasks assigned to this step lead to rollback, then 4KAgent will take a compromise policy and go back to the original plan to execute subsequent restoration tasks.

\subsection{Face Restoration Pipeline}

Human face regions are often the most visually sensitive and semantically important components in an image. However, conventional super-resolution methods struggle to maintain identity consistency, natural skin textures, and perceptual quality when restoring faces, especially in heavily degraded portraits. To address this, 4KAgent incorporates a dedicated \textbf{Face Restoration pipeline}, which is selectively triggered within the restoration workflow. 
The Face Restoration pipeline is embedded as a submodule in 4KAgent and will only be invoked \emph{after the super-resolution restoration step},
ensuring that face quality refinement is seamlessly integrated into the iterative restoration loop.

\begin{wrapfigure}{R}{0.6\textwidth}
\vspace{-11pt}
\begin{center}
\includegraphics[width=0.6\textwidth]{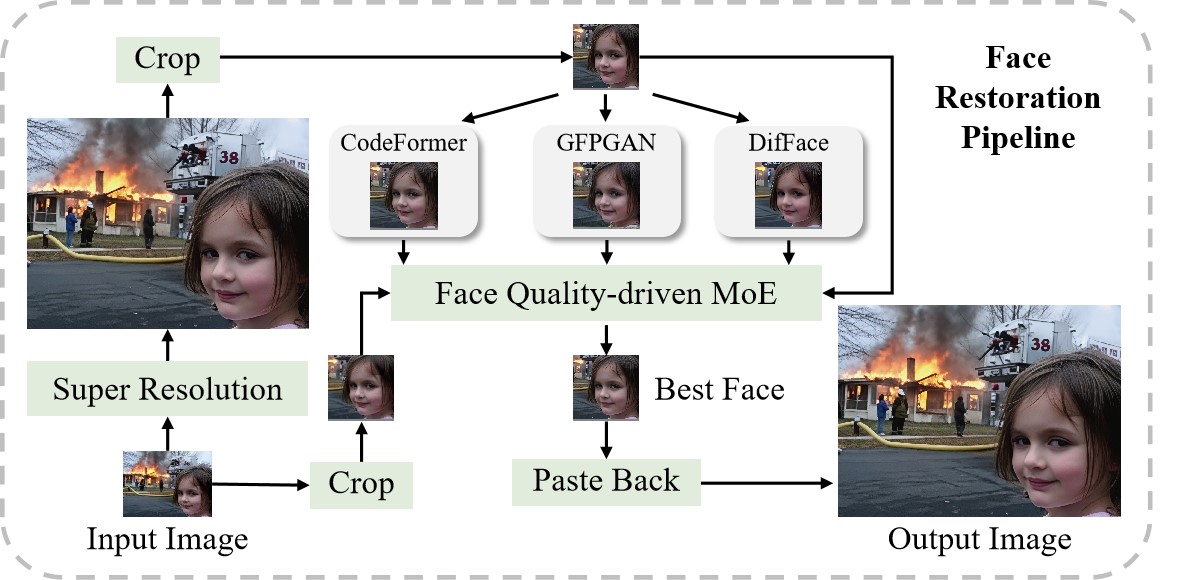}
\end{center}
\vspace{-9pt}
\caption{Face restoration pipeline overview.}
\label{fig:face_restore}
\vspace{-9pt}
\end{wrapfigure}

The overall framework of the face restore pipeline in the 4KAgent system is shown in~\cref{fig:face_restore}.
First, 4KAgent will detect and crop faces in the input image $\{F_{I}^{l}, l=1\sim L\}$ ($L$ is the number of faces in the image $I$). Then, if \textbf{super-resolution} is in the restoration plan and the resulting image $I_{k}$ of super-resolution step does not trigger the rollback mechanism, 4KAgent will detect and crop faces in the resulting image $\{F_{I_{k}}^{l}, l=1\sim L'\}$ ($L'$ is the number of faces in the image $I_{k}$). If $L = L'$, then for each face in $I_{k}$, different advanced face restoration methods are applied, yielding restored faces $\{T_{i}^{f}(F_{I_{k}}^{l}), i=0\sim N^{f}\}$. Here, $T_{i}^{f}$ is a face restoration tool in the toolbox, $T_{0}^{f}$ is an identical function, and $N^{f}$ is the number of face restoration tools in the toolbox.

Likewise, we also conduct Q-MoE policy here: 4KAgent selects the best face based on the quality score $Q_{s}^{f}$. The quality score $Q_{s}^{f}$ not only considers the face quality, but also the identity preservation:
\begin{equation}
    Q_{s}^{f}(T_{i}^{f}(F_{I_{k}}^{l})) = w_{\text{IP}}*\text{IP}(T_{i}^{f}(F_{I_{k}}^{l}), F_{I}^{l}) + w_{\text{IQA}}*(Q_{nr}(T_{i}^{f}(F_{I_{k}}^{l})) / 4 + Q_{\text{CF}}(T_{i}^{f}(F_{I_{k}}^{l}))),
\end{equation}
where $l=1\sim L$. $\text{IP}$ calculates the cosine similarity of face features, extracted using ArcFace \cite{deng2019arcface}. $\text{CF}$ indicates CLIB-FIQA \cite{ou2024clib}, which is an advanced face IQA metric. 4KAgent combines the no-reference quality score used in the reflection stage and the CLIB-FIQA score to assess the face quality. After obtaining quality score $Q_{s}^{f}$, 4KAgent select the best face $F_{out}^{l}$:
\begin{equation}
  F_{out}^{l} = \mathop{\arg\max}_{Q_{s}^{f}}(T_{0}^{f}(F_{I_{k}}^{l}), T_{1}^{f}(F_{I_{k}}^{l}), ..., T_{N_{f}}^{f}(F_{I_{k}}^{l})).
\end{equation}
4KAgent will paste $F_{out}^{l}$ back to the original image $I_{k}$, then proceeding to the next step.

\subsection{Profile Module}
\label{ssec:profile}
To enhance the flexibility and applicability of our 4KAgent system, we develop the \textbf{Profile Module}, enabling dynamic customization for diverse image restoration scenarios, per user's needs. 
Specifically, the Profile Module acts like a system prompt for LLM applications, allowing fine-grained control through the following seven configuration parameters: 
\begin{enumerate}[leftmargin=16pt,nosep,topsep=0pt]
\vspace{-2pt}
\item \textbf{Perception Agent}: Specifies the choice of LLM / VLM employed by the Perception Agent. [\texttt{Default}: \texttt{Llama-vision}] 
\item \textbf{Upscale to 4K}: Determines whether to upscale to 4K resolution. [\texttt{Default}: \texttt{True}] 
\item \textbf{Scale Factor}: Explicitly defines the upscale factor for the entire pipeline. (\texttt{Default}: \texttt{4}, \texttt{Options}: [\texttt{2,4,8,16}]). This parameter overrides ``\textbf{Upscale to 4K}'' when specified.
\item \textbf{Restore Option}: Explicitly sets the restoration task(s) to be applied. If set to \texttt{None}, restoration task(s) are determined automatically by the Perception Agent. (\texttt{Default}: \texttt{None}) 
\item \textbf{Face Restore}: Toggles activation of the dedicated face restore pipeline. (\texttt{Default}: \texttt{True})
\item \textbf{Brightening}: Controls the activation of image brightening, which may cause color shifts in restored images. Provided as [Optional] to maintain image color fidelity.  (\texttt{Default}: \texttt{False})
\item \textbf{Restore Preference}: Defines whether to prioritize higher perceptual quality or higher fidelity in image restoration. (\texttt{Options}: [\texttt{Perception}, \texttt{Fidelity}], \texttt{Default}: \texttt{Perception}). Here we respect the perception-distortion tradeoff~\cite{blau2018perception,zheng2024video}, deeming models that optimize for distortion metrics (\eg, PSNR, SSIM~\cite{wang2004image}) as \texttt{Fidelity} models while methods trained for perceptual quality (\eg, NIQE~\cite{mittal2012making}, MUSIQ~\cite{ke2021musiq}) as \texttt{Perception} models.
\end{enumerate}

The Profile Module offers exceptional configurability, enabling seamless adaptation to a wide range of restoration tasks without requiring model retraining or domain-specific fine-tuning. To the best of our knowledge, 4KAgent is a first-of-its-kind framework that enjoys unprecedented robustness and generalizability: each distinct restoration scenario can be addressed by simply selecting an appropriate configuration profile, thanks to which 4KAgent consistently achieves excellent performance across a variety of challenging restoration domains—including natural images, portraits, AI-generated images (AIGC), medical imaging, scientific microscopy, and remote sensing images—all without additional training or adaptation. Comprehensive details on predefined profiles in 4KAgent and their naming conventions are further elaborated in~\cref{sec:model_card}.

\section{Model Card}
\label{sec:model_card}
\subsection{Profile}
4KAgent is highly flexible based on the profile setting. Users can easily customize 4KAgent by pre-selecting a profile defined in 4KAgent. We pre-define profile examples in~\cref{table:4kagent_profiles}, which cover most use cases and include all the modes used in our experiments. This feature offers easy and intuitive customization for new, unseen scenarios identified by the customers.
\vspace{-3pt}
\begin{table}[ht]
    \centering
        \vspace{-3pt}
    \fontsize{6.5pt}{9.5pt}\selectfont
    \setlength{\tabcolsep}{2.5pt}
    \caption{Pre-defined Profiles in 4KAgent.}
    \label{table:4kagent_profiles}
   \resizebox{\textwidth}{!}{%
    \begin{tabular}{l l l l l l l l}
        \toprule
        \textbf{Profile Nickname} & \textbf{Perception Agent} & \textbf{Upscale to 4K} & \textbf{Scale Factor} & \textbf{Restore Option} & \textbf{Face Restore} & \textbf{Brightening} & \textbf{Restore Preference}  \\
        \midrule
        Gen4K-P & DepictQA \cite{you2024depicting} & True & None & None & True & False & Perception \\
        \midrule
        Gen4K-F & DepictQA \cite{you2024depicting} & True & None & None & True & False & Fidelity  \\
        \midrule
        Aer4K-P & Llama-3.2-Vision \cite{llama32vision} & True & None & None & False & False & Perception  \\
        \midrule
        Aer4K-F & Llama-3.2-Vision \cite{llama32vision} & True & None & None & False & False & Fidelity  \\
        \midrule
        ExpSR-s4-P & Llama-3.2-Vision \cite{llama32vision} & False & 4 & super-resolution & False & False & Perception  \\
        \midrule
        ExpSR-s4-F & Llama-3.2-Vision \cite{llama32vision} & False & 4 & super-resolution & False & False & Fidelity \\
        \midrule
        ExpSR-s2-F & Llama-3.2-Vision \cite{llama32vision} & False & 2 & super-resolution & False & False & Fidelity  \\
        \midrule
        ExpSR-s8-F & Llama-3.2-Vision \cite{llama32vision} & False & 8 & super-resolution & False & False & Fidelity  \\
        \midrule
        GenSR-s4-P & DepictQA \cite{you2024depicting} & False & 4 & None & False & False & Perception  \\
        \midrule
        GenMIR-P & DepictQA \cite{you2024depicting} & False & 4 & None & False & True & Perception \\
        \midrule
        ExpSRFR-s4-P & Llama-3.2-Vision \cite{llama32vision} & False & 4 & super-resolution & True & False & Perception  \\
        \midrule
        GenSRFR-s4-P & DepictQA \cite{you2024depicting} & False & 4 & None & True & False & Perception \\
        \bottomrule
    \end{tabular}
    }
    \vspace{-5pt}
\end{table}

\textbf{Profile naming convention:} We combine restoration type, restoration task, and restoration preference to construct the profile name. For example, \textbf{Gen} indicates a \underline{Gen}eral image, \textbf{4K} indicates ``Upscale to \underline{4K}'' on, and \textbf{P} indicates to restore the image with high \underline{P}erceptual quality. \textbf{Aer} indicates \underline{Aer}ial image, \textbf{Exp} corresponds to \underline{Exp}licit setting, indicating that the profile has explicitly set the restoration task (\eg, \textbf{SR}, which indicates \underline{S}uper-\underline{R}esolution). \textbf{MIR} indicates \underline{M}ultiple \underline{I}mage \underline{R}estoration. \textbf{FR} indicates \underline{F}ace \underline{R}estoration. \textbf{s4} indicates to upscale the image by a \underline{s}cale factor of \underline{4}. 

4KAgent supports various VLMs or LLMs in the Perception Agent, enabling effective analysis of image content and degradation. 
Specifically, users can select either DepictQA \cite{you2024depicting} or Llama-3.2-Vision (11B) \cite{llama32vision} as available options, but can also be extended to other more recent VLM models, \eg, Qwen2.5-VL~\cite{bai2025qwen2}. 
For the VLMs or LLMs to schedule the restoration plan, users can choose from GPT-4 \cite{achiam2023gpt}, or Llama-3.2-Vision. 
This is configured by the \textbf{Perception Agent} in the profile module. For example, when it is set to \textbf{Llama-3.2-Vision}, the Llama-3.2-Vision model serves as the core engine to perceive image content and degradation, and then schedules the restoration plan $P_{I}$. As DepictQA is fine-tuned for image quality assessment (IQA), when it is set as the VLM in the perception agent, 4KAgent will use Llama-3.2-Vision to obtain the image description $C_{I}$ and use GPT-4 \cite{achiam2023gpt} to schedule the restoration plan.

\subsection{Model Zoo}
\label{ssec:model-zoo}

The 4KAgent system supports nine distinct image restoration models in the toolbox: \ding{182} Brightening, \ding{183} Defocus Deblurring, \ding{184} Motion Deblurring, \ding{185} Dehazing, \ding{186} Denoising, \ding{187} Deraining, \ding{188} JPEG Compression Artifact Removal (JPEG CAR), \ding{189} Super Resolution, and \ding{190} Face Restoration. 
For each of these tasks, we integrate advanced state-of-the-art methods into our comprehensive restoration toolbox. Detailed correspondences between restoration tasks and their respective methods are presented below, where `QF' denotes the Quality Factor and `BQF' indicates methods that are blind to the Quality Factor in the JPEG CAR task.

\noindent
\begin{minipage}[t]{0.245\textwidth}
\vspace{-6pt}
  \begin{taskblock}[top=6pt,bottom=6pt,before skip=2pt,after skip=4pt, colback=PastelRed, colframe=PastelRedFrame]{Brightening}
    \begin{itemize}[nosep,left=0pt,label={}]
        \item CLAHE \cite{zuiderveld1994contrast}    
        \item Constant Shift (C=40)  
        \item DiffPlugin \cite{liu2024diff} 
        \item FourierDiff \cite{lv2024fourier} 
        \item Gamma Correction \\ ($\gamma=2/3$)    
        \item MAXIM \cite{tu2022maxim}   
    \end{itemize}
  \end{taskblock}

  \begin{taskblock}[top=6pt,bottom=6pt,before skip=2pt,after skip=2pt, colback=PastelBlue, colframe=PastelBlueFrame]{Denoising}
    \begin{itemize}[nosep,left=0pt,label={}]
        \item MAXIM \cite{tu2022maxim}   
        \item MPRNet \cite{zamir2021multi}
        \item NAFNet \cite{chen2022simple}
        \item Restormer \cite{zamir2022restormer}
        \item X-Restormer \cite{chen2024comparative}   
        \item SwinIR \cite{liang2021swinir}   
    \end{itemize}
  \end{taskblock}
\end{minipage}
\hfill
\begin{minipage}[t]{0.245\textwidth}
\vspace{-6pt}
  \begin{taskblock}[top=6pt,bottom=6pt,before skip=2pt,after skip=4pt, colback=PastelOrange, colframe=PastelOrangeFrame]{Defocus Deblurring}
    \begin{itemize}[nosep,left=0pt,label={}]
        \item ConvIR \cite{cui2024revitalizing}   
        \item DiffPlugin \cite{liu2024diff}  
        \item DRBNet \cite{ruan2022learning} 
        \item IFAN \cite{lee2021iterative} 
        \item LaKDNet \cite{ruan2023revisiting}    
        \item Restormer \cite{zamir2022restormer}    
    \end{itemize}
  \end{taskblock}

  \begin{taskblock}[top=6pt,bottom=6pt,before skip=2pt,after skip=2pt, colback=PastelPurple, colframe=PastelPurpleFrame]{Motion Deblurring}
    \begin{itemize}[nosep,left=0pt,label={}]
        \item EVSSM \cite{kong2024efficient}   
        \item LaKDNet \cite{ruan2023revisiting}  
        \item MAXIM \cite{tu2022maxim}  
        \item MPRNet \cite{zamir2021multi}
        \item NAFNet \cite{chen2022simple} 
        \item Restormer \cite{zamir2022restormer} 
        \item X-Restormer \cite{chen2024comparative}    
    \end{itemize}
  \end{taskblock}
\end{minipage}
\hfill
\begin{minipage}[t]{0.245\textwidth}
\vspace{-6pt}
  \begin{taskblock}[before skip=2pt,after skip=1pt, colback=PastelYellow, colframe=PastelYellowFrame]{Dehazing}
    \begin{itemize}[nosep,left=0pt,label={}]
        \item DehazeFormer \cite{song2023vision}   
        \item DiffPlugin \cite{liu2024diff}  
        \item MAXIM \cite{tu2022maxim} 
        \item RIDCP \cite{wu2023ridcp} 
        \item X-Restormer \cite{chen2024comparative}    
    \end{itemize}
  \end{taskblock}

  \begin{taskblock}[before skip=2pt,after skip=1pt, colback=PastelBrown,   colframe=PastelBrownFrame]{Deraining}
    \begin{itemize}[nosep,left=0pt,label={}]
        \item DiffPlugin \cite{liu2024diff}   
        \item MAXIM \cite{tu2022maxim}   
        \item MPRNet \cite{zamir2021multi}  
        \item Restormer \cite{zamir2022restormer} 
        \item X-Restormer \cite{chen2024comparative}    
    \end{itemize}
  \end{taskblock}

  \begin{taskblock}[before skip=2pt,after skip=1pt, colback=PastelPink,   colframe=PastelPinkFrame]{Face Restoration}
    \begin{itemize}[nosep,left=0pt,label={}]
        \item GFPGAN \cite{wang2021towards}  
        \item CodeFormer \cite{zhou2022towards}  
        \item DifFace \cite{yue2024difface}  
    \end{itemize}
  \end{taskblock}
\end{minipage}
\hfill
\begin{minipage}[t]{0.245\textwidth}
\vspace{-6pt}
  \begin{taskblock}[top=2pt,bottom=2pt,before skip=2pt,after skip=4pt, colback=PastelGreen, colframe=PastelGreenFrame]{Super Resolution}
    \begin{itemize}[nosep,left=0pt,label={}]
        \item DiffBIR \cite{lin2024diffbir}   
        \item DRCT \cite{hsu2024drct}  
        \item HAT-L \cite{chen2023hat}
        \item HAT-GAN \cite{chen2023hat}
        \item HMA \cite{chu2024hmanet} 
        \item OSEDiff \cite{wu2024one} 
        \item PiSA-SR \cite{sun2024pixel}    
        \item SwinIR \cite{liang2021swinir} 
        \item SwinIR (Real-ISR) \cite{liang2021swinir}
        \item SwinFIR \cite{zhang2022swinfir}   
        \item X-Restormer \cite{chen2024comparative}   
    \end{itemize}
  \end{taskblock}

  \begin{taskblock}[top=2pt,bottom=2pt, before skip=2pt,after skip=2pt, colback=PastelYellow, colframe=PastelYellowFrame]{JPEG CAR}
    \begin{itemize}[nosep,left=0pt,label={}]
        \item FBCNN \cite{jiang2021towards} (QF=5)   
        \item FBCNN \cite{jiang2021towards} (QF=90)  
        \item FBCNN \cite{jiang2021towards} (BQF) 
        \item SwinIR \cite{liang2021swinir} (QF=40) 
    \end{itemize}
  \end{taskblock}
\end{minipage} 

As previously mentioned in~\cref{ssec:profile}, users can tailor the model by adjusting the \textbf{Restore Preference} setting, which prioritizes either perceptual quality or fidelity. We achieve this by partitioning our toolbox methods into perception-oriented and fidelity-oriented categories. For example, the Super-Resolution tools in the toolbox are split into:
\vspace{-5pt}
\begin{enumerate}[leftmargin=*,noitemsep]
\item \textbf{Fidelity-based:} HAT-L \cite{chen2023hat}, X-Restormer \cite{chen2024comparative}, SwinFIR \cite{zhang2022swinfir}, HMA \cite{chu2024hmanet}, DRCT \cite{hsu2024drct}
\item \textbf{Perception-based:} DiffBIR \cite{lin2024diffbir}, HAT-GAN \cite{chen2023hat}, OSEDiff \cite{wu2024one}, PiSA-SR \cite{sun2024pixel}, SwinIR (Real-ISR) \cite{liang2021swinir} 
\end{enumerate}
Therefore, when \textbf{Restore Preference} is set to \textbf{Perception}, 4KAgent will only use the Perception-based methods to restore the image, efficiently meeting the user’s restoration requirements.

We develop a \textbf{Fast4K} mode for 4KAgent. Specifically, when the size of the input image at the current step of the restoration plan exceeds a predefined threshold $s_{t}$, 4KAgent automatically excludes methods with long inference times from the toolbox, such as DiffBIR (a 50-step diffusion-based method) in the super-resolution toolbox. Users can adjust $s_{t}$ to control the running time of 4KAgent.

\subsection{Prompts}
In 4KAgent, we enable the VLM / LLM to perceive image degradations and formulate a restoration plan via customized system prompts. In this section, we present the details of these prompts in 4KAgent.
When the \textbf{Perception Agent} in the profile module selects \textbf{DepictQA}, we use the same prompt as in AgenticIR \cite{zhu2024intelligent} for DepictQA to assess the image degradations, and GPT-4 to construct the restoration plan. When setting \textbf{Llama-3.2-Vision} in \textbf{Perception Agent}, we design tailored prompts for degradation reasoning and planning, as shown below, where \{$\cdot$\} represents slots to fill according to the context, and the content inside comes from external input. For the restoration experience $E$ in 4KAgent, we employ the restoration experience from AgenticIR.
\begin{tcolorbox}[colback=gray!5!white,colframe=black!50!gray, fontupper=\small, title=Prompt for Llama-Vision in Degradation Reasoning]
        \texttt{You are an expert tasked with image quality assessment (IQA) and well-versed in popular IQA metrics, including CLIPIQA+, TOPIQ\_NR, MUSIQ, and NIQE. Note that for NIQE, a lower score indicates better image quality, whereas for the other metrics, higher scores generally reflect better quality. Here's an image to restore, along with its corresponding quality scores evaluated using the aforementioned IQA metrics. First, please describe the content and style of the input image, the description must not contain its image quality. Second, please assess the image based on both the metric scores and your prior visual knowledge, with respect to the following two degradations: noise, motion blur, defocus blur, haze, rain, jpeg compression artifact. Images may suffer from one or more of these degradations. **Do not output any explanations or comments.** **Strictly return only a JSON object** containing degradation types and image content/style description. The keys in the JSON object should be: `degradations` and `image\_description`. Information about the input image: IQA metrics: \textbf{\{iqa\_result\}}. (\textbf{\{iqa\_result\}} corresponds to $Q_{I}$ in~\cref{ssec:perception_agent}.)} 
\end{tcolorbox}

\begin{tcolorbox}[colback=gray!5!white,colframe=black!50!gray, fontupper=\small, title=Prompt for Llama-Vision in Planning (Rollback)]
        \texttt{You are an expert in image restoration. Given an image of low quality, your task is to guide the user to utilize various tools to enhance its quality. The input image requires a list of restoration tasks. Your goal is to make a plan (the order of the tasks) based on the task list. The final output should be formatted as a JSON object containing the restoration plan (the correct order of the tasks). The key in the JSON object should be: `plan`. Information about the input image: }  \\
        \texttt{Its description is: \textbf{\{image\_description\}} ($C_{I}$),} \\
        \texttt{It suffer from degradations \textbf{\{degradations\}} ($D_{I}$),} \\
        \texttt{The list of restoration tasks: \textbf{\{tasks\}} ($A_{I}$ / $A_{I}^{R}$),} \\
        \texttt{For your information, based on past trials, we have the following experience in making a restoration plan: \textbf{\{experience\}} ($E$). Based on this experience, please give the correct order of the tasks in the restoration plan. The restoration plan must be a permutation of \textbf{\{tasks\}} in the order you determine. (Besides, in attempts just now, we found the result is unsatisfactory if \textbf{\{failed\_tries\}} ($S_{I}$)} \\
        \texttt{is conducted first. Remember not to arrange \textbf{\{failed\_tries\}} in the first place.) **Do not output any explanations or comments.** **Strictly return only a JSON object** containing plan. The keys in the JSON object should be: `plan`.} 
\end{tcolorbox}

\subsection{Implementation Details}
\textbf{Computing Resource.} As a multi-agent system, 4KAgent supports multi-GPU deployment. Specifically, 4KAgent assigns different agents (Perception Agent, Restoration Agent) on different GPUs to conserve memory. Most of our experiments were conducted using two NVIDIA RTX 4090 GPUs.

\textbf{Hyper-parameters.} Hyperparameters in 4KAgent reside in the Restoration Agent, namely the weights used to compute the execution quality scores $Q_{s}$ and $Q_s^{f}$ in execution, as well as the quality threshold $\eta$ in rollback. Specifically, in 4KAgent, we set $w_{\text{NIQE}}=1.0, w_{\text{MUSIQ}}=0.01, w_{\text{MANIQA}}=1.0, w_{\text{CLIPIQA}}=1.0$ for $Q_{s}$, $w_{\text{IP}}=0.001, w_{\text{IQA}}=1.0$ for $Q_{s}^{f}$, and $\eta=0.5$ for rollback.

\section{Experiment Overview}
\label{sec:experiment-p}
We evaluate 4KAgent on a variety of complex degradation and super-resolution (SR) tasks, demonstrate its flexible profile-driven modes for different restoration requirements, validate its generalization to multiple image domains, and quantify the contributions of each core component via ablation studies.
Specifically, we test 4KAgent on a wide range of \textbf{11} image SR tasks on \textbf{26} benchmarks. The summary of datasets used in experiments is shown in~\cref{table:benchmarks}, which can be classified as natural degraded images (\S\ref{sec:experiment-i},\ref{sec:experiment-ii}), AI-generated images (\S\ref{sec:experiment-iii}), and scientific images (\S\ref{sec:experiment-iv}). Then, we perform an ablation study on \textbf{Q-MoE} policy and \textbf{Face restoration pipeline} in 4KAgent with a runtime analysis (\S\ref{sec:ablation}).

First, we evaluate 4KAgent on natural image restoration / super-resolution tasks under general settings, including classical image super-resolution (4$\times$) (\S\ref{ssec:classical-sr}), real-world image super-resolution (4$\times$) (\S\ref{ssec:real-sr}), multiple-degradation image restoration (\S\ref{ssec:multiple-sr}), and face restoration (4$\times$) (\S\ref{ssec:face-sr}). Next, we assess its performance in more challenging scenarios, such as large scale factor super-resolution (16$\times$) (\S\ref{ssec:16x-sr}) and joint restoration with 4K upscaling (\S\ref{ssec:div4k-sr}). Finally, we extend 4KAgent to diverse domains by testing its capabilities on AIGC images (\S\ref{ssec:aigc-sr}) and scientific imagery (\S\ref{sec:experiment-iv}), including remote sensing (\S\ref{ssec:sci-sr-remote-sensing}), microscopy (\S\ref{ssec:sci-sr-ii-micro}), pathology (\S\ref{ssec:sci-sr-iii-patho}), and medical images (\S\ref{ssec:sci-sr-iiii-med}). To comprehensively evaluate the performance of 4KAgent, we disable the Fast4K mode in all our experiments.

\section{Experiment Part I: 4$\times$ Natural Image Super-Resolution}
\label{sec:experiment-i}
\subsection{Classical Image Super-Resolution}
\label{ssec:classical-sr}
\begin{wraptable}[34]{r}{0.53\textwidth}
\vspace{-\intextsep}
    \centering
    \renewcommand{\arraystretch}{1.13}
    \fontsize{6.5pt}{9.5pt}\selectfont    \setlength{\tabcolsep}{2.5pt}
    \caption{Dataset summary in 4KAgent experiments.}
    \label{table:benchmarks}
    \begin{tabular}{l |l r}
        \toprule
        \textbf{Task} & \textbf{Dataset} & \textbf{\#Test Images} \\
        \midrule
         & Set5 \cite{bevilacqua2012low} & 5 \\
         & Set14 \cite{zeyde2010single} & 14 \\
         \textbf{Classical SR (\S\ref{ssec:classical-sr})} & B100 \cite{martin2001database} & 100 \\
         & Urban100 \cite{huang2015single} & 100 \\
         & Manga109 \cite{matsui2017sketch} & 109 \\
        \midrule
        \textbf{Real-World SR (\S\ref{ssec:real-sr})} & RealSR \cite{cai2019toward} & 93 \\
         & DrealSR \cite{wei2020component} & 100 \\
        \midrule
         & MiO-Group A \cite{zhu2024intelligent} & 640 \\
         \textbf{Multiple-Degradation IR (\S\ref{ssec:multiple-sr})} & MiO-Group B \cite{zhu2024intelligent} & 400 \\
         & MiO-Group C \cite{zhu2024intelligent} & 400 \\
        \midrule
        \textbf{Face Restoration (\S\ref{ssec:face-sr})} & WebPhoto-Test \cite{wang2021towards} & 407 \\
        \midrule
        \textbf{Large Scale Factor SR (\S\ref{ssec:16x-sr})} & RealSRSet \cite{zhang2021designing} & 20 \\
        \midrule
        \textbf{Joint Image Restoration + } & DIV4K-50 (Ours) & 50 \\
        \textbf{4K Upscaling (\S\ref{ssec:div4k-sr})} & &  \\
        \midrule
        \textbf{AI-Generated  } & GenAIBench-4K~\cite{li2024genaibench} & 100 \\
        \textbf{Content 4K SR (\S\ref{ssec:aigc-sr})} & DiffusionDB-4K~\cite{wang2022diffusiondb} & 100 \\
        \midrule
         & AID~\cite{7907303} & 135 \\
         \textbf{Remote Sensing SR (\S\ref{ssec:sci-sr-remote-sensing})} & DIOR~\cite{li2020object}  & 154 \\
         & DOTA~\cite{8578516} & 183 \\
         & WorldStrat~\cite{cornebise2022open} & 100 \\
        \midrule
        \textbf{Fluorescence Microscopy} & SR-CACO-2~\cite{belharbi2024sr} & 300 \\
        \textbf{Image SR (\S\ref{ssec:sci-sr-ii-micro})} &  &  \\
        \midrule
        \textbf{Pathology Image SR (\S\ref{ssec:sci-sr-iii-patho})} & bcSR~\cite{jia2023channelattention} & 200 \\
        \midrule
         & Chest X-ray 2017~\cite{kermany2018identifying} & 624 \\
         & Chest X-ray 14~\cite{wang2017chestx}  & 880 \\
         \textbf{Medical Image SR (\S\ref{ssec:sci-sr-iiii-med})} & US-Case~\cite{US-CASE2025} & 111 \\
         & MMUS1K~\cite{Ultrasound} & 100 \\
         & DRIVE~\cite{staal2004ridge} & 20 \\
        \bottomrule
    \end{tabular}
\end{wraptable}
\paragraph{Settings.} In this experiment, we follow the standard super-resolution (SR) evaluation protocol~\cite{zhang2018image, liang2021swinir}, and assess the performance of 4KAgent on widely-used benchmark datasets, including Set5~\cite{bevilacqua2012low}, Set14~\cite{zeyde2010single}, B100~\cite{martin2001database}, Urban100~\cite{huang2015single}, and Manga109~\cite{matsui2017sketch}. In addition to PSNR and SSIM~\cite{wang2004image}, we adopt a range of quality metrics for a more comprehensive evaluation, including LPIPS~\cite{zhang2018unreasonable}, DISTS~\cite{ding2020image}, FID~\cite{heusel2017gans}, NIQE~\cite{mittal2012making}, CLIPIQA~\cite{wang2023exploring}, MUSIQ~\cite{ke2021musiq}, and MANIQA-pipal~\cite{yang2022maniqa}. 

Specifically, PSNR and SSIM are computed on the Y channel in the YCbCr space. They are used to measure the fidelity of images. LPIPS and DISTS are computed in the RGB space, are used to measure the perceptual quality of images. FID is used to evaluate the distance of distributions between the ground truth and the restored images. NIQE, CLIPIQA, MUSIQ, and MANIQA-pipal are used to evaluate the perceptual quality of images without reference images.

Thanks to the high flexibility of our 4KAgent, which is governed by configurable profiles, we customize the 4KAgent in this experiment using three specific profiles: \textbf{ExpSR-s4-F}, \textbf{ExpSR-s4-P}, and \textbf{GenSR-s4-P}.

For comparison, we employ state-of-the-art fidelity-based methods (\eg, SwinIR \cite{liang2021swinir}, X-Restormer \cite{chen2024comparative}, HAT-L \cite{chen2023hat}) and perception-based methods (\eg, DiffBIR \cite{lin2024diffbir}, OSEDiff \cite{wu2024one}, PiSA-SR \cite{sun2024pixel}). In addition, we include AgenticIR \cite{zhu2024intelligent} in this experiment for agentic system comparison.

\paragraph{Quantitative Comparison}
It should be noted that, once the user sets the \textbf{Restore Option} to \textbf{super-resolution} in the profile, the 4KAgent system can be seen as a quality-driven Mixture-of-Expert system for image super-resolution. In this mode, the system sequentially invokes every super-resolution tool in its toolbox based on the \textbf{Restore Option} setting in the profile, then selects the best result based on the quality score $Q_{s}$. Accordingly, we group 4KAgent with \textbf{ExpSR-s4-F} and \textbf{ExpSR-s4-P} profile to \textbf{\textit{Fidelity based method}} and \textbf{\textit{Perception based method}}. 
\begin{table*}[!h]
\renewcommand{\arraystretch}{1.2}
\centering
\fontsize{6.5pt}{7.5pt}\selectfont
\setlength{\tabcolsep}{4pt}
\caption{Quantitative comparison on classical image super-resolution benchmarks (Set5, Set14, B100). The top three performances of each metric are marked in \textbf{bold}, \underline{underline}, \textit{italic} respectively. For Agentic systems, we only \textbf{bold} the best performance.}
\label{table:performance_classicsr_set5}
\begin{tabular}{l|l|ccccccccc}
\toprule
{\textbf{Dataset}} & {\textbf{Method}}   & \textbf{PSNR$\uparrow$}&  \textbf{SSIM$\uparrow$} &  \textbf{LPIPS$\downarrow$}  &  \textbf{DISTS$\downarrow$}  &  \textbf{FID$\downarrow$}  &  \textbf{NIQE$\downarrow$}  &  \textbf{CLIPIQA$\uparrow$} & \textbf{MUSIQ$\uparrow$}  &  \textbf{MANIQA$\uparrow$}      \\
\midrule
\multirow{18}{*}{Set5}
& \textbf{\textit{\underline{Fidelity based method}}} \\
& SwinIR \cite{liang2021swinir} & 32.92 & 0.9044 & 0.1669 & 0.1567 & 57.37 & 7.24 & 0.6179 & 59.98 & 0.6095   \\
& X-Restormer \cite{chen2024comparative} & 33.15 & 0.9057 & 0.1636 & 0.1564 & 60.24 & \textit{7.07} & \textit{0.6368} & 60.09 & 0.6169 \\ 
& DRCT \cite{hsu2024drct} & 33.26 & 0.9067 & 0.1616 & \textit{0.1526} & \textbf{52.25} & \underline{6.94} & \textbf{0.6406} & \textit{60.21} & 0.6100   \\ 
& HAT-L \cite{chen2023hat} & \textit{33.29} & \underline{0.9082} & \textbf{0.1582} & \underline{0.1542} & 56.95 & 7.11 & \underline{0.6389} & \textbf{60.44} & \underline{0.6212}  \\ 
& HMA \cite{chu2024hmanet} & \textbf{33.39} & \textbf{0.9089} & \underline{0.1587} & \textbf{0.1535} & \underline{54.61} & 7.11 & 0.6338 & \underline{60.39} & \textbf{0.6241}   \\
& \cellcolor{LightGray!70}\textbf{4KAgent (ExpSR-s4-F)} & \cellcolor{LightGray!70}\underline{33.34} & \cellcolor{LightGray!70}\textit{0.9081} & \cellcolor{LightGray!70}\textit{0.1589} & \cellcolor{LightGray!70}0.1549 & \cellcolor{LightGray!70}\textit{56.62} & \cellcolor{LightGray!70}\textbf{6.90} & \cellcolor{LightGray!70}0.6294 & \cellcolor{LightGray!70}60.02 & \cellcolor{LightGray!70}\textit{0.6177}   \\
\cmidrule{2-11} 
& \textbf{\textit{\underline{Perception based method}}} \\
& SwinIR (Real-ISR) \cite{liang2021swinir}  & \textbf{28.48} & \textbf{0.8446} & 0.1632 & \underline{0.1590} & \underline{63.58} & 7.46 & 0.7072 & 62.43 & 0.6153   \\
& DiffBIR \cite{lin2024diffbir} & 26.41 & 0.7510 & 0.2059 & 0.1888 & 72.79 & 6.06 & \textbf{0.8405} & \textbf{70.23} & \textit{0.6767}  \\  
& OSEDiff \cite{wu2024one} & 26.21 & \textit{0.8063} & \underline{0.1583} & \textit{0.1647} & \textit{67.50} & \textbf{5.78} & 0.7973 & 68.76 & 0.6698   \\
& PiSA-SR \cite{sun2024pixel} & \underline{27.56} & \underline{0.8189} & \textbf{0.1318} & \textbf{0.1516} & \textbf{62.94} & \textit{5.87} & \textit{0.8086} & \textit{69.87} & \textbf{0.6904}   \\  
& \cellcolor{LightGray!70}\textbf{4KAgent (ExpSR-s4-P)} & \cellcolor{LightGray!70}\textit{26.88} & \cellcolor{LightGray!70}0.7899 & \cellcolor{LightGray!70}\textit{0.1591} & \cellcolor{LightGray!70}0.1657 & \cellcolor{LightGray!70}70.63 & \cellcolor{LightGray!70}\underline{5.79} & \cellcolor{LightGray!70}\underline{0.8245} & \cellcolor{LightGray!70}\underline{69.93} & \cellcolor{LightGray!70}\underline{0.6808}   \\ 
\cmidrule{2-11} 
& \textbf{\textit{\underline{Agentic System}}} \\
& AgenticIR \cite{zhu2024intelligent} & 23.68 & 0.6711 & 0.2737 & 0.2190 & 124.96 & \textbf{6.59} & \textbf{0.7750} & \textbf{71.88} & \textbf{0.7079}   \\
& \cellcolor{LightGray!70}\textbf{4KAgent (GenSR-s4-P)} & \cellcolor{LightGray!70}\textbf{26.25} & \cellcolor{LightGray!70}\textbf{0.7672} & \cellcolor{LightGray!70}\textbf{0.1785} & \cellcolor{LightGray!70}\textbf{0.1836} & \cellcolor{LightGray!70}\textbf{89.02} & \cellcolor{LightGray!70}6.72 & \cellcolor{LightGray!70}0.7396 & \cellcolor{LightGray!70}70.39 & \cellcolor{LightGray!70}0.6811   \\
\midrule
\multirow{18}{*}{Set14}
& \textbf{\textit{\underline{Fidelity based method}}} \\
& SwinIR \cite{liang2021swinir} & 29.09 & 0.7950 & 0.2671 & 0.1574 & 70.49 & 6.19 & 0.5252 & 63.10 & 0.5891   \\
& X-Restormer \cite{chen2024comparative} & 29.16 & 0.7963 & 0.2659 & 0.1557 & 69.86 & 6.22 & \underline{0.5332} & 62.91 & 0.5925 \\ 
& DRCT \cite{hsu2024drct} & \textbf{29.57} & \textit{0.8009} & 0.2617 & \textit{0.1524} & \textit{67.84} & \underline{6.09} & \textbf{0.5362} & \textit{63.12} & 0.5932   \\ 
& HAT-L \cite{chen2023hat} & \textit{29.46} & \underline{0.8014} & \textbf{0.2565} & \underline{0.1516} & \textbf{66.61} & \textit{6.11} & 0.5267 & \underline{63.23} & \underline{0.5986}  \\ 
& HMA \cite{chu2024hmanet} & \underline{29.51} & \textbf{0.8019} & \underline{0.2567} & \textbf{0.1510} & 69.41 & 6.25 & 0.5278 & 63.00 & \textbf{0.6012}   \\
& \cellcolor{LightGray!70}\textbf{4KAgent (ExpSR-s4-F)} & \cellcolor{LightGray!70}29.43 & \cellcolor{LightGray!70}0.7989 & \cellcolor{LightGray!70}\textit{0.2593} & \cellcolor{LightGray!70}0.1528 & \cellcolor{LightGray!70}\underline{67.83} & \cellcolor{LightGray!70}\textbf{5.95} & \cellcolor{LightGray!70}\textit{0.5315} & \cellcolor{LightGray!70}\textbf{63.45} & \cellcolor{LightGray!70}\textit{0.5970}   \\ 
\cmidrule{2-11} 
& \textbf{\textit{\underline{Perception based method}}} \\
& SwinIR (Real-ISR) \cite{liang2021swinir}  & \textbf{25.91} & \textbf{0.7187} & \textit{0.2244} & \textit{0.1508} & \underline{96.19} & 4.45 & 0.6506 & 66.82 & 0.6054   \\ 
& DiffBIR \cite{lin2024diffbir} & 24.73 & 0.6349 & 0.2338 & 0.1545 & \textit{100.51} & \textit{4.34} & \textit{0.7553} & \textbf{72.97} & \textit{0.6869}  \\  
& OSEDiff \cite{wu2024one} & 24.30 & \textit{0.6663} & 0.2389 & 0.1524 & 101.03 & 4.61 & 0.7264 & 70.02 & 0.6674   \\
& PiSA-SR \cite{sun2024pixel} & \textit{24.76} & \underline{0.6716} & \textbf{0.1993} & \textbf{0.1343} & \textbf{89.91} & \underline{4.16} & \underline{0.7643} & \textit{71.81} & \textbf{0.7015}   \\ 
& \cellcolor{LightGray!70}\textbf{4KAgent (ExpSR-s4-P)} & \cellcolor{LightGray!70}\underline{24.76} & \cellcolor{LightGray!70}0.6471 & \cellcolor{LightGray!70}\underline{0.2158} & \cellcolor{LightGray!70}\underline{0.1467} & \cellcolor{LightGray!70}101.99 & \cellcolor{LightGray!70}\textbf{4.01} & \cellcolor{LightGray!70}\textbf{0.7740} & \cellcolor{LightGray!70}\underline{72.54} & \cellcolor{LightGray!70}\underline{0.6956}   \\
\cmidrule{2-11} 
& \textbf{\textit{\underline{Agentic System}}} \\
& AgenticIR \cite{zhu2024intelligent} & 21.98 & 0.6064  & 0.2807 & 0.1812 & 129.29 & 4.58 & 0.7449 & 72.48 & 0.6804   \\
& \cellcolor{LightGray!70}\textbf{4KAgent (GenSR-s4-P)} & \cellcolor{LightGray!70}\textbf{23.40} & \cellcolor{LightGray!70}\textbf{0.6340} & \cellcolor{LightGray!70}\textbf{0.2484} & \cellcolor{LightGray!70}\textbf{0.1749} & \cellcolor{LightGray!70}\textbf{125.29} & \cellcolor{LightGray!70}\textbf{4.29} & \cellcolor{LightGray!70}\textbf{0.7604} & \cellcolor{LightGray!70}\textbf{73.64} & \cellcolor{LightGray!70}\textbf{0.7061}  \\
\midrule
\multirow{18}{*}{B100}
& \textbf{\textit{\underline{Fidelity based method}}} \\
& SwinIR \cite{liang2021swinir} & 27.92 & 0.7489 & 0.3548 & 0.2005 & 94.57 & 6.27 & 0.5373 & 57.71 & 0.5860   \\
& X-Restormer \cite{chen2024comparative} & 27.99 & 0.7508 & 0.3521 & 0.1972 & 90.52 & 6.21 & 0.5427 & 57.91 & 0.5935 \\ 
& DRCT \cite{hsu2024drct} & \underline{28.10} & 0.7535 & 0.3480 & \textbf{0.1947} & \textbf{87.76} & \underline{6.06} & \textit{0.5499} & \textit{58.78} & 0.5895   \\ 
& HAT-L \cite{chen2023hat} & 28.08 & \underline{0.7547} & \textbf{0.3440} & \textit{0.1952} & 89.52 & 6.20 & 0.5477 & 58.71 & \textit{0.5991}  \\ 
& HMA \cite{chu2024hmanet} & \textbf{28.12} & \textbf{0.7559} & \underline{0.3442} & 0.1953 & \underline{88.46} & \textit{6.17} & \textbf{0.5534} & \underline{59.11} & \textbf{0.6043}   \\
& \cellcolor{LightGray!70}\textbf{4KAgent (ExpSR-s4-F)} & \cellcolor{LightGray!70}\textit{28.09} & \cellcolor{LightGray!70}\textit{0.7540} & \cellcolor{LightGray!70}\textit{0.3453} & \cellcolor{LightGray!70}\underline{0.1950} & \cellcolor{LightGray!70}\textit{88.89} & \cellcolor{LightGray!70}\textbf{6.02} & \cellcolor{LightGray!70}\underline{0.5516} & \cellcolor{LightGray!70}\textbf{59.12} & \cellcolor{LightGray!70}\underline{0.5994}   \\
\cmidrule{2-11} 
& \textbf{\textit{\underline{Perception based method}}} \\
& SwinIR (Real-ISR) \cite{liang2021swinir}  & \textbf{25.42} & \textbf{0.6711} & 0.2500 & 0.1699 & 92.65 & \textit{4.00} & 0.6322 & 62.78 & 0.6085   \\
& DiffBIR \cite{lin2024diffbir} & \textit{24.99} & 0.6156 & 0.2719 & 0.1666 & 84.99 & \underline{3.92} & \underline{0.7483} & 68.23 & \textit{0.6750}  \\  
& OSEDiff \cite{wu2024one} & 24.35 & \textit{0.6495} & \textit{0.2408} & \textit{0.1634} & \underline{73.23} & 4.08 & \textit{0.7422} & \underline{68.54} & 0.6725   \\
& PiSA-SR \cite{sun2024pixel} & \underline{25.00} & \underline{0.6520} & \textbf{0.2111} & \textbf{0.1471} & \textbf{61.82} & 4.04 & 0.7384 & \textit{68.47} & \underline{0.6829}   \\  
& \cellcolor{LightGray!70}\textbf{4KAgent (ExpSR-s4-P)} & \cellcolor{LightGray!70}24.64 & \cellcolor{LightGray!70}0.6294 & \cellcolor{LightGray!70}\underline{0.2387} & \cellcolor{LightGray!70}\underline{0.1606} & \cellcolor{LightGray!70}\textit{73.64} & \cellcolor{LightGray!70}\textbf{3.86} & \cellcolor{LightGray!70}\textbf{0.7546} & \cellcolor{LightGray!70}\textbf{69.42} & \cellcolor{LightGray!70}\textbf{0.6851}   \\  
\cmidrule{2-11} 
& \textbf{\textit{\underline{Agentic System}}} \\
& AgenticIR \cite{zhu2024intelligent} & 22.51 & 0.5853 & 0.3078 & 0.1907 & 102.92 & 4.08 & \textbf{0.7474} & 68.36 & 0.6752   \\
& \cellcolor{LightGray!70}\textbf{4KAgent (GenSR-s4-P)} & \cellcolor{LightGray!70}\textbf{23.64} & \cellcolor{LightGray!70}\textbf{0.6246} & \cellcolor{LightGray!70}\textbf{0.2572} & \cellcolor{LightGray!70}\textbf{0.1702} & \cellcolor{LightGray!70}\textbf{78.80} & \cellcolor{LightGray!70}\textbf{3.93} & \cellcolor{LightGray!70}0.7354 & \cellcolor{LightGray!70}\textbf{69.44} & \cellcolor{LightGray!70}\textbf{0.6844}  \\
\bottomrule
\end{tabular}
\end{table*}

\begin{table*}[!h]
\renewcommand{\arraystretch}{1.2}
\centering
\fontsize{6.5pt}{7.5pt}\selectfont
\setlength{\tabcolsep}{4pt}
\caption{Quantitative comparison on classical image super-resolution benchmarks (Urban100 and Manga109). The top three performances of each metric are marked in \textbf{bold}, \underline{underline}, \textit{italic} respectively. For Agentic systems, we only \textbf{bold} the best performance.}
\label{table:performance_classicsr_urban100}
\begin{tabular}{l|l|ccccccccc}
\toprule
{\textbf{Dataset}} & {\textbf{Method}}   & \textbf{PSNR$\uparrow$}&  \textbf{SSIM$\uparrow$} &  \textbf{LPIPS$\downarrow$}  &  \textbf{DISTS$\downarrow$}  &  \textbf{FID$\downarrow$}  &  \textbf{NIQE$\downarrow$}  &  \textbf{CLIPIQA$\uparrow$} & \textbf{MUSIQ$\uparrow$}  &  \textbf{MANIQA$\uparrow$}      \\
\midrule
\multirow{18}{*}{Urban100}
& \textbf{\textit{\underline{Fidelity based method}}} \\
& SwinIR \cite{liang2021swinir} & 27.45 & 0.8254 & 0.1840 & 0.1533 & 3.58 & \textit{5.50} & 0.5003 & 70.00 & 0.6693   \\
& X-Restormer \cite{chen2024comparative} & 27.64 & 0.8288 & 0.1805 & 0.1504 & 3.65 & 5.61 & 0.4953 & 70.00 & 0.6746 \\ 
& DRCT \cite{hsu2024drct} & \textbf{28.78} & \textit{0.8492} & 0.1623 & \textbf{0.1388} & \underline{2.92} & \underline{5.45} & \textbf{0.5271} & 70.48 & 0.6778   \\ 
& HAT-L \cite{chen2023hat} & 28.58 & \underline{0.8495} & \underline{0.1598} & 0.1411 & \textbf{2.87} & 5.55 & 0.5054 & \textit{70.62} & \underline{0.6866}  \\ 
& HMA \cite{chu2024hmanet} & \underline{28.69} & \textbf{0.8511} & \textbf{0.1583} & \textit{0.1405} & \textit{2.93} & 5.61 & \textit{0.5084} & \textbf{70.75} & \textbf{0.6893}  \\
& \cellcolor{LightGray!70}\textbf{4KAgent (ExpSR-s4-F)} & \cellcolor{LightGray!70}\textit{28.59} & \cellcolor{LightGray!70}0.8479 & \cellcolor{LightGray!70}\textit{0.1599} & \cellcolor{LightGray!70}\underline{0.1399} & \cellcolor{LightGray!70}2.97 & \cellcolor{LightGray!70}\textbf{5.31} & \cellcolor{LightGray!70}\underline{0.5235} & \cellcolor{LightGray!70}\underline{70.70} & \cellcolor{LightGray!70}\textit{0.6833}   \\ 
\cmidrule{2-11} 
& \textbf{\textit{\underline{Perception based method}}} \\
& SwinIR (Real-ISR) \cite{liang2021swinir}  & \textbf{23.24} & \textbf{0.7184} & \underline{0.1908} & \underline{0.1365} & \textbf{25.36} & \textbf{4.29} & 0.6169 & 71.99 & 0.6578  \\ 
& DiffBIR \cite{lin2024diffbir} & \textit{22.51} & 0.6397 & 0.2011 & 0.1395 & \textit{26.10} & 4.79 & \textbf{0.7185} & \underline{73.10} & \textit{0.6956}  \\  
& OSEDiff \cite{wu2024one} & 21.88 & 0.6572 & 0.2185 & 0.1479 & 38.13 & 4.67 & 0.6593 & 72.35 & 0.6822   \\
& PiSA-SR \cite{sun2024pixel} & 22.36 & \underline{0.6704} & \textbf{0.1823} & \textbf{0.1297} & 28.51 & \underline{4.43} & \textit{0.6814} & \textit{72.93} & \textbf{0.7020}   \\
& \cellcolor{LightGray!70}\textbf{4KAgent (ExpSR-s4-P)} & \cellcolor{LightGray!70}\underline{22.56} & \cellcolor{LightGray!70}\textit{0.6582} & \cellcolor{LightGray!70}\textit{0.1955} & \cellcolor{LightGray!70}\textit{0.1378} & \cellcolor{LightGray!70}\underline{25.55} & \cellcolor{LightGray!70}\textit{4.53} & \cellcolor{LightGray!70}\underline{0.7092} & \cellcolor{LightGray!70}\textbf{73.65} & \cellcolor{LightGray!70}\underline{0.6981}   \\
\cmidrule{2-11} 
& \textbf{\textit{\underline{Agentic System}}} \\
& AgenticIR \cite{zhu2024intelligent} & 22.03 & \textbf{0.6615} & 0.2147 & 0.1507 & \textbf{31.09} & 4.65 & 0.6790 & 73.10 & 0.6873   \\
& \cellcolor{LightGray!70}\textbf{4KAgent (GenSR-s4-P)} & \cellcolor{LightGray!70}\textbf{22.27} & \cellcolor{LightGray!70}0.6545 & \cellcolor{LightGray!70}\textbf{0.2073} & \cellcolor{LightGray!70}\textbf{0.1444} & \cellcolor{LightGray!70}32.29 & \cellcolor{LightGray!70}\textbf{4.43} & \cellcolor{LightGray!70}\textbf{0.7001} & \cellcolor{LightGray!70}\textbf{73.57} & \cellcolor{LightGray!70}\textbf{0.6961}  \\
\midrule
\multirow{18}{*}{Manga109}
& \textbf{\textit{\underline{Fidelity based method}}} \\
& SwinIR \cite{liang2021swinir} & 32.05 & 0.9260 & 0.0926 & 0.0761 & 1.88 & 5.32 & \textbf{0.6385} & \textbf{70.32} & 0.6117   \\
& X-Restormer \cite{chen2024comparative} & 32.40 & 0.9279 & 0.0909 & 0.0748 & 1.88 & 5.48 & 0.6325 & \underline{70.05} & 0.6123 \\ 
& DRCT \cite{hsu2024drct} & 32.84 & 0.9307 & 0.0889 & 0.0685 & 1.49 & \underline{5.08} & \underline{0.6362} & 69.77 & 0.6087   \\ 
& HAT-L \cite{chen2023hat} & \underline{33.08} & \underline{0.9334} & \underline{0.0845} & \textit{0.0684} & \textit{1.48} & 5.26 & 0.6160 & 69.76 & \underline{0.6145} \\ 
& HMA \cite{chu2024hmanet} & \textbf{33.20} & \textbf{0.9344} & \textbf{0.0835} & \textbf{0.0682} & \textbf{1.47} & \textit{5.24} & 0.6208 & 69.92 & \textbf{0.6196}  \\
& \cellcolor{LightGray!70}\textbf{4KAgent (ExpSR-s4-F)} & \cellcolor{LightGray!70}\textit{32.87} & \cellcolor{LightGray!70}\textit{0.9316} & \cellcolor{LightGray!70}\textit{0.0860} & \cellcolor{LightGray!70}\underline{0.0683} & \cellcolor{LightGray!70}\underline{1.48} & \cellcolor{LightGray!70}\textbf{4.95} & \cellcolor{LightGray!70}\textit{0.6329} & \cellcolor{LightGray!70}\textit{69.99} & \cellcolor{LightGray!70}\textit{0.6125}   \\ 
\cmidrule{2-11} 
& \textbf{\textit{\underline{Perception based method}}} \\
& SwinIR (Real-ISR) \cite{liang2021swinir}  & \textbf{26.29} & \textbf{0.8553} & \textbf{0.1367} & \textbf{0.0948} & \textbf{24.59} & \textbf{4.30} & 0.6316 & 70.28 & 0.5868  \\
& DiffBIR \cite{lin2024diffbir} & 23.57 & 0.7297 & 0.1923 & 0.1275 & \underline{30.11} & 4.55 & \textbf{0.7804} & \textit{74.51} & \underline{0.6787}  \\  
& OSEDiff \cite{wu2024one} & 23.74 & \textit{0.7980} & \textit{0.1703} & \textit{0.1181} & 41.54 & 4.78 & 0.6874 & 72.51 & 0.6538   \\
& PiSA-SR \cite{sun2024pixel} & \underline{24.02} & \underline{0.8119} & \underline{0.1450} & \underline{0.1161} & 34.11 & \textit{4.35} & \textit{0.7277} & \underline{74.76} & \textit{0.6779}   \\  
& \cellcolor{LightGray!70}\textbf{4KAgent (ExpSR-s4-P)} & \cellcolor{LightGray!70}\textit{23.76} & \cellcolor{LightGray!70}0.7615 & \cellcolor{LightGray!70}0.1776 & \cellcolor{LightGray!70}0.1231 & \cellcolor{LightGray!70}\textit{33.36} & \cellcolor{LightGray!70}\underline{4.32} & \cellcolor{LightGray!70}\underline{0.7678} & \cellcolor{LightGray!70}\textbf{75.08} & \cellcolor{LightGray!70}\textbf{0.6801}   \\ 
\cmidrule{2-11} 
& \textbf{\textbf{\textit{\underline{Agentic System}}}} \\
& AgenticIR \cite{zhu2024intelligent} & \textbf{23.70} & 0.7550  & 0.1862 & \textbf{0.1246} & \textbf{34.01} & 4.38 & 0.7450 & 73.98 & 0.6597   \\
& \cellcolor{LightGray!70}\textbf{4KAgent (GenSR-s4-P)} & \cellcolor{LightGray!70}23.12 & \cellcolor{LightGray!70}\textbf{0.7556} & \cellcolor{LightGray!70}\textbf{0.1834} & \cellcolor{LightGray!70}0.1264 & \cellcolor{LightGray!70}34.58 & \cellcolor{LightGray!70}\textbf{4.23} & \cellcolor{LightGray!70}\textbf{0.7652} & \cellcolor{LightGray!70}\textbf{75.02} & \cellcolor{LightGray!70}\textbf{0.6797}  \\
\bottomrule
\end{tabular}
\end{table*}
Experimental results are shown in~\cref{table:performance_classicsr_set5,table:performance_classicsr_urban100}. 
For the commonly used fidelity metrics PSNR and SSIM in the classical image SR task, 4KAgent with \textbf{ExpSR-s4-F} profile shows competitive performance compared to state-of-the-art fidelity-based methods, ranking among the top three on Set5, B100, Urban100, and Manga109 datasets. For perception-based methods, we focus more on perceptual metrics such as NIQE, CLIPIQA, MUSIQ, and MANIQA. By simply switching the profile from \textbf{ExpSR-s4-F} to \textbf{ExpSR-s4-P}, 4KAgent achieves strong performance among state-of-the-art perception-based methods, ranking among the top two across most metrics on all classical SR benchmarks.
For comparison across agentic systems, 4KAgent outperforms AgenticIR in most metrics on classical SR benchmarks, especially on the Set14 and B100 datasets.

\paragraph{Qualitative Comparison}
\begin{figure*}[!h]
\centering
\includegraphics[width=\textwidth]{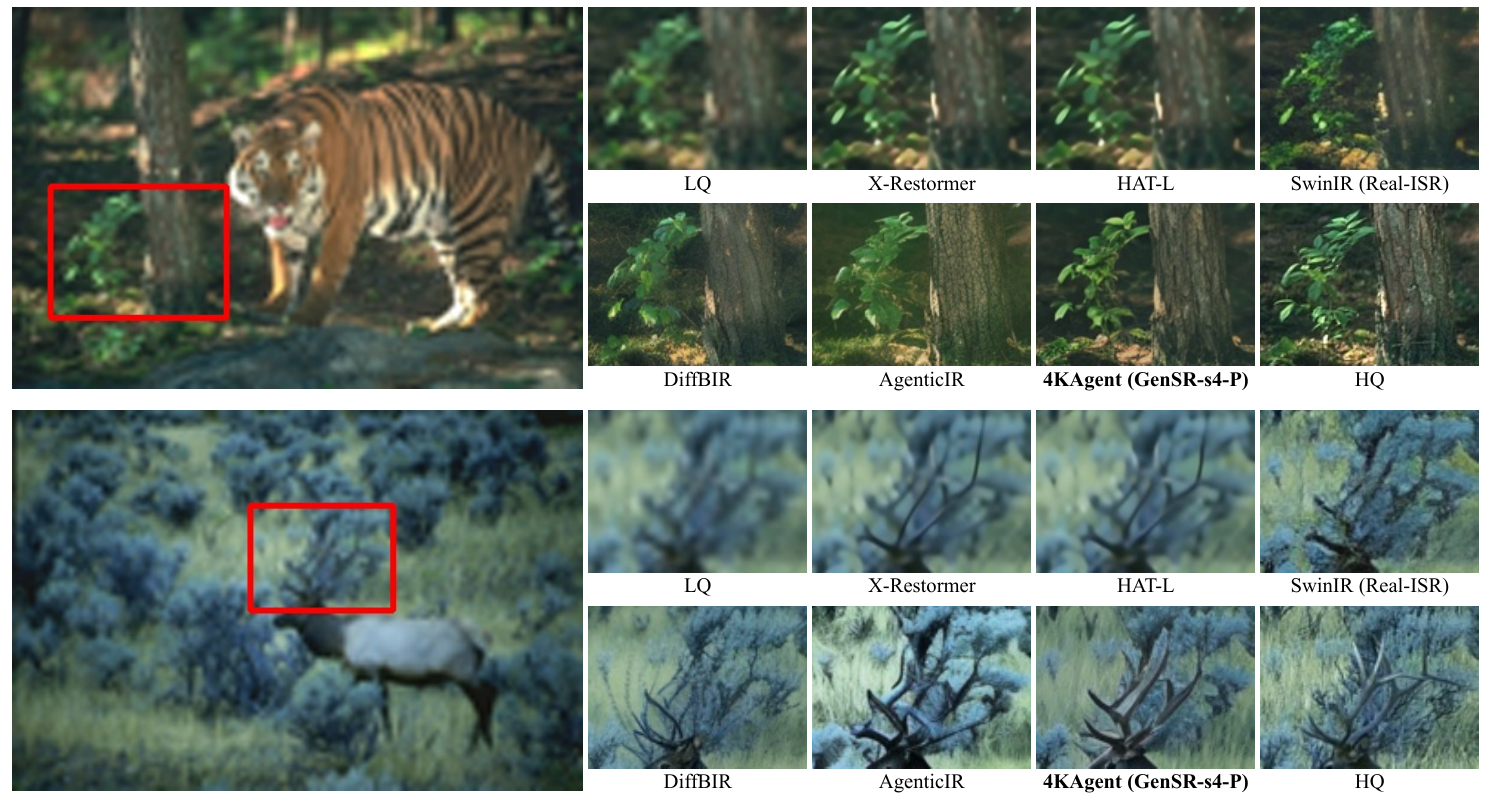}
\caption{Visual comparisons on classical image SR task. (Please zoom in to see details.)}
\vspace{-10pt}
\label{fig:classicsr_compare}
\end{figure*}
For visual comparison, we select two leading fidelity-based methods (X-Restormer, HAT-L), two perception-based methods (SwinIR (Real-ISR), DiffBIR), as well as one agentic system (AgenticIR) as baselines. For 4KAgent, we present images under \textbf{GenSR-s4-P} profile for a comprehensive comparison.
Visual comparisons in~\cref{fig:classicsr_compare} reveal that fidelity-based methods tend to produce overly smooth or blurred details (\eg, HAT-L), even trained with real-world image SR setting (\eg, SwinIR (Real-ISR)), which is visually unpleasant. Diffusion-based method (\eg, DiffBIR) generates rich but unrealistic details. AgenticIR performs well in detail generation but still lacks realism and exhibits noticeable color shifts. 4KAgent delivers both richer and more accurate details than these methods. For instance, it faithfully reproduces the fine stripes on tree bark in the top row and the intricate structure of antlers in the bottom row.

\paragraph{Discussions}
In the context of classical image super-resolution (SR), fidelity-based methods prioritize reconstruction accuracy, measured by PSNR and SSIM, resulting in outputs that often appear overly smooth or blurred. In contrast, perception-based methods optimize for high perceptual quality, reflected in metrics like NIQE, CLIPIQA, MUSIQ, and MANIQA, though often at the expense of fidelity. For example, diffusion-based approaches (\eg, DiffBIR) may hallucinate rich but unrealistic textures. AgenticIR, while capable of generating sharper details, sometimes introduces color shifts or artifacts that undermine visual plausibility. 4KAgent offers configurable flexibility through its profile system, allowing it to operate either as a fidelity-based system (\textbf{ExpSR-s4-F}) or as a perception-based system (\textbf{ExpSR-s4-P}). Quantitatively, 4KAgent delivers competitive PSNR and SSIM scores under the fidelity-based profile, and achieves leading performance in perceptual metrics (NIQE, CLIPIQA, MUSIQ, MANIQA) under the perception-based profile. Qualitatively, 4KAgent consistently produces images with rich, realistic details. The flexibility of 4KAgent allows it to strike a superior balance: it can be easily tuned for maximum visual fidelity or for maximum perceptual appeal without extra training or adaptation, which avoids the common drawbacks of existing SR systems.

\subsection{Real-World Image Super-Resolution}
\label{ssec:real-sr}
\paragraph{Settings.}
In this experiment, we follow prior real-world image super-resolution (SR) methods~\cite{wang2024exploiting, wang2024sinsr, wu2024one, sun2024pixel} and adopt widely used real-world image SR datasets (RealSR \cite{cai2019toward}, DrealSR \cite{wei2020component}) for evaluation. Following the common practice in recent works~\cite{wu2024one, wang2024exploiting, wang2024sinsr}, the real-world data are center-cropped from these datasets with size $128\times128$ for low-quality (LQ) images and $512\times512$ for high-quality (HQ) images. In this experiment, we customize 4KAgent with two different profiles: \textbf{ExpSR-s4-P} and \textbf{GenSR-s4-P}. We compare 4KAgent with state-of-the-art real-world image SR methods, including ResShift \cite{yue2023resshift}, StableSR \cite{wang2024exploiting}, DiffBIR \cite{lin2024diffbir}, PASD \cite{yang2024pixel}, SeeSR \cite{wu2024seesr}, SinSR \cite{wang2024sinsr}, OSEDiff \cite{wu2024one}, and PiSA-SR \cite{sun2024pixel}. In addition, we employ AgenticIR in this experiment for agentic system comparison. The evaluation is conducted using the same metrics as those employed in the Classical Image SR experiment (\S\ref{ssec:classical-sr}).

\paragraph{Quantitative Comparison}
Experiment results on real-world image super-resolution datasets are shown in~\cref{table:performance_realsr}. 
For the real-world image super-resolution task, we concern more about perceptual metrics, such as NIQE, CLIPIQA, MUSIQ, and MANIQA. Real-world image SR methods have achieved promising results on these metrics. AgenticIR, which contains the DiffBIR in its toolbox, outperforms DiffBIR in most perceptual metrics, proving that agentic systems have better potential in solving the real-world SR problem. 4KAgent goes a step further and outperforms AgenticIR in most metrics, achieving better perceptual quality with better fidelity (PSNR and SSIM), regardless of profile setting. In addition, 4KAgent sets a new state-of-the-art performance on perceptual metrics.

% For visual comparison, we select four leading real-world image super-resolution methods—StableSR, DiffBIR, SinSR, and OSEDiff—as well as one agentic system, AgenticIR, as baselines. The visual results are presented in~\cref{fig:realsr_compare}. While previous methods are able to recover rich details from the low-quality (LQ) images, their results often lack realism and fidelity. For example, in the top row, OSEDiff reconstructs clothing that appears more like jackets, whereas the high-quality (HQ) reference image shows down jackets. In contrast, 4KAgent generates sharper and more realistic details. Notably, it accurately recovers the texture of the down jacket in the top row and restores the clarity of the number ‘27’ in the bottom row.

\begin{figure*}[!h]
\centering
\includegraphics[width=\textwidth]{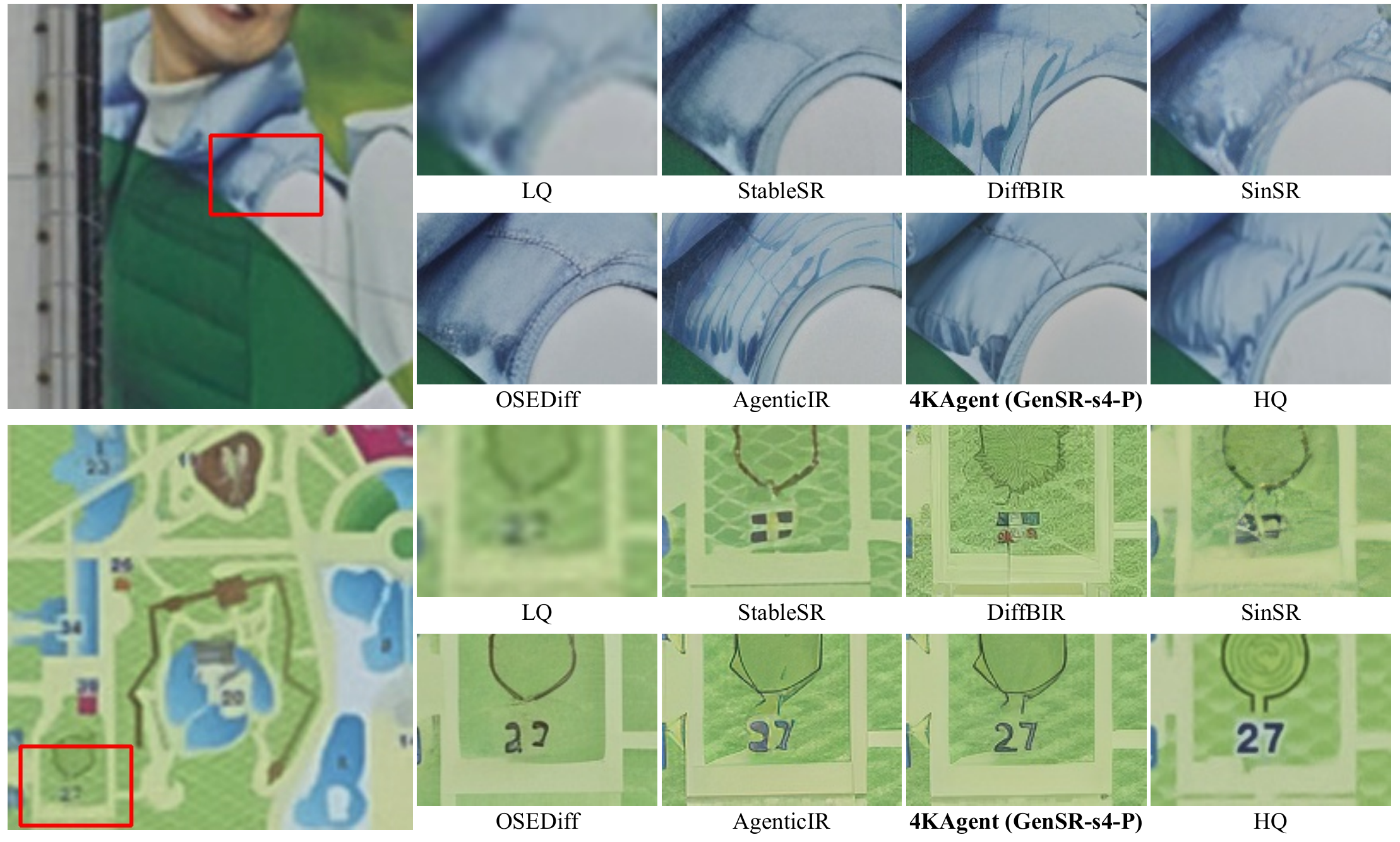}
\caption{Visual comparisons on real-world image SR task. (Please zoom in to see details)}
\label{fig:realsr_compare}
\end{figure*}

\paragraph{Qualitative Comparison}
For visual comparison, we select four leading real-world image super-resolution methods (StableSR, DiffBIR, SinSR, OSEDiff) as well as one agentic system (AgenticIR) as baselines.
The visual results are presented in~\cref{fig:realsr_compare}. 
While previous methods are able to recover rich details from the LQ image, their results often lack realism and fidelity. For example, in the top row, OSEDiff reconstructs clothing that appears more like jackets, whereas the HQ reference image shows down jackets.
4KAgent produces sharper and more realistic details, such as the texture of the down jacket in the top row and the clarity of the number `27' in the bottom row.
\begin{table*}[!h]
\renewcommand{\arraystretch}{1.2}
\centering
\fontsize{6.5pt}{7.5pt}\selectfont
\setlength{\tabcolsep}{4pt}
\caption{Quantitative comparison on real-world image super-resolution benchmarks (RealSR and DrealSR). The top three performances of each metric are marked in \textbf{bold}, \underline{underline}, \textit{italic} respectively.}
\label{table:performance_realsr}
\begin{tabular}{l|l|ccccccccc}
\toprule
{\textbf{Dataset}} & {\textbf{Method}}   & \textbf{PSNR$\uparrow$}&  \textbf{SSIM$\uparrow$} &  \textbf{LPIPS$\downarrow$}  &  \textbf{DISTS$\downarrow$}  &  \textbf{FID$\downarrow$}  &  \textbf{NIQE$\downarrow$}  &  \textbf{CLIPIQA$\uparrow$} & \textbf{MUSIQ$\uparrow$}  &  \textbf{MANIQA$\uparrow$}      \\
\midrule
\multirow{11}{*}{RealSR}
& ResShift \cite{yue2023resshift} & \textbf{26.31} & \underline{0.7411} & 0.3489 & 0.2498 & 142.81 & 7.27 & 0.5450 & 58.10 & 0.5305 \\
& StableSR \cite{wang2024exploiting} & 24.69 & 0.7052 & 0.3091 & \textit{0.2167} & 127.20 & 5.76 & 0.6195 & 65.42 & 0.6211 \\
& DiffBIR \cite{lin2024diffbir} & 24.88 & 0.6673 & 0.3567 & 0.2290 & 124.56 & 5.63 & 0.6412 & 64.66 & 0.6231 \\
& PASD \cite{yang2024pixel} & 25.22 & 0.6809 & 0.3392 & 0.2259 & \textbf{123.08} & \textit{5.18} & 0.6502 & 68.74 & 0.6461 \\
& SeeSR \cite{wu2024seesr} & 25.33 & 0.7273 & \textit{0.2985} & 0.2213 & 125.66 & 5.38 & 0.6594 & 69.37 & 0.6439 \\
& SinSR \cite{wang2024sinsr} & \underline{26.30} & \textit{0.7354} & 0.3212 & 0.2346 & 137.05 & 6.31 & 0.6204 & 60.41 & 0.5389 \\
& OSEDiff \cite{wu2024one} & 25.15 & 0.7341 & \underline{0.2921} & \underline{0.2128} & \underline{123.50} & 5.65 & \textit{0.6693} & 69.09 & 0.6339 \\
& PiSA-SR \cite{sun2024pixel} & \textit{25.50} & \textbf{0.7417} & \textbf{0.2672} & \textbf{0.2044} & \textit{124.09} & 5.50 & \underline{0.6702} & \textit{70.15} & \textit{0.6560} \\
& AgenticIR \cite{zhu2024intelligent} & 22.45 & 0.6447 & 0.3745 & 0.2503 & 140.38 & 5.81 & 0.6506 & 65.87 & 0.6210 \\
& \cellcolor{LightGray!70}\textbf{4KAgent (ExpSR-s4-P)}  & \cellcolor{LightGray!70}24.60 & \cellcolor{LightGray!70}0.6839 & \cellcolor{LightGray!70}0.3253 & \cellcolor{LightGray!70}0.2292 & \cellcolor{LightGray!70}127.64 & \cellcolor{LightGray!70}\underline{5.09} & \cellcolor{LightGray!70}\textbf{0.7078} & \cellcolor{LightGray!70}\underline{70.97} & \cellcolor{LightGray!70}\textbf{0.6602} \\
& \cellcolor{LightGray!70}\textbf{4KAgent (GenSR-s4-P)} & \cellcolor{LightGray!70}22.55 & \cellcolor{LightGray!70}0.6557 & \cellcolor{LightGray!70}0.3509 & \cellcolor{LightGray!70}0.2468 & \cellcolor{LightGray!70}134.63 & \cellcolor{LightGray!70}\textbf{4.78} & \cellcolor{LightGray!70}0.6666 & \cellcolor{LightGray!70}\textbf{71.77} & \cellcolor{LightGray!70}\underline{0.6564} \\
\midrule
\multirow{11}{*}{DrealSR}
& ResShift \cite{yue2023resshift} & \textbf{28.45} & 0.7632 & 0.4073 & 0.2700 & 175.92 & 8.28 & 0.5259 & 49.86 & 0.4573 \\
& StableSR \cite{wang2024exploiting} & 28.04 & 0.7460 & 0.3354 & \textit{0.2287} & \textit{147.03} & 6.51 & 0.6171 & 58.50 & 0.5602 \\
& DiffBIR \cite{lin2024diffbir} & 26.84 & 0.6660 & 0.4446 & 0.2706 & 167.38 & 6.02 & 0.6292 & 60.68 & 0.5902 \\
& PASD \cite{yang2024pixel} & 27.48 & 0.7051 & 0.3854 & 0.2535 & 157.36 & \textit{5.57} & 0.6714 & 64.55 & 0.6130 \\
& SeeSR \cite{wu2024seesr} & 28.26 & \textit{0.7698} & \textit{0.3197} & 0.2306 & 149.86 & 6.52 & 0.6672 & 64.84 & 0.6026 \\
& SinSR \cite{wang2024sinsr} & \underline{28.41} & 0.7495 & 0.3741 & 0.2488 & 177.05 & 7.02 & 0.6367 & 55.34 & 0.4898 \\
& OSEDiff \cite{wu2024one} & 27.92 & \textbf{0.7835} & \underline{0.2968} & \textbf{0.2165} & \underline{135.29} & 6.49 & 0.6963 & 64.65 & 0.5899 \\
& PiSA-SR \cite{sun2024pixel} & \textit{28.31} & \underline{0.7804} & \textbf{0.2960} & \underline{0.2169} & \textbf{130.61} & 6.20 & \textit{0.6970} & \textit{66.11} & \textit{0.6156} \\
& AgenticIR \cite{zhu2024intelligent} & 23.06 & 0.6145 & 0.4775 & 0.2973 & 182.02 & 6.11 & 0.6542 & 63.59 & 0.5927 \\
& \cellcolor{LightGray!70}\textbf{4KAgent (ExpSR-s4-P)}  & \cellcolor{LightGray!70}26.00 & \cellcolor{LightGray!70}0.6535 & \cellcolor{LightGray!70}0.4257 & \cellcolor{LightGray!70}0.2717 & \cellcolor{LightGray!70}170.19 & \cellcolor{LightGray!70}\underline{5.51} & \cellcolor{LightGray!70}\textbf{0.7167} & \cellcolor{LightGray!70}\underline{67.72} & \cellcolor{LightGray!70}\textbf{0.6397} \\
& \cellcolor{LightGray!70}\textbf{4KAgent (GenSR-s4-P)} & \cellcolor{LightGray!70}23.11 & \cellcolor{LightGray!70}0.6126 & \cellcolor{LightGray!70}0.4579 & \cellcolor{LightGray!70}0.2866 & \cellcolor{LightGray!70}178.36 & \cellcolor{LightGray!70}\textbf{4.65} & \cellcolor{LightGray!70}\underline{0.7092} & \cellcolor{LightGray!70}\textbf{69.30} & \cellcolor{LightGray!70}\underline{0.6219}  \\
\bottomrule
\end{tabular}
\end{table*}

\paragraph{Discussions}
The real-world image super-resolution task is more challenging than the classical image super-resolution task as it contains more complex distortions than the synthetic downsample, which can also be seen from the comparison of the LQ image and HQ image in the dataset.
Under this challenging setting, agentic systems prove their advantage by analyzing the distortion and restoring the image properly.
4KAgent further proves its superiority by consistently outperforming AgenticIR in most quantitative metrics. In particular, 4KAgent sets a new state-of-the-art performance for no-reference perceptual metrics, demonstrating that its design effectively elevates perceived realism. Qualitatively, these gains translate into visibly sharper and more believable details. By dynamically leveraging multiple SR experts and selecting the optimal result, 4KAgent shows its superiority in the challenging real-world image super-resolution task.

\subsection{Multiple-Degradation Image Restoration}
\label{ssec:multiple-sr}
\paragraph{Settings.}
We follow the experimental protocol of AgenticIR, evaluating 4KAgent on the Group A, B, and C test sets, which together contain 1,440 LQ images generated by applying 16 combinations of degradations to images from the MiO100 dataset \cite{kong2024preliminary}. In this experiment, we configure 4KAgent with the \textbf{GenMIR-P} profile. We compare 4KAgent with several leading all-in-one models, including AirNet \cite{li2022all}, PromptIR \cite{potlapalli2023promptir}, MiOIR \cite{kong2024towards}, DA-CLIP \cite{luo2023controlling}, InstructIR \cite{conde2024instructir}, AutoDIR \cite{jiang2024autodir}, as well as agentic systems: AgenticIR \cite{zhu2024intelligent} and MAIR \cite{jiang2025multi}. For evaluation, we adopt a comprehensive set of metrics, including PSNR, SSIM, LPIPS, MANIQA, CLIPIQA, and MUSIQ.

\paragraph{Quantitative Comparison}
Experimental results are shown in~\cref{table:performance_mir}.
In the multiple-degradation image restoration (IR) task, agentic systems once again prove their superiority, outperforming all-in-one methods in all metrics. Among the agentic systems, 4KAgent performs the best, achieving a new state-of-the-art performance on PSNR, MANIQA, CLIPIQA, and MUSIQ. Specifically, for no-reference perceptual metrics (MANIQA, CLIPIQA, MUSIQ), 4KAgent outperforms all compared methods by a noticeable margin (\eg, 4.2 lead of MUSIQ on Group C). For SSIM and LPIPS metrics, 4KAgent remains competitive, ranking among the top two on Group A and Group C subsets.
\begin{table*}[!h]
\renewcommand{\arraystretch}{1.2}
\centering
\fontsize{6.5pt}{7.5pt}\selectfont
\setlength{\tabcolsep}{6pt}
\caption{Quantitative comparison of multiple-degradation image restoration tasks on three subsets (Group A, B, and C) from the MiO100 dataset. The top three performances of each metric are marked in \textbf{bold}, \underline{underline}, \textit{italic} respectively.}
\label{table:performance_mir}
\begin{tabular}{l|l|cccccc}
\toprule
{\textbf{Degradations}} & {\textbf{Method}}   & \textbf{PSNR$\uparrow$}&  \textbf{SSIM$\uparrow$} &  \textbf{LPIPS$\downarrow$}   &  \textbf{MANIQA$\uparrow$}  &  \textbf{CLIPIQA$\uparrow$}  &  \textbf{MUSIQ$\uparrow$}   \\ 
\midrule
\multirow{9}{*}{Group A}
& AirNet \cite{li2022all} & 19.13 & 0.6019 & 0.4283 & 0.2581 & 0.3930 & 42.46  \\  
& PromptIR \cite{potlapalli2023promptir} & 20.06 & 0.6088 & 0.4127 & 0.2633 & 0.4013 & 42.62   \\
& MiOIR \cite{kong2024towards} & 20.84 & 0.6558 & 0.3715 & 0.2451 & 0.3933 & 47.82   \\
& DA-CLIP \cite{luo2023controlling} & 19.58 & 0.6032  & 0.4266 & 0.2418 & 0.4139 & 42.51   \\  
& InstructIR \cite{conde2024instructir} & 18.03 & 0.5751 & 0.4429 & 0.2660 & 0.3528 & 45.77   \\ 
& AutoDIR \cite{jiang2024autodir} & 19.64 & 0.6286 & 0.3967 & 0.2500 & 0.3767 & 47.01 \\ 
& AgenticIR \cite{zhu2024intelligent} & \underline{21.04} & \textbf{0.6818} & \textit{0.3148} & \textit{0.3071} & \textit{0.4474} & \textit{56.88}   \\ 
& MAIR \cite{jiang2025multi} & \textit{21.02} & \textit{0.6715} & \textbf{0.2963} & \underline{0.3330} & \underline{0.4751} & \underline{59.19} \\ 
% \rowcolor{LightGray!70}
& \cellcolor{LightGray!70}\textbf{4KAgent (GenMIR-P)} & \cellcolor{LightGray!70}\textbf{21.48}  & \cellcolor{LightGray!70}\underline{0.6720}  & \cellcolor{LightGray!70}\underline{0.3019}  & \cellcolor{LightGray!70}\textbf{0.3748} & \cellcolor{LightGray!70}\textbf{0.5544}  & \cellcolor{LightGray!70}\textbf{63.19}  \\
\midrule
\multirow{9}{*}{Group B}
& AirNet \cite{li2022all} & 19.31 & 0.6567 & 0.3670 & 0.2882 & 0.4274 & 47.88  \\  
& PromptIR \cite{potlapalli2023promptir} & 20.47 & 0.6704 & 0.3370 & 0.2893 & 0.4289 & 48.10   \\
& MiOIR \cite{kong2024towards} & \textit{20.56} & \textit{0.6905} & 0.3243 & 0.2638 & 0.4330 & 51.87   \\
& DA-CLIP \cite{luo2023controlling} & 18.56 & 0.5946  & 0.4405 & 0.2435 & 0.4154 & 43.70   \\  
& InstructIR \cite{conde2024instructir} & 18.34 & 0.6235 & 0.4072 & 0.3022 & 0.3790 & 50.94   \\ 
& AutoDIR \cite{jiang2024autodir} & 19.90 & 0.6643 & 0.3542 & 0.2534 & 0.3986 & 49.64 \\ 
& AgenticIR \cite{zhu2024intelligent} & 20.55 & \textbf{0.7009} & \textit{0.3072} & \textit{0.3204} & \textit{0.4648} & \textit{57.57}   \\ 
& MAIR \cite{jiang2025multi} & \underline{20.92} & \underline{0.7004} & \textbf{0.2788} & \underline{0.3544} & \underline{0.5084} & \underline{60.98} \\ 
& \cellcolor{LightGray!70}\textbf{4KAgent (GenMIR-P)} & \cellcolor{LightGray!70}\textbf{20.95}  & \cellcolor{LightGray!70}0.6727  & \cellcolor{LightGray!70}\underline{0.3017}  & \cellcolor{LightGray!70}\textbf{0.3734} & \cellcolor{LightGray!70}\textbf{0.5505}  & \cellcolor{LightGray!70}\textbf{62.69}   \\ 
\midrule
\multirow{9}{*}{Group C}
& AirNet \cite{li2022all} & 17.95 & 0.5145 & 0.5782 & 0.1854 & 0.3113 & 30.12  \\  
& PromptIR \cite{potlapalli2023promptir} & 18.51 & 0.5166 & 0.5756 & 0.1906 & 0.3104 & 29.71   \\
& MiOIR \cite{kong2024towards} & 15.63 & 0.4896 & 0.5376 & 0.1717 & 0.2891 & 37.95   \\
& DA-CLIP \cite{luo2023controlling} & 18.53 & 0.5320  & 0.5335 & 0.1916 & 0.3476 & 33.87   \\  
& InstructIR \cite{conde2024instructir} & 17.09 & 0.5135 & 0.5582 & 0.1732 & 0.2537 & 33.69   \\ 
& AutoDIR \cite{jiang2024autodir} & 18.61 & 0.5443 & 0.5019 & 0.2045 & 0.2939 & 37.86 \\ 
& AgenticIR \cite{zhu2024intelligent} & \textit{18.82} & \textit{0.5474} & \textit{0.4493} & \textit{0.2698} & \textit{0.3948} & \textit{48.68}   \\ 
& MAIR \cite{jiang2025multi} & \underline{19.42} & \underline{0.5544} & \textbf{0.4142} & \underline{0.2798} & \underline{0.4239} & \underline{51.36} \\ 
& \cellcolor{LightGray!70}\textbf{4KAgent (GenMIR-P)} & \cellcolor{LightGray!70}\textbf{19.77}  & \cellcolor{LightGray!70}\textbf{0.5629} & \cellcolor{LightGray!70}\underline{0.4271} & \cellcolor{LightGray!70}\textbf{0.3545}  & \cellcolor{LightGray!70}\textbf{0.5233}  & \cellcolor{LightGray!70}\textbf{55.56}  \\
\bottomrule
\end{tabular}
\end{table*}

\begin{figure*}[!h]
\centering
\includegraphics[width=\textwidth]{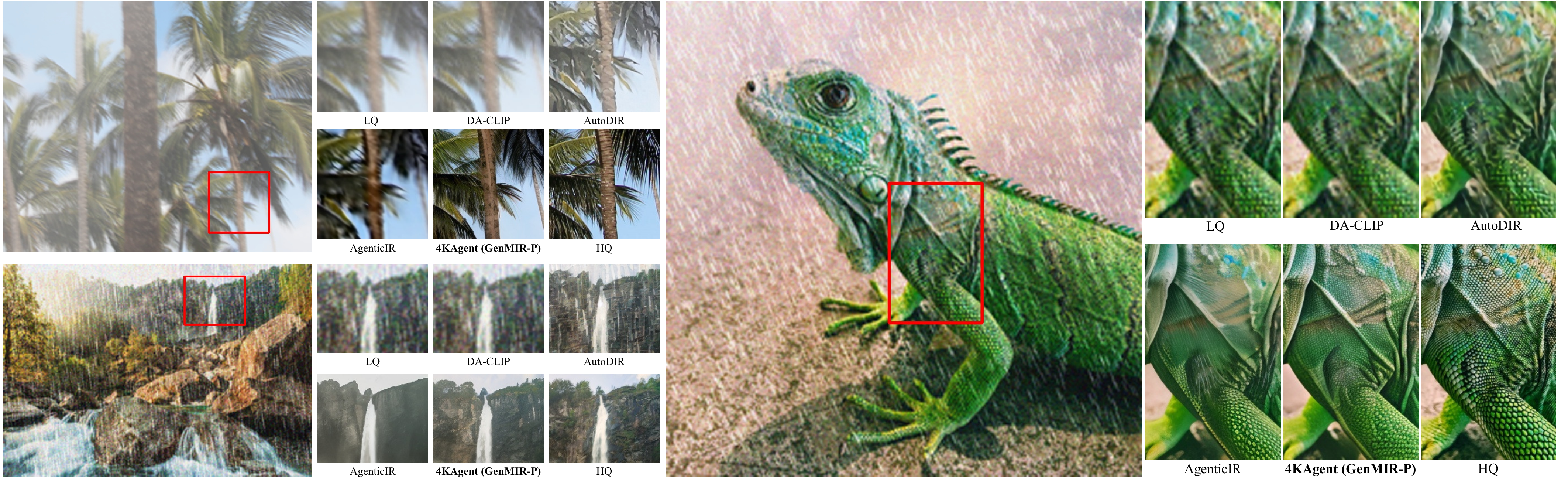}
\caption{Visual comparisons on multiple-degradation IR task. (Please zoom in to see details)}
\label{fig:mio_compare}
\vspace{-5pt}
\end{figure*}

\paragraph{Qualitative Comparison}
For visual comparison, we select two leading all-in-one methods (DA-CLIP, AutoDIR) as well as an agentic system (AgenticIR) as baselines.
Visual comparisons are shown in~\cref{fig:mio_compare}, 
all-in-one methods performs limited under this setting, especially when restoring complex distortions, such as rain drops. AgenticIR achieves promising results, proving the potential of agentic systems in dealing with complex distortion tasks. 4KAgent goes a step further, generating images with high-grained details and more consistent with the high-quality (HQ) reference image. For instance, the natural stripes on the tree trunks and the fine leaf textures, the intricate waterfall ripples and mountain contours on the left, and the skin of the lizard on the right. These results demonstrate the superiority of 4KAgent under complex distortions.

\paragraph{Discussions}
Concluding from experiment results, agentic systems have shown their superiority in the multi-degradation image restoration tasks, where each low-quality (LQ) image is affected by 2 $\sim$ 3 types of distortions. Under this challenging setting, 4KAgent exhibits a clear advantage in handling complex, multi-degraded inputs, outperforming both conventional all-in-one methods and previous agentic systems. Quantitatively, 4KAgent achieves state-of-the-art results across multiple metrics, including PSNR, MANIQA, CLIPIQA, and MUSIQ, highlighting its strong capability in enhancing both fidelity and perceptual quality. Qualitatively, this metric superiority translates into more faithful restoration of fine-grained patterns and textures, even under severe and heterogeneous distortions.

\subsection{Face Restoration}
\label{ssec:face-sr}
\paragraph{Settings.}
In this section, we evaluate 4KAgent on the real-world face restoration benchmark, the WebPhoto-test~\cite{wang2021towards} dataset, which contains 407 low-quality face images collected from the Internet. As face restoration pipeline is a module subsequent to super-resolution task in 4KAgent, we first downsample the images by a factor of 4 to generate low-quality (LQ) images. In this experiment, we configure 4KAgent with the \textbf{GenSRFR-s4-P} profile.

We compare 4KAgent with state-of-the-art face restoration methods, including CodeFormer~\cite{zhou2022towards}, GFPGAN~\cite{wang2021towards}, and DifFace~\cite{yue2024difface}, as well as an agentic system AgenticIR~\cite{zhu2024intelligent}. For face restoration methods, we set the scaling factor to 4. As there are no high-quality (HQ) references, we evaluate performance with four no-reference perceptual metrics (NIQE, CLIPIQA, MUSIQ, and MANIQA-pipal) and two advanced face-specific IQA metrics (CLIB-FIQA \cite{ou2024clib} and DSL-FIQA \cite{chen2024dsl}). 

\paragraph{Quantitative Comparison}
Experimental results are shown in~\cref{table:performance_face}. 
AgenticIR performs worse than previous face restoration methods in terms of general perceptual metrics and face IQA metrics. 4KAgent outperforms AgenticIR on every metric by a clear margin (\eg, 19.67 lead on MUSIQ).
Moreover, 4KAgent achieves the best scores on general no-reference perceptual metrics and delivers competitive results on face IQA metrics, ranking second in both CLIB-FIQA and DSL-FIQA.
\begin{table*}[!h]
\renewcommand{\arraystretch}{1.2}
\centering
\fontsize{6.5pt}{7.5pt}\selectfont
\setlength{\tabcolsep}{4pt}
\caption{Quantitative comparison on face restoration benchmark (WebPhoto-Test). The top three performances of each metric are marked in \textbf{bold}, \underline{underline}, \textit{italic} respectively.}
\label{table:performance_face}
\begin{tabular}{l|l|cccccc}
\toprule
{\textbf{Dataset}} & {\textbf{Method}}   &   \textbf{NIQE$\downarrow$}  &  \textbf{CLIPIQA$\uparrow$} & \textbf{MUSIQ$\uparrow$}  &  \textbf{MANIQA$\uparrow$}  &  \textbf{CLIB-FIQA$\uparrow$}  & \textbf{DSL-FIQA$\uparrow$}     \\
\midrule
\multirow{5}{*}{WebPhoto-Test}
& GFPGAN \cite{wang2021towards} & 5.12   & \textit{0.6792} & \underline{74.21} & \textit{0.6379} & \textit{0.6590} & \textbf{0.7732} \\
& CodeFormer \cite{zhou2022towards} & \textit{4.58}   & \underline{0.6884} & \textit{73.87} & \underline{0.6415} & \textbf{0.6840} & \textit{0.7435} \\
& DifFace \cite{yue2024difface} & \underline{4.20}   & 0.5831 & 65.31 & 0.5891 & 0.6511 & 0.6189 \\
& AgenticIR \cite{zhu2024intelligent} & 6.85  & 0.5731 & 56.25 & 0.5465 & 0.5978 & 0.5289 \\
& \cellcolor{LightGray!70}\textbf{4KAgent (GenSRFR-s4-P)} & \cellcolor{LightGray!70}\textbf{4.15}  & \cellcolor{LightGray!70}\textbf{0.7077} & \cellcolor{LightGray!70}\textbf{75.92} & \cellcolor{LightGray!70}\textbf{0.6576} & \cellcolor{LightGray!70}\underline{0.6671} & \cellcolor{LightGray!70}\underline{0.7683} \\
\bottomrule
\end{tabular}
\vspace{-6pt}
\end{table*}

\begin{figure*}[!h]
\centering
\includegraphics[width=\textwidth]{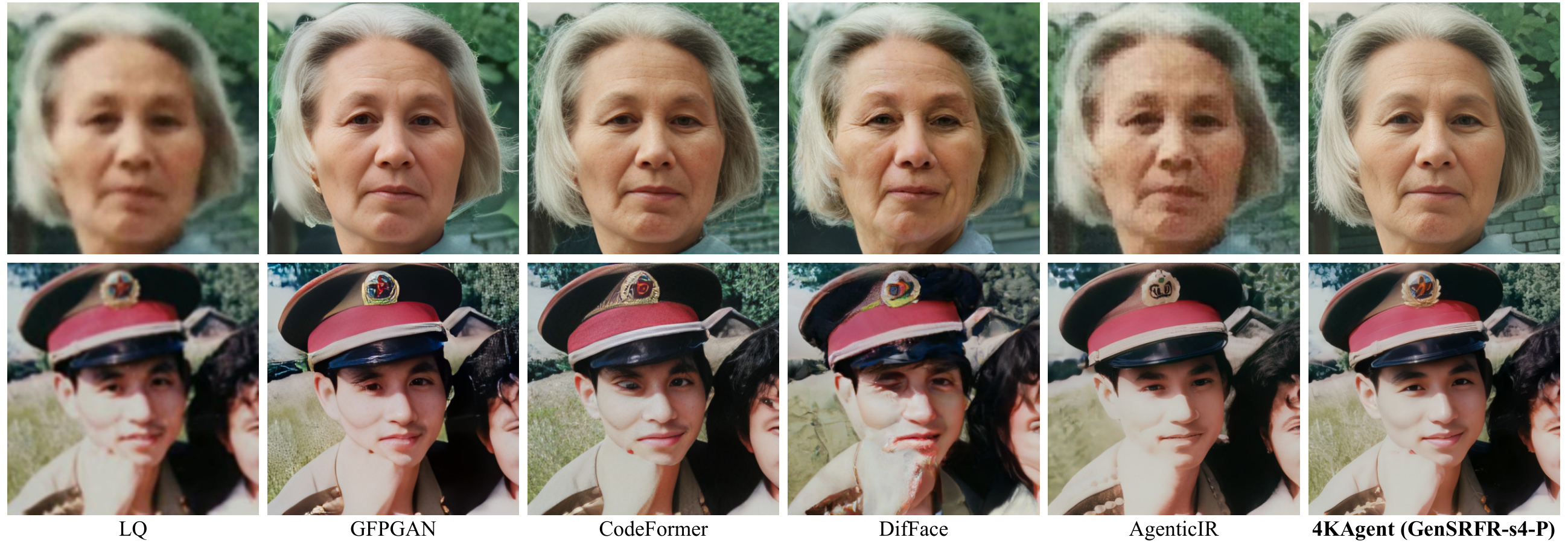}
\caption{Visual comparisons on face restoration task. (Please zoom in to see details)}
\label{fig:face_compare}
\vspace{-10pt}
\end{figure*}

\paragraph{Qualitative Comparison}
Visual comparisons are shown in~\cref{fig:face_compare}. 
Compared with other methods, 4KAgent demonstrates a clear advantage in restoring realistic facial details, such as fine hair strands and natural skin textures. Moreover, it achieves superior restoration performance in non-facial regions, such as the wall and leaves in the first row and the logo on the hat in the second row. 
By consistently delivering high-quality restoration in both facial and non-facial areas, 4KAgent produces more visually pleasing and perceptually balanced results overall.

\paragraph{Discussions}
In the face restoration scenario, 4KAgent effectively addresses both facial and contextual degradations. Quantitatively, 4KAgent achieves the best performance on general no-reference perceptual metrics and delivers competitive scores on face IQA metrics, demonstrating the superiority of its system design and face Q-MoE policy. Qualitatively, this translates into more natural and richly detailed facial features, such as individual hair strands and realistic skin texture, while also enhancing background elements, producing overall more visually pleasing outputs. 
Among agentic systems, AgenticIR applies a uniform processing pipeline without a dedicated face restoration module, which limits its performance on face restoration tasks. Benefit from the face restoration pipeline and profile design, 4KAgent can be tailored as a face restoration expert, achieving superior results. We further present how these two designs enhance the face restoration ability of 4KAgent in the ablation study.

\section{Experiment Part II: 16$\times$ Natural Image Super-Resolution}
\label{sec:experiment-ii}

Experiment Part I (\S\ref{sec:experiment-i}) demonstrates the flexibility and superiority of 4KAgent on general image restoration and super-resolution tasks. In this section, we evaluate 4KAgent on more challenging restoration tasks. First, we assess its performance on a 16$\times$ real-world image super-resolution task. Next, we introduce a new image restoration dataset, \textbf{DIV4K-50}, designed to restore 256$\times$256 resolution images with multiple distortions and upscale them to 4096$\times$4096 resolution, and we evaluate our 4KAgent on this new benchmark.

\subsection{Large Scale Factor (16$\times$) Image Super-Resolution}
\label{ssec:16x-sr}
\paragraph{Settings.}
In this experiment, we explore a challenging setting, 16$\times$ real-world image upscaling. For the dataset in this experiment, we adopt the RealSRSet dataset~\cite{zhang2021designing}, which consists of 20 real-world low-quality images for real-world image super-resolution task. We configure 4KAgent with the \textbf{Gen4K-P} profile for this experiment. Based on the original resolution of images in this dataset, 4KAgent will upscale each image with a scale factor of 16.

We compare 4KAgent against previous methods, including HAT-L~\cite{chen2023hat}, DiffBIR~\cite{lin2024diffbir}, OSEDiff~\cite{wu2024one}, and PiSA-SR~\cite{sun2024pixel}, under two distinct settings: (1) 4$\times$$\rightarrow$ 4$\times$: first upscale the low-quality image by a scale factor of 4, then upscale the upscaled images to obtain the 16$\times$ upscaled images. (2) 16$\times$: directly upscale the low-quality image by a scale factor of 16. We also extend AgenticIR \cite{zhu2024intelligent} for a larger scale-factor (16$\times$) image super-resolution for agentic system comparison. As there are no corresponding high-quality (HQ) reference images in the RealSRSet dataset, we evaluate the result images on four no-reference perceptual metrics: NIQE, CLIPIQA, MUSIQ, and MANIQA-pipal.

\paragraph{Quantitative Comparison}
Experiment results are shown in~\cref{table:performance_realsrset}. As the largest scale factor of the pre-trained model in HAT-L is 4, we apply the 4$\times$$\rightarrow$ 4$\times$ setting for HAT-L for 16$\times$ upscaling.
Fidelity-based method struggles to deliver satisfactory performance in perceptual metrics under this setting. Recent perceptual-based real-world image super-resolution methods perform well in these metrics, even with the 16$\times$ setting. For example, DiffBIR with 16$\times$ setting achieves the best NIQE and MANIQA scores.
Among agentic systems, 4KAgent outperforms AgenticIR on every metric by a clear margin (\eg, 6.13 lead on MUSIQ).
In addition, 4KAgent achieves the best performance on MUSIQ and the second-best performance on NIQE. For CLIPIQA and MANIQA metrics, 4KAgent also delivers competitive performance, ranking among the top three across all methods.
\begin{table*}[!h]
\renewcommand{\arraystretch}{1.2}
\centering
\fontsize{6.5pt}{7.5pt}\selectfont
\setlength{\tabcolsep}{6pt}
\caption{Quantitative comparison on RealSRSet dataset under 16$\times$ upscaling. The top three performances of each metric are marked in \textbf{bold}, \underline{underline}, \textit{italic} respectively.}
\label{table:performance_realsrset}
\begin{tabular}{l|l|cccc}
\toprule
{\textbf{Dataset}} & {\textbf{Method}}   &   \textbf{NIQE$\downarrow$}  &  \textbf{CLIPIQA$\uparrow$} & \textbf{MUSIQ$\uparrow$}  &  \textbf{MANIQA$\uparrow$}        \\
\midrule
\multirow{9}{*}{RealSRSet}
& HAT-L \cite{chen2023hat} (4$\times$$\rightarrow$ 4$\times$) & 10.59 & 0.3885 & 25.06 & 0.3060 \\
& DiffBIR \cite{lin2024diffbir}  (4$\times$$\rightarrow$ 4$\times$) & \textit{3.63}  & \underline{0.7867} & 44.86 & \underline{0.6076} \\
& DiffBIR \cite{lin2024diffbir} (16$\times$) & \textbf{2.80}  & 0.7583 & 47.54 & \textbf{0.6099} \\
& OSEDiff \cite{wu2024one} (4$\times$$\rightarrow$ 4$\times$) & 5.40  & 0.7665 & \underline{48.42} & 0.5362  \\
& OSEDiff \cite{wu2024one} (16$\times$) & 4.66  & 0.6483 & 35.33 & 0.4581  \\
& PiSA-SR \cite{sun2024pixel} (4$\times$$\rightarrow$ 4$\times$) & 5.70  & \textbf{0.7883} & \textit{48.20} & 0.5464  \\
& PiSA-SR \cite{sun2024pixel} (16$\times$) & 4.88  & 0.6384 & 35.90 & 0.4128  \\
& AgenticIR \cite{zhu2024intelligent} & 4.86  & 0.6775 & 44.71 & 0.5236 \\
& \cellcolor{LightGray!70}\textbf{4KAgent (Gen4K-P)} & \cellcolor{LightGray!70}\underline{3.53} & \cellcolor{LightGray!70}\textit{0.7794} & \cellcolor{LightGray!70}\textbf{50.84} & \cellcolor{LightGray!70}\textit{0.5913}  \\
\bottomrule
\end{tabular}
\end{table*}

\begin{figure*}[!h]
\centering
\includegraphics[width=\textwidth]{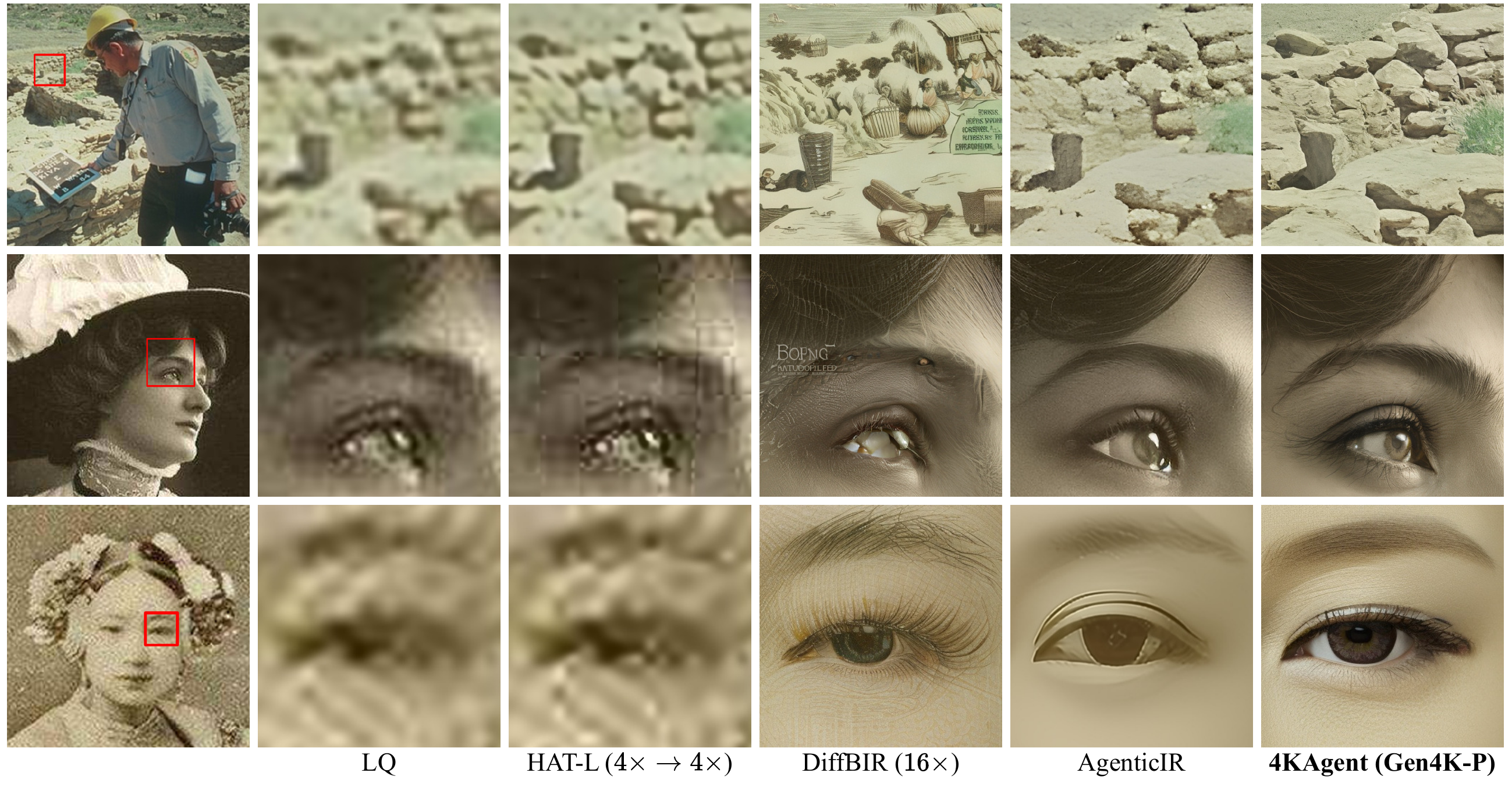}
\vspace{-15pt}
\caption{Visual comparisons on RealSRSet dataset (16$\times$ upscaling). (Please zoom in to see details)}
\label{fig:realsrset_compare}
\vspace{-5pt}
\end{figure*}

\paragraph{Qualitative Comparison}
For visual comparison, we select three representative methods to benchmark against 4KAgent: 
(1) HAT-L (4$\times$$\rightarrow$ 4$\times$): As a representative fidelity-based method, we investigate its performance under a large-scale upscaling setting. (2) DiffBIR (16$\times$): As shown in~\cref{table:performance_realsrset}, DiffBIR with 16$\times$ setting achieves the best performance on NIQE and MANIQA. Therefore, we include it to assess its visual quality. (3) AgenticIR: Selected for agentic system comparison. 
Visual comparisons are shown in~\cref{fig:realsrset_compare}. 

HAT-L (4$\times$$\rightarrow$ 4$\times$) shows limited enhancement over the low-quality input, leading to notably blurred textures.
DiffBIR (16$\times$) produces visually rich but often unrealistic hallucinations, and in some cases even alters the semantic content of the scene (\eg, the first row), which is visually unappealing.
AgenticIR generates visually plausible results but lacks sufficient sharpness and fine-grained details.
4KAgent generates high-grained and realistic details: the rock and grass textures in the first row, and the hair strands, eyebrow patterns, and naturally expressive eyes in the second and third rows are all much more faithfully restored with high-grained details.

\paragraph{Discussions}
In the challenging 16$\times$ upscaling scenario, 4KAgent delivers competitive quantitative results alongside high-grained and realistic qualitative results, compared to other methods.
In addition, traditional fidelity-oriented methods such as HAT-L struggle to recover fine details and instead produce overly smoothed and blurred results. This highlights the limitation of fidelity-driven pipelines under extreme magnification levels. Therefore, for such high-scale upscaling tasks, it is essential to configure 4KAgent with a perception-oriented profile (\eg, setting \textbf{Restore Option} to \textbf{Perception} in the profile module) to better prioritize realistic texture synthesis. As image resolution approaches 4K and beyond, existing no-reference perceptual metrics may become misaligned with human judgment of visual quality. This discrepancy underscores the need for developing new no-reference perceptual metrics specifically designed for ultra-high-resolution images.

\subsection{Joint Restoration \& 4K Upscaling}
\label{ssec:div4k-sr}
\paragraph{Settings.}
In this section, we bring 4KAgent to the most challenging setting: Joint multiple image restoration and 4K upscaling. 
As there are no previous methods and dataset target at this setting, we construct a new evaluation dataset, \textbf{DIV4K-50}, constructed from the Aesthetic-4K dataset \cite{zhang2025diffusion4k} to rigorously test end-to-end restoration and ultra-high-scale SR. Specifically, we select 50 images from the Aesthetic-4K dataset based on its content, then center-crop each image to 4096$\times$4096 as the high-quality (HQ) ground truth. Then we downsample HQ images to 256$\times$256 and randomly apply combinations of defocus blur, motion blur, additive Gaussian noise, and JPEG compression to generate the corresponding low-quality (LQ) images. In this experiment, we also configure 4KAgent with the \textbf{Gen4K-P} profile. Comparing methods and experimental settings are the same as in~\cref{ssec:16x-sr}.

\paragraph{Quantitative Comparison}
Quantitative comparisons are shown in~\cref{table:performance_div4k}. 
Similar to the experiment result in~\cref{ssec:16x-sr}, real-world image super-resolution methods perform competitively on perceptual metrics under this challenging setting. 
For example, DiffBIR achieves the best score on NIQE and CLIPIQA metrics. For agentic systems, 4KAgent outperforms AgenticIR on every metric. Additionally, 4KAgent achieves the best performance on MUSIQ and MANIQA metrics, and the second-best performance on NIQE and CLIPIQA metrics.
\begin{table*}[!h]
\renewcommand{\arraystretch}{1.2}
\centering
\fontsize{6.5pt}{7.5pt}\selectfont
\setlength{\tabcolsep}{6pt}
\caption{Quantitative comparison on DIV4K-50 dataset. The top three performances of each metric are marked in \textbf{bold}, \underline{underline}, \textit{italic} respectively.}
\label{table:performance_div4k}
\begin{tabular}{l|l|cccc}
\toprule
{\textbf{Dataset}} & {\textbf{Method}}   &   \textbf{NIQE$\downarrow$}  &  \textbf{CLIPIQA$\uparrow$} & \textbf{MUSIQ$\uparrow$}  &  \textbf{MANIQA$\uparrow$}        \\
\midrule
\multirow{9}{*}{DIV4K-50}
& HAT-L \cite{chen2023hat} (4$\times$ $\rightarrow$ 4$\times$) & 11.86  & 0.4699 & 22.82 & 0.3270 \\
& DiffBIR \cite{lin2024diffbir} (4$\times$ $\rightarrow$ 4$\times$) & \textit{3.36}  & \textbf{0.7588} & 37.17 & \underline{0.5916} \\
& DiffBIR \cite{lin2024diffbir} (16$\times$) & \textbf{2.65}  & 0.7078 & 38.59 & \textit{0.5858} \\
& OSEDiff \cite{wu2024one} (4$\times$ $\rightarrow$ 4$\times$) & 4.88  & \textit{0.7201} & \underline{39.88} & 0.5482  \\
& OSEDiff \cite{wu2024one} (16$\times$) & 8.37  & 0.5680 & 25.07 & 0.4210  \\
& PiSA-SR \cite{sun2024pixel} (4$\times$ $\rightarrow$ 4$\times$) & 5.01  & 0.7141 & 38.22 & 0.5364  \\
& PiSA-SR \cite{sun2024pixel} (16$\times$) & 9.30  & 0.5549 & 24.51 & 0.3861  \\
& AgenticIR \cite{zhu2024intelligent} & 5.13  & 0.5614 & \textit{39.55}  & 0.4814 \\
& \cellcolor{LightGray!70}\textbf{4KAgent (Gen4K-P)} & \cellcolor{LightGray!70}\underline{3.15} & \cellcolor{LightGray!70}\underline{0.7585} & \cellcolor{LightGray!70}\textbf{44.16} & \cellcolor{LightGray!70}\textbf{0.5928}  \\
\bottomrule
\end{tabular}
\end{table*}

\begin{figure*}[!t]
\centering
\includegraphics[width=\textwidth]{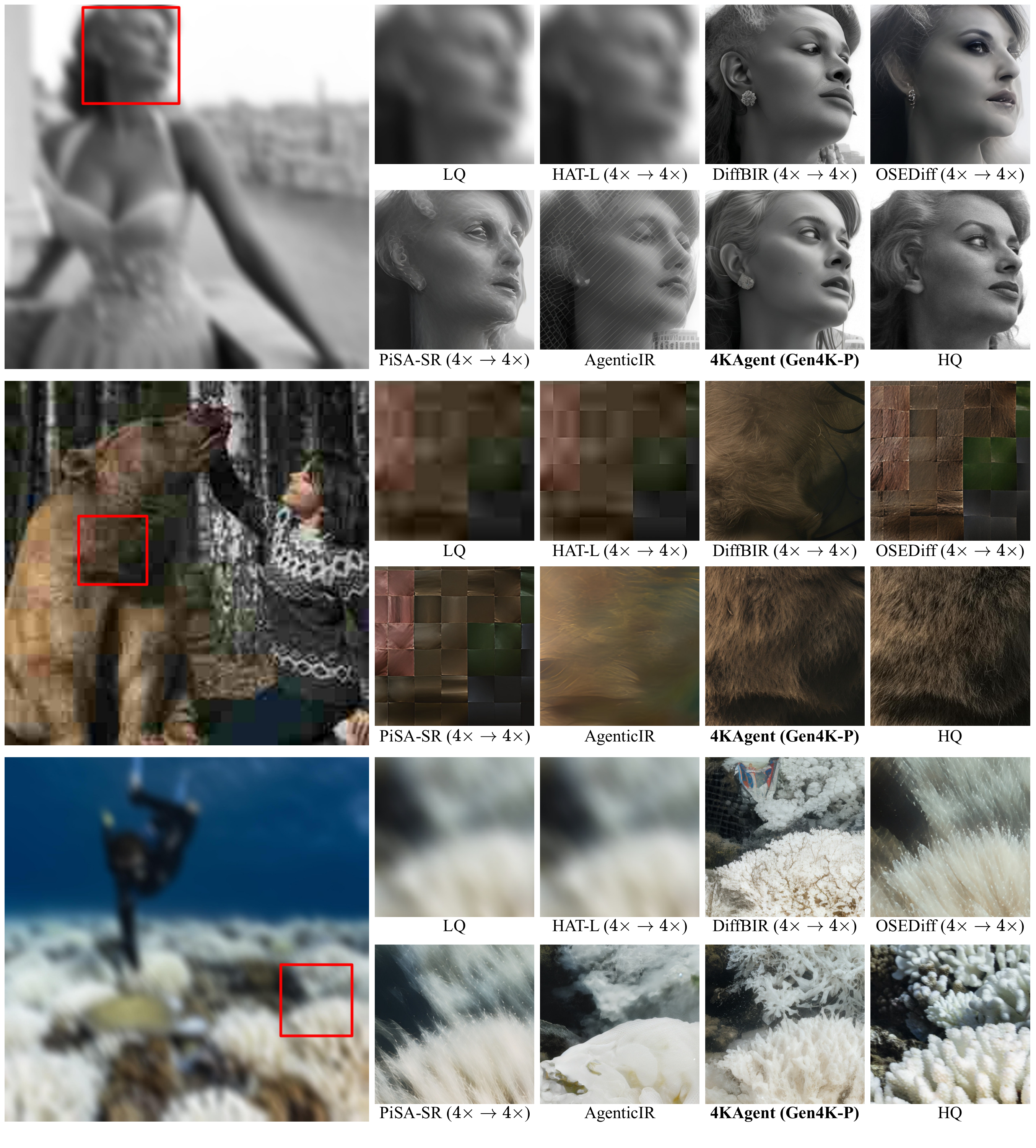}
\vspace{-15pt}
\caption{Visual comparisons on DIV4K-50 dataset. (Please zoom in to see details)}
\label{fig:div4k_compare}
\vspace{-4pt}
\end{figure*}

\paragraph{Qualitative Comparison}
As shown in~\cref{ssec:16x-sr}, directly upscaling images with the 16$\times$ setting often produces visually rich but unrealistic artifacts.
To ensure a fair and meaningful qualitative comparison in this experiment, we select previous methods with the 4$\times$$\rightarrow$ 4$\times$ setting, along with AgenticIR, as baselines.
Qualitative comparisons are shown in~\cref{fig:div4k_compare}.
Real-world image super-resolution methods generally recover more details than fidelity-based method.
However, their outputs still exhibit noticeable distortions. For example, the generated patches from OSEDiff and PiSA-SR in the middle row retain visible JPEG compression artifacts, which degrade the overall visual quality. While DiffBIR achieves the most favorable visual results among these methods, its outputs still suffer from either blurring or unrealistic artifacts.
AgenticIR performs competitively but tends to produce insufficiently sharp details.
4KAgent consistently reconstructs finer and more natural details. Notable examples include the facial features in the top row, the bear’s fur in the middle row, and the intricate coral textures in the bottom row, highlighting the superiority of our method.

\paragraph{Discussions}
The DIV4K-50 benchmark presents an extremely challenging setting, where fully recovering low-quality 256$\times$256 inputs to match the 4096$\times$4096 ground-truth images is virtually unattainable. Therefore, our focus shifts towards generating richly detailed and visually authentic textures.
Existing real-world image super-resolution methods struggle to fully correct the compounded degradations present in challenging scenarios. While these methods achieve competitive scores on perceptual metrics, their outputs tend to suffer from unrealistic hallucinated textures. Prior agentic systems, despite their effectiveness in handling multiple degradations, show limitations in maintaining sufficient sharpness when upscaling to 4K resolutions. 4KAgent demonstrates its capability to simultaneously address multiple degradations and extreme scaling factors, effectively reconstructing natural, fine-grained details with high visual fidelity.

\section{Experiment Part III: AI-Generated Content (AIGC) 4K Super-Resolution}
\label{sec:experiment-iii}

Text-to-visual models have ushered in a new era of high-quality image synthesis in AI-generated content. While existing models exhibit impressive capabilities in interpreting and following complex user instructions, their limitation to relatively low-resolution outputs (\eg, 1024 $\times$ 1024) poses significant challenges for applications requiring ultra-high visual fidelity, such as digital content creation and cinematic production. Scaling diffusion models for high-resolution generation entails computational overhead and access to large-scale high-resolution training data. As a practical alternative, pre-trained diffusion models can be repurposed for ultra-high-resolution image generation.

\subsection{AI-Generated Content (AIGC) 4K Super-Resolution}
\label{ssec:aigc-sr}
In this section, we investigate super-resolution in AIGC scenarios by comparing direct 4K generation with 1K upscaling using our 4KAgent. 

\paragraph{Settings.} As no prior method and dataset targets this setting, we sample 200 prompts from two standard AIGC benchmarks~\cite{wang2022diffusiondb,li2024genaibench}. For each prompt, 1K-resolution images were generated using several representative text-to-image models, including Flux.1-dev~\cite{flux2024}, Stable Diffusion 3 (SD3)~\cite{esser2024scaling}, PixArt-$\Sigma$~\cite{chen2024pixart}, SANA~\cite{xie2024sana}, and GPT-4o~\cite{hurst2024gpt4o}. In parallel, we employed native 4K-capable models such as SANA~\cite{xie2024sana} and Diffusion-4K~\cite{zhang2025diffusion4k} to directly synthesize 4K-resolution outputs. Due to stringent safety protocols, GPT-4o yielded only 39 valid 1K-resolution images from DiffusionDB prompts. We refer to the resulting datasets as GenAIBench-4K and DiffusionDB-4K.

We use the \textbf{ExpSR-s4-P} profile in 4KAgent here. To assess perceptual quality, we employ no-reference perceptual metrics. However, we observe that these metrics\,---\,particularly MUSIQ\,---\,are not tailored for evaluating ultra-high-resolution images, likely due to their inability to capture fine-grained details through a multi-scale architecture. To mitigate this limitation, we introduce \textbf{MUSIQ-P}, a patch-applied variant that computes MUSIQ scores over non-overlapping $512 \times 512$ patches and averages them, thereby improving sensitivity to localized artifacts in ultra-high-resolution content.

\paragraph{Quantitative Comparison.}\cref{table:performance-aigc-dataset} presents the quantitative results across three strategies: (1) native 4K generation, (2) 1K-resolution generation, and (3) 1K-resolution images upscaled by 4KAgent. On GenAIBench-4K, the SANA-1K + 4KAgent pipeline achieves a NIQE score of 3.03 and a CLIPIQA of 0.7050, significantly outperforming SANA-4K (NIQE 4.02, CLIPIQA 0.6172). Similarly, PixArt-$\Sigma$ + 4KAgent obtains the best NIQE score (2.76) among all methods, while SD3-Medium + 4KAgent achieves the best CLIPIQA (0.7169) and MANIQA (0.5155) scores. On DiffusionDB-4K, several models such as SANA-1K, GPT-4o, PixArt-$\Sigma$, and SD3-Medium, when upscaled with 4KAgent, achieve significantly lower NIQE and higher CLIPIQA scores. Although MUSIQ-P scores for the upscaled images show a slight decrease relative to their 1K counterparts, the difference remains marginal, suggesting limited perceptual degradation during upscaling.
\begin{table*}[!h]
\renewcommand{\arraystretch}{1.25}
\setlength{\tabcolsep}{4pt}
\centering
\scriptsize
\caption{Comprehensive quantitative comparison of AIGC  4$\times$ Super-Resolution. The top three performances of each metric are marked in \textbf{bold}, \underline{underline}, \textit{italic} respectively. MUSIQ-P$^*$ indicates a patch-applied variant of the MUSIQ metric for evaluating ultra-high-resolution (4K) images.}
\label{table:performance-aigc-dataset}
\begin{tabular}{l|cccc|cccc}
\toprule
\textbf{Dataset} 
& \multicolumn{4}{c|}{\textbf{GenAIBench-4K~\cite{li2024genaibench}}} 
& \multicolumn{4}{c}{\textbf{DiffusionDB-4K~\cite{wang2022diffusiondb}}} \\
\midrule
\textbf{Model} 
& \textbf{NIQE$\downarrow$} & \textbf{CLIPIQA$\uparrow$} & \textbf{MUSIQ-P$^*$$\uparrow$} & \textbf{MANIQA$\uparrow$}
& \textbf{NIQE$\downarrow$} & \textbf{CLIPIQA$\uparrow$} & \textbf{MUSIQ-P$^*$$\uparrow$} & \textbf{MANIQA$\uparrow$} \\
\midrule
SANA-4K~\cite{xie2024sana} & 4.02 & 0.6172 & 47.93 & 0.3673 & 3.74 & 0.6005 & 48.66 & 0.3425 \\
Diffusion-4K~\cite{zhang2025diffusion4k} & 6.38 & 0.5049 & 35.07 & 0.3535 & 6.55 & 0.5056 & 35.87 & 0.3404 \\
\midrule
SANA-1K~\cite{xie2024sana} & 4.18 & \underline{0.7147} & \textbf{66.30} & 0.4814 & 3.80 & 0.6910 & \underline{67.99} & \underline{0.5104} \\
\cellcolor{LightGray!70}\hspace{10mm}+ 4KAgent & \cellcolor{LightGray!70}3.03 & \cellcolor{LightGray!70}0.7050 & \cellcolor{LightGray!70}57.97 & \cellcolor{LightGray!70}0.4735 & \cellcolor{LightGray!70}3.04 & \cellcolor{LightGray!70}0.7082 & \cellcolor{LightGray!70}60.48 & \cellcolor{LightGray!70}0.4715 \\
GPT4o~\cite{hurst2024gpt4o} & 5.69 & 0.6607 & \textit{64.43} & 0.4997 & 5.13 & 0.6275 & 62.53 & 0.4398 \\
\cellcolor{LightGray!70}\hspace{10mm}+ 4KAgent & \cellcolor{LightGray!70}3.56 & \cellcolor{LightGray!70}0.7016 & \cellcolor{LightGray!70}58.28 & \cellcolor{LightGray!70}0.4976 & \cellcolor{LightGray!70}3.40 & \cellcolor{LightGray!70}0.6867 & \cellcolor{LightGray!70}56.67 & \cellcolor{LightGray!70}0.4711 \\
FLUX.1-dev~\cite{flux2024} & 6.18 & 0.6768 & 61.02 & \textit{0.5018} & 5.33 & \textbf{0.7509} & \textbf{69.69} & \textbf{0.5835} \\
\cellcolor{LightGray!70}\hspace{10mm}+ 4KAgent & \cellcolor{LightGray!70}\textit{2.98} & \cellcolor{LightGray!70}\textit{0.7078} & \cellcolor{LightGray!70}58.19 & \cellcolor{LightGray!70}\underline{0.5034} & \cellcolor{LightGray!70}\textit{3.04} & \cellcolor{LightGray!70}\underline{0.7440} & \cellcolor{LightGray!70}60.88 & \cellcolor{LightGray!70}\textit{0.5056} \\
PixArt-$\Sigma$~\cite{chen2024pixart} & 4.12 & 0.6960 & 63.74 & 0.4415 & 3.66 & 0.6892 & \textit{66.54} & 0.4386 \\
\cellcolor{LightGray!70}\hspace{10mm}+ 4KAgent & \cellcolor{LightGray!70}\textbf{2.76} & \cellcolor{LightGray!70}0.7077 & \cellcolor{LightGray!70}56.71 & \cellcolor{LightGray!70}0.4699 & \cellcolor{LightGray!70}\textbf{2.88} & \cellcolor{LightGray!70}\textit{0.7092} & \cellcolor{LightGray!70}58.85 & \cellcolor{LightGray!70}0.4659 \\
SD3-Medium~\cite{esser2024scaling} & 5.03 & 0.6922 & \underline{64.68} & 0.4767 & 4.38 & 0.6667 & 65.99 & 0.4413 \\
\cellcolor{LightGray!70}\hspace{10mm}+ 4KAgent & \cellcolor{LightGray!70}\underline{2.99} & \cellcolor{LightGray!70}\textbf{0.7169} & \cellcolor{LightGray!70}60.22 & \cellcolor{LightGray!70}\textbf{0.5155} & \cellcolor{LightGray!70}\underline{2.99} & \cellcolor{LightGray!70}0.7066 & \cellcolor{LightGray!70}59.35 & \cellcolor{LightGray!70}0.4747 \\ 
\bottomrule
\end{tabular}
\end{table*}

To further assess semantic and aesthetic fidelity, we report PickScore~\cite{kirstain2023pickscore} in~\cref{table:aigc-dataset-pickscore}, which quantitatively captures diversity and human-aligned visual quality. On GenAIBench-4K, models enhanced with 4KAgent outperform their native 4K counterparts. On DiffusionDB-4K, the performance gap is smaller, which may be attributed to the dataset's richer and more descriptive prompt content.
\begin{table*}[!h]
\renewcommand{\arraystretch}{1.25}
\setlength{\tabcolsep}{5pt}
\centering
\scriptsize
\captionsetup{font=footnotesize}
\caption{Comparison of PickScore-Based~\cite{kirstain2023pickscore} Quantitative Evaluation Between Native 4K and 4KAgent-Upscaled 1K Models.}
\label{table:aigc-dataset-pickscore}
\begin{tabular}{l|c|cc|cc}
\toprule
\textbf{Dataset} & \textbf{Avg. Prompt Length} & \textbf{SANA-4K} & \cellcolor{LightGray!70}\textbf{SANA-1K + 4KAgent} & \textbf{Diffusion-4K} & \cellcolor{LightGray!70}\textbf{Flux.1-dev + 4KAgent}\\
\midrule
GenAIBench-4K~\cite{li2024genaibench} & 12.13 & 0.4482 & \cellcolor{LightGray!70}\textbf{0.5518} & 0.2389 & \cellcolor{LightGray!70}\textbf{0.7611} \\
DiffusionDB-4K~\cite{wang2022diffusiondb} & 25.29 & 0.4893 & \cellcolor{LightGray!70}\textbf{0.5107} & 0.2406 & \cellcolor{LightGray!70}\textbf{0.7594}\\
\bottomrule
\end{tabular}
\end{table*}

\begin{figure*}[!h]
\centering
\footnotesize
\begin{tabular}{@{}c@{}}
\includegraphics[width=\textwidth]{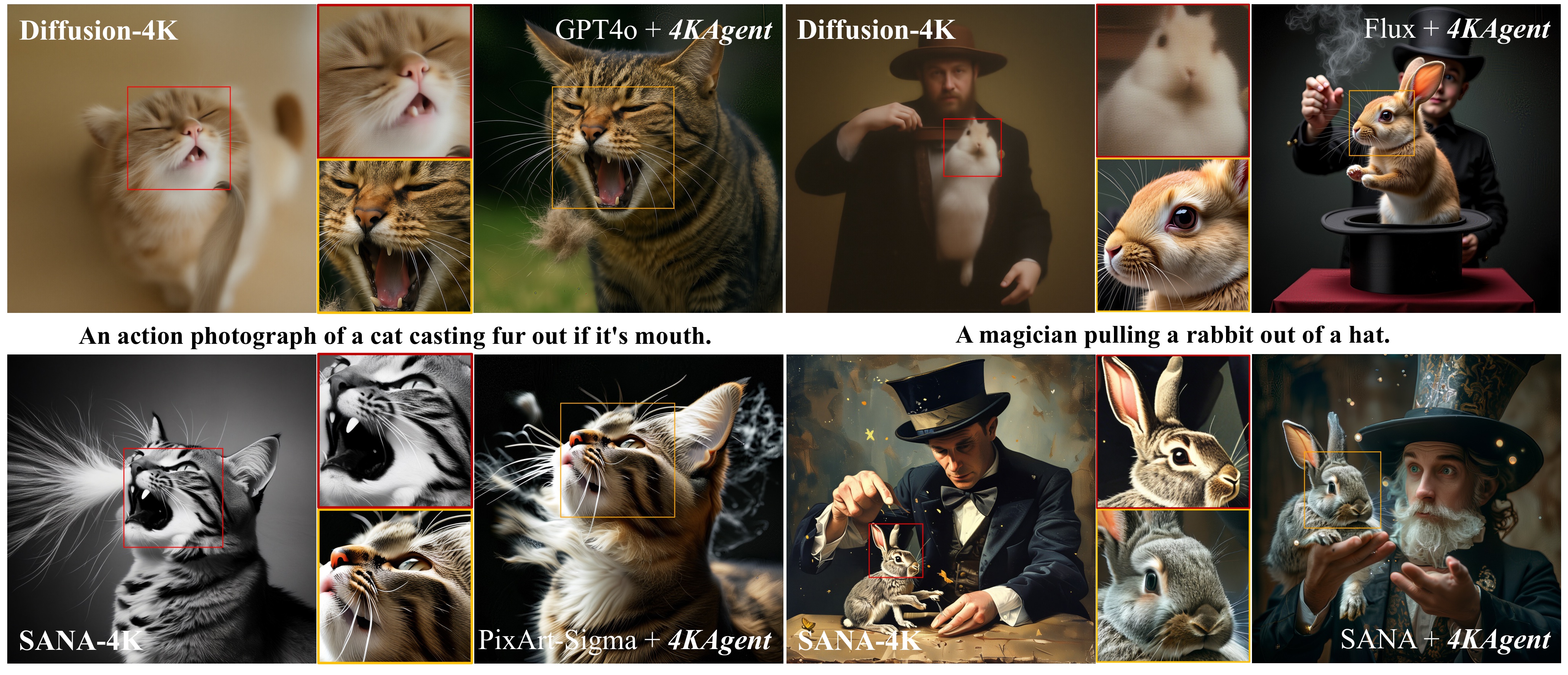} \\
\includegraphics[width=\textwidth]{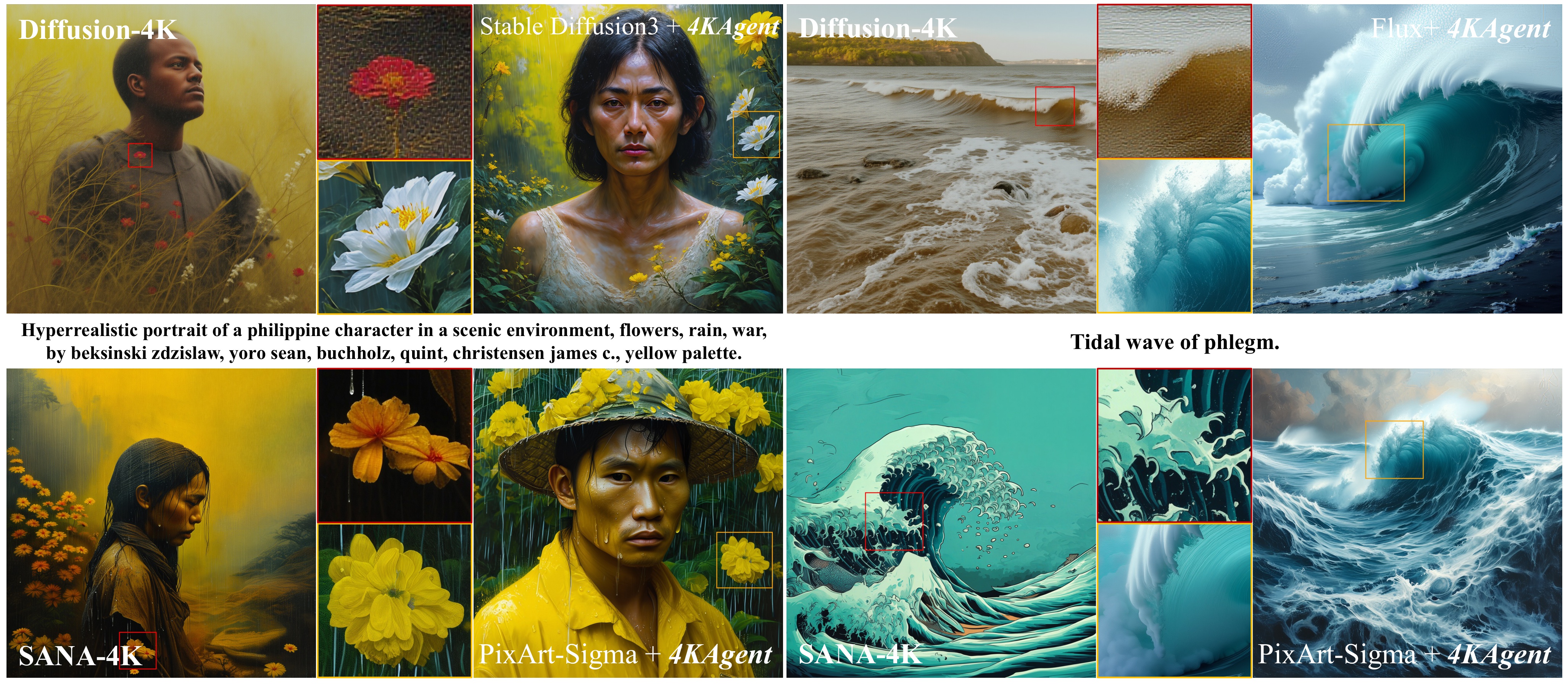}
\end{tabular}
\vspace{-5pt}
\caption{Visual comparison between native 4K image generation and 1K image generation methods with 4KAgent, using identical prompts. 4K images from Diffusion-4K and SANA-4K are displayed on the left, while the corresponding outputs enhanced by 4KAgent are shown on the right.}
\vspace{-10pt}
\label{fig:aigc_4k_comparsion}
\end{figure*}

\begin{figure*}[!h]
\centering
\footnotesize
\begin{tabular}{@{}c@{}}
\includegraphics[width=\textwidth]{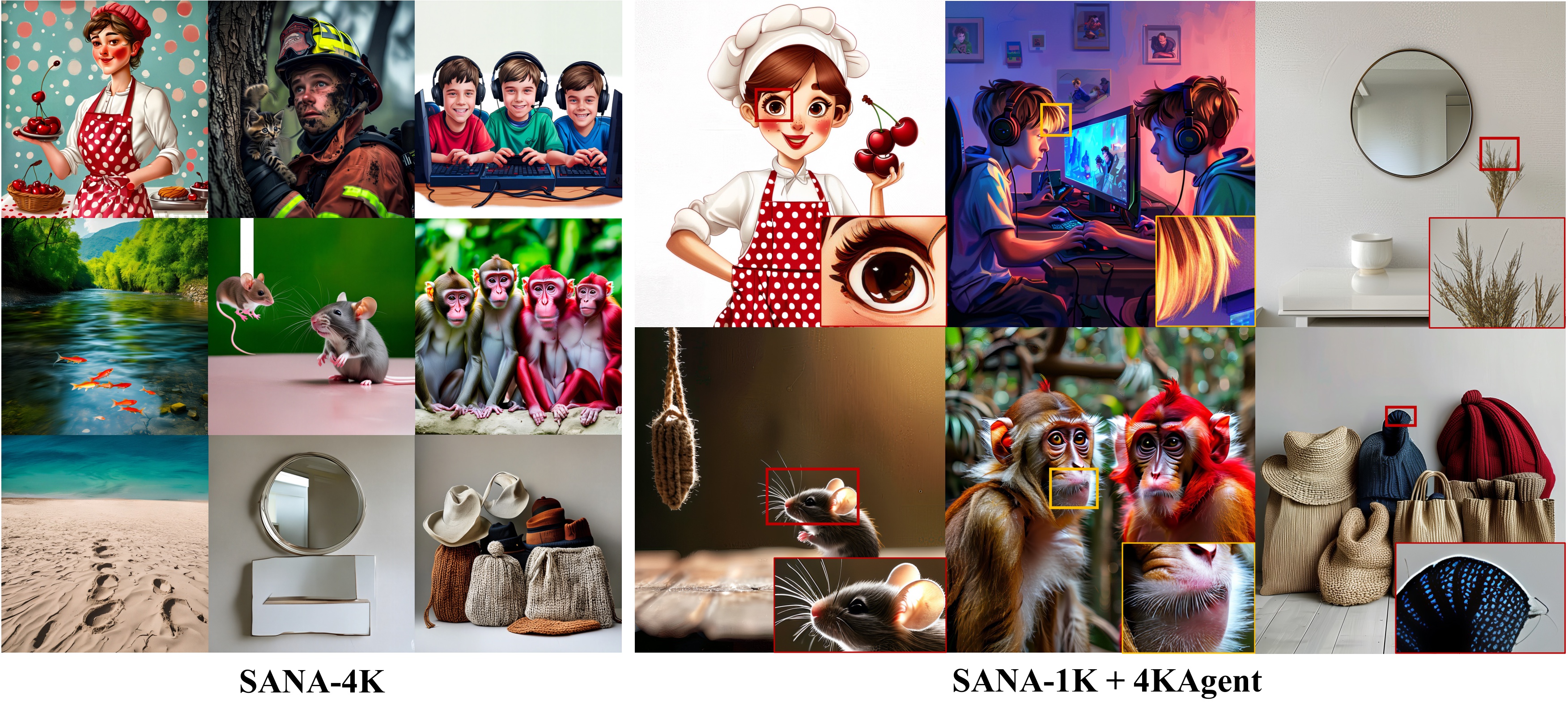}
\end{tabular}
\vspace{-5pt}
\caption{Visual comparison of aesthetic preference alignment between SANA-4K and SANA-1K+4KAgent using identical prompts sampled from GenAIBench-4K. SANA-1K+4KAgent yields superior aesthetic alignment and richer high-resolution details, highlighted in the zoomed-in patches.}
\vspace{-10pt}
\label{fig:aigc_sana4k_sana1k_4kagent_comparsion}
\end{figure*}

\paragraph{Qualitative Comparison}\cref{fig:aigc_4k_comparsion} shows qualitative results of applying 4KAgent to various base models under identical prompts. Across different models, 4KAgent consistently enhances visual fidelity and preserves fine-grained details. As shown in~\cref{fig:aigc_sana4k_sana1k_4kagent_comparsion}, images upscaled from SANA-1K using 4KAgent exhibit richer textures and stronger aesthetic alignment than those generated natively at 4K resolution by SANA-4K.

\paragraph{Discussions}
The application of our 4KAgent in AIGC scenarios leads to substantial improvements in image quality when upscaling 1K-resolution images to 4K.
\textbf{First}, when applied to 1K-resolution inputs, 4KAgent consistently achieves notable gains across multiple quantitative benchmarks, enabling more detailed and accurate reconstructions in the resulting 4K outputs. As traditional metrics are not specifically tuned for ultra-high-resolution images, we adopted the adaptive MUSIQ-P metric to enable a perceptually-focused evaluation. The results indicate that 4K-upscaled images achieve perceptual quality scores comparable to their original 1K counterparts.
\textbf{Second}, 4KAgent demonstrates strong capability in synthesizing high-fidelity visual details and intricate textures. However, as with many traditional super-resolution methods, this training-free framework occasionally introduces unintended bokeh-like artifacts, particularly in blurred background regions. Given 4KAgent's modular and scalable design, we believe integrating task-specific profile configurations and perceptual alignment strategies could further reduce artifacts and improve robustness in diverse AIGC applications.

\section{Experiment Part IV: Scientific Imagery Super-Resolution}
\label{sec:experiment-iv}
\subsection{Remote Sensing Image Super-Resolution}
\label{ssec:sci-sr-remote-sensing}
High-resolution satellite imagery is foundational for a wide spectrum of remote sensing tasks, including urban planning, environmental monitoring, and disaster response \cite{he2021spatial,shermeyer2019effects}. However, due to cost, bandwidth, and sensing constraints, acquiring such high-resolution imagery globally at very high frequency remains very expensive or even impractical. Recent advances in deep learning-based super-resolution have provided a promising alternative by reconstructing high-fidelity imagery from lower-resolution observations \cite{kowaleczko2023real}. In this section, we evaluate 4KAgent against state-of-the-art baselines on a diverse set of real-world satellite image super-resolution datasets.

\paragraph{Settings.}
We evaluate our models on four benchmark datasets covering varied land-use patterns and sensing characteristics:
\begin{itemize}[leftmargin=*]
    \item AID~\cite{7907303} is a large-scale dataset constructed to benchmark aerial scene classification methods. It includes over 10,000 high-resolution aerial images across 30 scene categories, such as airports, industrial areas, and farmlands. Each image has a resolution of 600$\times$600 with a spatial resolution of 0.5-8 m/pixel. Images exhibit high intra-class diversity and low inter-class variation, making them well-suited for evaluating generalization in SR tasks.
    \item DIOR~\cite{li2020object} is a comprehensive object detection benchmark in the remote sensing domain, containing 23,463 images and 192,472 annotated object instances across 20 categories. The resolution of images is 800$\times$800, and the spatial resolutions range from 0.5m to 30m. These images exhibit high diversity in resolution, imaging conditions, and object scale.
    \item DOTA~\cite{8578516} consists of 2,806 ultra-high-resolution aerial images collected from various sensors. It features over 188,000 labeled object instances with arbitrary orientations. Each image is in resolution about 4000$\times$4000. The combination of large-scale, fine-grained annotations and high inter-scene variability makes DOTA particularly valuable for evaluating perceptual fidelity.
    \item WorldStrat~\cite{cornebise2022open} is a unique dataset designed for real-world satellite image super-resolution tasks, with globally stratified land use coverage across 10,000 $\text{km}^2$. It pairs 1054$\times$1054 pixel high-resolution (SPOT 6/7, 1.5 m/pixel) and temporally matched low-resolution (Sentinel-2, 10 m/pixel) imagery for thousands of regions worldwide. Importantly, unlike synthetic degradation benchmarks, WorldStrat contains real cross-sensor-captured low-resolution (LR) and high-resolution (HR) image pairs, introducing natural misalignment and color mismatches due to different sensor characteristics.
\end{itemize}

From each dataset, we select 100-200 representative scenes. Following~\cite{xiao2024frequency,xiao2023ediffsr}, for AID, DIOR, and DOTA, HR images are downsampled using bicubic interpolation to generate corresponding LR inputs. For WorldStrat, we adopt the dataset’s official pre-processing pipeline, selecting LR images that are temporally closest to each HR acquisition. Notably, due to the different sensors used for LR and HR captures, RGB content may exhibit significant variation, posing a realistic challenge for super-resolution. To test the generalization ability of our models, we evaluate on a spectrum of resolution scales: 1) 4$\times$ SR (128$\rightarrow$512) on DIOR and DOTA datasets; 2) 4$\times$ SR (160$\rightarrow$640) on AID and WorldStrat datasets; 3) 4$\times$ SR (512$\rightarrow$2048) for high-res DOTA scenes; 4) 16$\times$ SR (e.g., 256$\rightarrow$4096) for DOTA scenes.

We evaluate 4KAgent with the \textbf{ExpSR-s4-F} profile and \textbf{ExpSR-s4-P} profile in 4$\times$ super resolution for \textbf{Fidelity} and \textbf{Perception} preference, respectively. Then we evaluate 4KAgent with the \textbf{Aer4K-F} profile and \textbf{Aer4K-P} profile in 16$\times$ super resolution for \textbf{Fidelity} and \textbf{Perception} preference, respectively. We benchmark 4KAgent against the following categories of SR models: 1) Expert aerial SR models: HAUNet~\cite{10145829}, TransENet~\cite{9654169}; 2) Fidelity-based SR models: HAT-L~\cite{chen2023hat}, PiSA-SR-PSNR~\cite{sun2024pixel}, and SwinIR~\cite{liang2021swinir}; 3) Perception-based SR Models: DiffBIR~\cite{lin2024diffbir}, OSEDiff~\cite{wu2024one}, HAT-GAN~\cite{chen2023hat}, PiSA-SR~\cite{sun2024pixel}, and SwinIR (Real-ISR)~\cite{liang2021swinir}. For `PiSA-SR-PSNR', we set the pixel guidance factor $\lambda_{pix}=1.0$ and semantic guidance factor $\lambda_{sem}=0$ for PiSA-SR~\cite{sun2024pixel} in inference. Additionally, we include AgenticIR for agentic system comparison. Together, these diverse datasets and models enable a comprehensive evaluation of 4KAgent in terms of both pixel fidelity and perceptual quality, across synthetic and real-world degradation settings.

\paragraph{Quantitative Comparison}
We report the comparison results on AID (4x SR, 160$\rightarrow$640), DIOR (4$\times$ SR, 128$\rightarrow$512), DOTA (4$\times$ SR, 128$\rightarrow$512), WorldStrat (4$\times$ SR, 160$\rightarrow$640), DOTA (4$\times$ SR, 512$\rightarrow$2048), and DOTA (16$\times$ 4K SR) in~\cref{table:performance-aid,table:performance-dior,table:performance-dota-128,table:performance-worldstrat,table:performance-dota-512,table:performance-dota-4k}, respectively. Across all six benchmark settings, 4KAgent with \textbf{Fidelity} preference consistently demonstrates superior performance in terms of \textbf{pixel-level reconstruction}. It ranks within the top three PSNR in four out of six tasks, and ranks within the top three in SSIM across all synthetic scenarios. This confirms its ability to preserve structural details across scales and domains. Importantly, 4KAgent also consistently outperforms AgenticIR by a large margin across all fidelity metrics and tasks, highlighting its effectiveness.

In terms of perceptual quality, 4KAgent with \textbf{Perception} preference achieves top performance in \textbf{perceptual quality} assessment metrics across multiple datasets. Notably, on the WorldStrat, 4KAgent (ExpSR-s4-P) ranks first in MUSIQ, MANIQA, and CLIPIQA. These results indicate that 4KAgent not only produces photorealistic outputs but also maintains robustness in real-world, sensor-misaligned scenarios. Compared to AgenticIR, 4KAgent with \textbf{Perception} preference shows clear gains across all perceptual IQA metrics, reaffirming the value of 4KAgent in balancing realism and structure across diverse and challenging remote sensing settings.
\begin{table*}[!h]
\renewcommand{\arraystretch}{1.2}
\centering
\fontsize{6.5pt}{7.5pt}\selectfont
\setlength{\tabcolsep}{2.5pt}
\caption{4$\times$ performance comparison of evaluated models on the AID dataset (160$\rightarrow$640). The top three performances of each metric are marked in \textbf{bold}, \underline{underline}, \textit{italic} respectively.}
\label{table:performance-aid}
\begin{tabular}{l|l|ccccccccc}
\toprule
{\textbf{DataSet}} & {\textbf{Model}}   & \textbf{PSNR$\uparrow$}&  \textbf{SSIM$\uparrow$} &  \textbf{LPIPS$\downarrow$}  &  \textbf{DISTS$\downarrow$}  &  \textbf{FID$\downarrow$}  &  \textbf{NIQE$\downarrow$}  &  \textbf{MUSIQ$\uparrow$}  &  \textbf{MANIQA$\uparrow$}  &  \textbf{CLIPIQA$\uparrow$}   \\ 
\midrule
\multirow{3}{*}{Fidelity-based SR}
& SwinIR~\cite{liang2021swinir} & \textit{28.4887} & \textbf{0.7422} & 0.4355 & 0.2473 & 186.1843 & 7.4295 & 50.1222 & 0.3108 & 0.2563 \\
& HAT-L~\cite{chen2023hat} & 27.1630 & 0.6835 & 0.4635 & 0.2330 & 126.9911 & 7.3586 & 36.4299 & 0.4149 & 0.4111 \\
& PiSA-SR-PSNR~\cite{sun2024pixel} & 27.9079 & \underline{0.7251} & 0.4273 & 0.4273 & 144.3109 & 7.2371 & 41.6239 & 0.4024 & 0.1831 \\
\midrule
\multirow{5}{*}{Perception-based SR}
& SwinIR (Real-ISR)~\cite{liang2021swinir} & 26.5090 & 0.6700 &\textbf{0.3344} & \textbf{0.1928} & 129.3879 & \textit{3.8690} & 60.6544 & 0.5641 & 0.5205 \\
& HAT-GAN~\cite{chen2023hat}& 25.7860 & 0.6643 & \textit{0.3522} & \textit{0.2137} & 140.2364 & 4.8258 & 55.4862 & 0.5618 & 0.3556 \\
& DiffBIR~\cite{lin2024diffbir} & 24.8343 & 0.5554 & 0.4466 & 0.2374 & 130.6386 & 4.8871 & \textit{65.9636} & \textit{0.6342} & \textbf{0.7302} \\
& OSEDiff~\cite{wu2024one} & 25.2220 & 0.6164 & \underline{0.3497} & 0.3497 & \textbf{91.5957} & \underline{3.6661} & 63.9855 & 0.6219 & 0.6074 \\
& PiSA-SR~\cite{sun2024pixel} & 24.5971 & 0.5903 & 0.3541 & 0.3541 & \underline{115.8287} & \textbf{3.4859} & \underline{66.0433} & \textbf{0.6555} & 0.6346 \\
\midrule
\multirow{2}{*}{Expert Aerial SR}
& HAUNet~\cite{10145829} & \underline{28.5136} & 0.7146 & 0.4327 & \underline{0.2083} & \textit{122.1656} & 7.2497 & 35.5021 & 0.4162 & 0.1706 \\
 & TransENet~\cite{9654169} & 28.0317 & 0.6983 & 0.4179 & 0.2109 & 125.9495 & 6.7700 & 35.1140 & 0.3776 & 0.1162 \\
\midrule
\multirow{3}{*}{Agentic System}
& AgenticIR~\cite{zhu2024intelligent} & 21.3431 & 0.5147 & 0.4600 & 0.2539 & 149.7191 & 4.5325 & 67.0257 & 0.6283 & \textit{0.6693} \\
& \cellcolor{LightGray!70}\textbf{4KAgent (ExpSR-s4-F)} & \cellcolor{LightGray!70}\textbf{28.5481} & \cellcolor{LightGray!70}\textit{0.7157} & \cellcolor{LightGray!70}0.4436 & \cellcolor{LightGray!70}0.2263 & \cellcolor{LightGray!70}127.4411 & \cellcolor{LightGray!70}7.5916 & \cellcolor{LightGray!70}37.5713 & \cellcolor{LightGray!70}0.3774 & \cellcolor{LightGray!70}0.4322 \\
& \cellcolor{LightGray!70}\textbf{4KAgent (ExpSR-s4-P)} & \cellcolor{LightGray!70}24.4212 & \cellcolor{LightGray!70}0.5354 & \cellcolor{LightGray!70}0.4696 &  \cellcolor{LightGray!70}0.2566& \cellcolor{LightGray!70}139.7775 &  \cellcolor{LightGray!70}4.8915& \cellcolor{LightGray!70}\textbf{68.1858} & \cellcolor{LightGray!70}\underline{0.6439} & \cellcolor{LightGray!70}\underline{0.6897} \\
\bottomrule
\end{tabular}
\end{table*}

\begin{table*}[!h]
\renewcommand{\arraystretch}{1.2}
\centering
\fontsize{6.5pt}{7.5pt}\selectfont
\setlength{\tabcolsep}{2.5pt}
\caption{4$\times$ performance comparison of evaluated models on the DIOR dataset (128$\rightarrow$512). The top three performances of each metric are marked in \textbf{bold}, \underline{underline}, \textit{italic} respectively.}
\label{table:performance-dior}
\begin{tabular}{l|l|ccccccccc}
\toprule
{\textbf{Type}} & {\textbf{Model}}   & \textbf{PSNR$\uparrow$}&  \textbf{SSIM$\uparrow$} &  \textbf{LPIPS$\downarrow$}  &  \textbf{DISTS$\downarrow$}  &  \textbf{FID$\downarrow$}  &  \textbf{NIQE$\downarrow$}  &  \textbf{MUSIQ$\uparrow$}  &  \textbf{MANIQA$\uparrow$}  &  \textbf{CLIPIQA$\uparrow$}   \\
\midrule
\multirow{3}{*}{Fidelity-based SR}
& SwinIR~\cite{liang2021swinir} & \textbf{27.8751} & \textbf{0.7257} & 0.4474 & 0.2488 & 223.4653 & 7.1247 & 51.6481 & 0.3122 & 0.2441 \\
& HAT-L~\cite{chen2023hat} & \textit{27.7355} & 0.6962 & 0.4586 & 0.2195 & 134.4649 & 7.1142 & 37.1784 & 0.4151 & 0.4194 \\
& PiSA-SR-PSNR~\cite{sun2024pixel} & 27.4176 & \underline{0.7118} & 0.4378 & 0.2202 & 167.7604 & 6.9455 & 42.5087 & 0.4095 & 0.1881 \\
\midrule
\multirow{5}{*}{Perception-based SR}
& SwinIR (Real-ISR)~\cite{liang2021swinir} & 26.4708 & 0.6698 & \textbf{0.3391} & \underline{0.1983} & 144.3900 & \textit{3.8921} & 60.6319 & 0.5552 & 0.5091 \\
& HAT-GAN~\cite{chen2023hat} & 26.8015 & 0.6848 & \underline{0.3398} & \textit{0.2073} & 149.8121 & 4.8459 & 55.8046 & 0.5592 & 0.3392 \\
& DiffBIR~\cite{lin2024diffbir} & 24.9254 & 0.5742 & 0.4201 & 0.2317 & 146.2642 & 4.9198 & \textit{66.4572} & \textit{0.6315} & \textbf{0.7078} \\
& OSEDiff~\cite{wu2024one} & 25.0470 & 0.6207 & \textit{0.3506} & \textbf{0.1875} & \textbf{127.7888} & \underline{3.6641} & 65.3934 & 0.6245 & 0.5976 \\
& PiSA-SR~\cite{sun2024pixel} & 24.4078 & 0.5932 & 0.3534 & 0.3534 & \textit{129.2724} & \textbf{3.5111} & \underline{67.6365} & \textbf{0.6571} & 0.6229 \\
\midrule
\multirow{2}{*}{Expert Aerial SR}
& HAUNet~\cite{10145829} & \underline{27.8221} & 0.6992 & 0.4527 & 0.2100 & \underline{128.8770} & 6.9586 & 35.5885 & 0.4089 & 0.1572 \\
& TransENet~\cite{9654169} & 27.3002 & 0.6824 & 0.4391 & 0.2113 & 129.4893 & 6.4986 & 34.7713 & 0.3750 & 0.0984 \\
\midrule
\multirow{3}{*}{Agentic System}
& AgenticIR~\cite{zhu2024intelligent} & 22.4811 & 0.5654 & 0.4668 & 0.2388 & 169.2341 & 4.7446 & 63.1399 & 0.5938 & \textit{0.6252} \\
& \cellcolor{LightGray!70}\textbf{4KAgent (ExpSR-s4-F)} & \cellcolor{LightGray!70}27.6761 & \cellcolor{LightGray!70}\textit{0.7062} & \cellcolor{LightGray!70}0.4309 & \cellcolor{LightGray!70}0.2250 & \cellcolor{LightGray!70}146.5618 & \cellcolor{LightGray!70}7.2555 & \cellcolor{LightGray!70}37.5543 & \cellcolor{LightGray!70}0.3811 & \cellcolor{LightGray!70}0.4368 \\
& \cellcolor{LightGray!70}\textbf{4KAgent (ExpSR-s4-P)} &  \cellcolor{LightGray!70}24.4893 &\cellcolor{LightGray!70}0.5795 & \cellcolor{LightGray!70}0.4374& \cellcolor{LightGray!70}0.2471 &\cellcolor{LightGray!70}160.6006 &\cellcolor{LightGray!70}4.6522 &\cellcolor{LightGray!70}\textbf{68.0117}  & \cellcolor{LightGray!70}\underline{0.6358} & \cellcolor{LightGray!70}\underline{0.6456} \\
\bottomrule
\end{tabular}
\end{table*}

\begin{table*}[!h]
\renewcommand{\arraystretch}{1.2}
\centering
\fontsize{6.5pt}{7.5pt}\selectfont
\setlength{\tabcolsep}{2.5pt}
\caption{4$\times$ performance comparison of evaluated models on the DOTA dataset (128$\rightarrow$512). The top three performances of each metric are marked in \textbf{bold}, \underline{underline}, \textit{italic} respectively.}
\label{table:performance-dota-128}
\begin{tabular}{l|l|ccccccccc}
\toprule
{\textbf{Type}} & {\textbf{Model}}   & \textbf{PSNR$\uparrow$}&  \textbf{SSIM$\uparrow$} &  \textbf{LPIPS$\downarrow$}  &  \textbf{DISTS$\downarrow$}  &  \textbf{FID$\downarrow$}  &  \textbf{NIQE$\downarrow$}  &  \textbf{MUSIQ$\uparrow$}  &  \textbf{MANIQA$\uparrow$}  &  \textbf{CLIPIQA$\uparrow$}   \\ 
\midrule
\multirow{3}{*}{Fidelity-based SR}
& HAT-L~\cite{chen2023hat} & \textbf{33.0720} & \textbf{0.8656} & \textbf{0.2448} & \underline{0.1471} & \underline{58.0105} & 6.6527 & 51.7547 & 0.5858 & 0.3725 \\
& PiSA-SR-PSNR~\cite{sun2024pixel} & 28.9623 & 0.7999 & 0.3415 & 0.2093 & 133.7664 & 7.7350 & 47.4847 & 0.4878 & 0.3094 \\
& SwinIR~\cite{liang2021swinir} & 30.5969 & 0.8254 & 0.3275 & 0.2215 & 143.2866 & 7.5391 & 54.3111 & 0.4526 & 0.2700 \\
\midrule
\multirow{5}{*}{Perception-based SR}
& DiffBIR~\cite{lin2024diffbir} & 25.8326 & 0.6489 & 0.3724 & 0.2340 & 115.1440 & 6.0906 & 64.9539 & \textit{0.6535} & \textit{0.6772} \\
& OSEDiff~\cite{wu2024one} & 26.3616 & 0.7156 & 0.3324 & 0.2133 & 126.4670 & \textit{5.4257} & 64.1220 & 0.6278 & 0.6736 \\
& HAT-GAN~\cite{chen2023hat} & 28.6557 & 0.7869 & 0.2751 & 0.1818 & 115.1743 & 5.7245 & 57.0159 & 0.5929 & 0.3691 \\
& PiSA-SR~\cite{sun2024pixel} & 25.8447 & 0.6921 & 0.3220 & 0.2081 & 112.1042 & \underline{4.9062} & \underline{66.3901} & \underline{0.6676} & \textbf{0.6855} \\
& SwinIR (Real-ISR)~\cite{liang2021swinir} & 28.9000 & 0.7883 & \textit{0.2657} & 0.1769 & 110.4552 & \textbf{4.7712} & 59.7489 & 0.5886 & 0.4519 \\
\midrule
\multirow{2}{*}{Expert Aerial SR}
& HAUNet~\cite{10145829} & \underline{32.8286} & \underline{0.8627} & \underline{0.2480} & \textbf{0.1428} & \textbf{57.3008} & 6.5917 & 50.7492 & 0.5711 & 0.3824 \\
& TransENet~\cite{9654169}  & 30.7214 & 0.8176 & 0.2883 & \textit{0.1553} & \textit{68.5120} & 6.2878 & 42.8957 & 0.4856 & 0.3441 \\
\midrule
\multirow{3}{*}{Agentic System}
& AgenticIR~\cite{zhu2024intelligent} & 19.9655 & 0.5973 & 0.4227 & 0.2620 & 137.2777 & 6.3126 & \textit{65.5596} & 0.6375 & 0.6198 \\
& \cellcolor{LightGray!70}\textbf{4KAgent (ExpSR-s4-F)} & \cellcolor{LightGray!70}\textit{31.3589} & \cellcolor{LightGray!70}\textit{0.8478} & \cellcolor{LightGray!70}0.2853 & \cellcolor{LightGray!70}0.1776 & \cellcolor{LightGray!70}88.0366 & \cellcolor{LightGray!70}7.0808 & \cellcolor{LightGray!70}50.6815 & \cellcolor{LightGray!70}0.5515 & \cellcolor{LightGray!70}0.3799 \\
& \cellcolor{LightGray!70}\textbf{4KAgent (ExpSR-s4-P)} & \cellcolor{LightGray!70}24.9224 & \cellcolor{LightGray!70}0.6427&  \cellcolor{LightGray!70}0.3884& \cellcolor{LightGray!70}0.2555 & \cellcolor{LightGray!70}131.0346 & \cellcolor{LightGray!70}6.1609& \cellcolor{LightGray!70}\textbf{67.0355}  & \cellcolor{LightGray!70}\textbf{0.6701} & \cellcolor{LightGray!70}\underline{0.6800} \\
\bottomrule
\end{tabular}
\end{table*}

\begin{table*}[!h]
\renewcommand{\arraystretch}{1.2}
\centering
\fontsize{6.5pt}{7.5pt}\selectfont
\setlength{\tabcolsep}{2.5pt}
\caption{4$\times$ performance comparison of evaluated models on the WorldStrat dataset (160$\rightarrow$640). The top three performances of each metric are marked in \textbf{bold}, \underline{underline}, \textit{italic} respectively.
}
\label{table:performance-worldstrat}
\begin{tabular}{l|l|ccccccccc}
\toprule
{\textbf{Type}} & {\textbf{Model}}   & \textbf{PSNR$\uparrow$}&  \textbf{SSIM$\uparrow$} &  \textbf{LPIPS$\downarrow$}  &  \textbf{DISTS$\downarrow$}  &  \textbf{FID$\downarrow$}  &  \textbf{NIQE$\downarrow$}  &  \textbf{MUSIQ$\uparrow$}  &  \textbf{MANIQA$\uparrow$}  &  \textbf{CLIPIQA$\uparrow$}   \\ 
\midrule
\multirow{3}{*}{Fidelity-based SR}
& HAT-L~\cite{chen2023hat} &21.2238 & 0.6480&0.3468 & 0.2108 & 145.3798  &9.0026 & 30.6655 & 0.2844 & 0.2339 \\
& PiSA-SR-PSNR~\cite{sun2024pixel} & 24.4312 & \textbf{0.7271} &0.3312 & 0.2283 & 142.8870 & 10.5327 &29.6462 & 0.2994 & 0.2509 \\
& SwinIR~\cite{liang2021swinir} & 18.0937 &0.6136 & 0.3451 & 0.2279 & 180.4954  & 8.2607 & 27.1954 &0.2104& 0.2355 \\
\midrule
\multirow{5}{*}{Perception-based SR}
& DiffBIR~\cite{lin2024diffbir} & 20.6485 & 0.5150& 0.6781& 0.3712 &227.6764 & \underline{7.5514} & \textit{53.5666} & \underline{0.5475} & \underline{0.6075} \\
& OSEDiff~\cite{wu2024one} & 25.9716 & 0.6316&  0.4460 &0.2562& 176.2589  &  8.6342 & 46.5092 & 0.4988 & 0.5096\\
& HAT-GAN~\cite{chen2023hat} & \underline{27.0796} & \underline{0.7241} & \underline{0.3199} & \underline{0.1978}& \textit{137.4441}  & 9.8474 & 30.3741 & 0.3587 & 0.2623 \\
& PiSA-SR~\cite{sun2024pixel} & 23.9304&0.6179& 0.4581& 0.2748 & 170.0426 & \textbf{7.2214} & 48.6414 & 0.5152 &0.5010 \\
& SwinIR (Real-ISR)~\cite{liang2021swinir} &  \textbf{27.9062} & 0.7120 & 0.3473& \textit{0.2074} &149.4428 & 10.1237 &32.9484& 0.3497 & 0.3060 \\
\midrule
\multirow{2}{*}{Expert Aerial SR}
& HAUNet~\cite{10145829} &\textit{26.1895} & \textit{0.7143}& \textbf{0.3141} &\textbf{0.1976}& \textbf{128.2747} & 10.6318& 28.4401 & 0.3129 &0.2701 \\
& TransENet~\cite{9654169} & 24.4879 & 0.6943 &\textit{0.3270}& 0.2106 & \underline{133.6959} & \textit{7.7765} & 27.8965 & 0.3152 & 0.2246 \\
\midrule
\multirow{3}{*}{Agentic System}
& AgenticIR~\cite{zhu2024intelligent} & 19.5883 & 0.5188 &0.6716 & 0.3686 & 224.8042&8.7079 & \underline{54.0649} & \textit{0.5166} & \textit{0.5402} \\
& \cellcolor{LightGray!70}\textbf{4KAgent (ExpSR-s4-F)} & \cellcolor{LightGray!70}22.3529 & \cellcolor{LightGray!70}0.6470 &  \cellcolor{LightGray!70}0.3702 & \cellcolor{LightGray!70}0.2324 & \cellcolor{LightGray!70}166.2731 & \cellcolor{LightGray!70}9.5364& \cellcolor{LightGray!70}34.3698 & \cellcolor{LightGray!70}0.3011& \cellcolor{LightGray!70}0.2875 \\
& \cellcolor{LightGray!70}\textbf{4KAgent (ExpSR-s4-P)} & \cellcolor{LightGray!70}20.1510 & \cellcolor{LightGray!70}0.5379 & \cellcolor{LightGray!70}0.6363& \cellcolor{LightGray!70}0.3664 & \cellcolor{LightGray!70}223.4866 & \cellcolor{LightGray!70}8.5528 & \cellcolor{LightGray!70}\textbf{56.8421} & \cellcolor{LightGray!70}\textbf{0.5547} & \cellcolor{LightGray!70}\textbf{0.6236} \\
\bottomrule
\end{tabular}
\end{table*}

\begin{table*}[!h]
\renewcommand{\arraystretch}{1.2}
\centering
\fontsize{6.5pt}{7.5pt}\selectfont
\setlength{\tabcolsep}{2.5pt}
\caption{4$\times$ performance comparison of evaluated models on the DOTA dataset (512$\rightarrow$2048). The top three performances of each metric are marked in \textbf{bold}, \underline{underline}, \textit{italic} respectively.}
\label{table:performance-dota-512}
\begin{tabular}{l|l|ccccccccc}
\toprule
{\textbf{Type}} & {\textbf{Model}} & \textbf{PSNR$\uparrow$} & \textbf{SSIM$\uparrow$} & \textbf{LPIPS$\downarrow$} & \textbf{DISTS$\downarrow$} & \textbf{FID$\downarrow$} & \textbf{NIQE$\downarrow$} & \textbf{MUSIQ$\uparrow$} & \textbf{MANIQA$\uparrow$} & \textbf{CLIPIQA$\uparrow$} \\
\midrule
\multirow{3}{*}{Fidelity-based SR}
& HAT-L~\cite{chen2023hat} & \textbf{38.4856} & \textbf{0.9101} & \textbf{0.1956} & \underline{0.1007} & \underline{0.2722} & 6.8076 & 39.5562 & 0.5006 & 0.3408 \\
& PiSA-SR-PSNR~\cite{sun2024pixel} & 31.5336 & 0.8519 & 0.3183 & 0.1797 & 40.7879 & 7.0981 & 38.6277 & 0.4226 & 0.2444 \\
& SwinIR~\cite{liang2021swinir} & 33.7463 & 0.8759 & 0.2990 & 0.1809 & 15.6964 & 6.7591 & 43.3724 & 0.4200 & 0.2847 \\
\midrule
\multirow{5}{*}{Perception-based SR}
& DiffBIR~\cite{lin2024diffbir} & 26.3032 & 0.6490 & 0.4035 & 0.1953 & 76.3522 & \textit{4.0360} & \textbf{57.2133} & \textbf{0.6265} & \textbf{0.7306} \\
& OSEDiff~\cite{wu2024one} & 28.5768 & 0.7567 & 0.3632 & 0.2127 & 70.4222 & 4.1286 & 51.4468 & 0.5871 & \underline{0.6907} \\
& HAT-GAN~\cite{chen2023hat} & 31.2735 & 0.8408 & 0.2720 & 0.1527 & 47.0606 & 4.6791 & 47.6011 & 0.5508 & 0.3551 \\
& PiSA-SR~\cite{sun2024pixel} & 27.6765 & 0.7303 & 0.3499 & 0.2175 & 54.5844 & \textbf{3.8036} & \textit{51.7604} & \textit{0.6135} & 0.6353 \\
& SwinIR (Real-ISR)~\cite{liang2021swinir} & 31.6933 & 0.8423 & 0.2564 & 0.1453 & 37.3714 & 4.1055 & 48.3376 & 0.5561 & 0.4038 \\
\midrule
\multirow{2}{*}{Expert Aerial SR}
& HAUNet~\cite{10145829} & \underline{38.2237} & \underline{0.9075} & \underline{0.2002} & \textbf{0.0974} & \textit{0.2984} & 6.6907 & 38.8776 & 0.4926 & 0.3471 \\
& TransENet~\cite{9654169} & 35.9776 & 0.8824 & 0.2431 & \textit{0.1267} & \textbf{0.2137} & 6.3886 & 34.8775 & 0.4345 & 0.2942 \\
\midrule
\multirow{3}{*}{Agentic System}
& AgenticIR~\cite{zhu2024intelligent} & 21.4719 & 0.7284 & 0.4157 & 0.2167 & 77.7286 & 4.7902 & 49.5913 & 0.5496 & 0.4853 \\
& \cellcolor{LightGray!70}\textbf{4KAgent (ExpSR-s4-F)} & \cellcolor{LightGray!70}\textit{36.7655} & \cellcolor{LightGray!70}\textit{0.9017} & \cellcolor{LightGray!70}\textit{0.2343} & \cellcolor{LightGray!70}0.1283 & \cellcolor{LightGray!70}11.0017 & \cellcolor{LightGray!70}7.0361 & \cellcolor{LightGray!70}38.7286 & \cellcolor{LightGray!70}0.4738 & \cellcolor{LightGray!70}0.3493 \\
& \cellcolor{LightGray!70}\textbf{4KAgent (ExpSR-s4-P)} & \cellcolor{LightGray!70}28.4281 & \cellcolor{LightGray!70}0.7513& \cellcolor{LightGray!70}0.3440 & \cellcolor{LightGray!70}0.2181 & \cellcolor{LightGray!70}41.1425& \cellcolor{LightGray!70}\underline{3.9267}& \cellcolor{LightGray!70}\underline{52.1735} & \cellcolor{LightGray!70}\underline{0.6264} & \cellcolor{LightGray!70}\textit{0.6608} \\
\bottomrule
\end{tabular}
\end{table*}

\begin{table*}[!h]
\renewcommand{\arraystretch}{1.2}
\centering
\fontsize{6.5pt}{7.5pt}\selectfont
\setlength{\tabcolsep}{2.5pt}
\caption{16$\times$ performance comparison of evaluated models on the DOTA dataset (4K resolution). The top three performances of each metric are marked in \textbf{bold}, \underline{underline}, \textit{italic} respectively.}
\label{table:performance-dota-4k}
\begin{tabular}{l|l|ccccccccc}
\toprule
{\textbf{Type}} & {\textbf{Model}}   & \textbf{PSNR$\uparrow$}&  \textbf{SSIM$\uparrow$} &  \textbf{LPIPS$\downarrow$}  &  \textbf{DISTS$\downarrow$}  &  \textbf{FID$\downarrow$}  &  \textbf{NIQE$\downarrow$}  &  \textbf{MUSIQ$\uparrow$}  &  \textbf{MANIQA$\uparrow$}  &  \textbf{CLIPIQA$\uparrow$}   \\ 
\midrule
\multirow{3}{*}{Fidelity-based SR}
& HAT-L~\cite{chen2023hat} & \textbf{23.9586} & \textbf{0.6362} & 0.6471 & 0.3219 & \textbf{82.7644} & 9.0807 & 31.5394 & 0.2565 & 0.3062 \\
& PiSA-SR-PSNR~\cite{sun2024pixel} & 22.6265 & 0.5994 & 0.7279 & 0.3368 & 110.5112 & 9.2583 & 24.0154 & 0.2066 & 0.1766 \\
& SwinIR~\cite{liang2021swinir} & 22.9425 & 0.6095 & 0.6860 & 0.3815 & 162.1128 & 9.7613 & 37.0268 & 0.2792 & 0.2814 \\
\midrule
\multirow{5}{*}{Perception-based SR}
& DiffBIR~\cite{lin2024diffbir} & 21.4093 & 0.4612 & 0.5595 &\textbf{0.2214} & 114.8595 & \textbf{3.4046} & \textbf{57.5771} & \underline{0.5030} & \textbf{0.7588} \\
& OSEDiff~\cite{wu2024one} & 22.0602 & 0.5544 & \underline{0.5450} & 0.2647 & 107.3622 & 4.1667 & 52.5278 & \textit{0.4430} & \underline{0.7287} \\
& HAT-GAN~\cite{chen2023hat} & 21.7525 & 0.5901 & 0.5590 & 0.2668 & 139.2047 & 5.4465 & 45.6411 & 0.2791 & 0.3448 \\
& PiSA-SR~\cite{sun2024pixel} & 22.1022 & 0.5761 & 0.5552 & 0.2517 & 100.3336 & 4.2723 & 48.0151 & 0.3194 & 0.5972 \\
& SwinIR (Real-ISR)~\cite{liang2021swinir} & 21.6770 & 0.5731 & \textbf{0.5431} & \textit{0.2431} & 129.7745 & \textit{3.7377} & 50.5413 & 0.3033 & 0.4885 \\
\midrule
\multirow{2}{*}{Expert Aerial SR}
& HAUNet~\cite{10145829} & \underline{23.6649} & \underline{0.6268} & 0.6922 & 0.3304 & \underline{86.2487} & 9.0018 & 26.2489 & 0.2207 & 0.2567 \\
& TransENet~\cite{9654169}  & 22.9690 & 0.5992 & 0.7449 & 0.3531 & \textit{97.5895} & 7.6931 & 21.3092 & 0.1903 & 0.1765 \\
\midrule
\multirow{3}{*}{Agentic System}
& AgenticIR~\cite{zhu2024intelligent} & 17.8736 & 0.4675 & 0.5928 & 0.2451 & 135.6437 & 3.8950 & \textit{54.3685} & 0.4301 & 0.6551 \\
& \cellcolor{LightGray!70}\textbf{4KAgent (Aer4K-F)} & \cellcolor{LightGray!70}\textit{23.4348} & \cellcolor{LightGray!70}\textit{0.6255} & \cellcolor{LightGray!70}0.6520 & \cellcolor{LightGray!70}0.3312 & \cellcolor{LightGray!70}105.6710 & \cellcolor{LightGray!70}9.0064 & \cellcolor{LightGray!70}33.6645 & \cellcolor{LightGray!70}0.2725 & \cellcolor{LightGray!70}0.3314 \\
& \cellcolor{LightGray!70}\textbf{4KAgent (Aer4K-P)} & \cellcolor{LightGray!70}21.9826 & \cellcolor{LightGray!70}0.5515& \cellcolor{LightGray!70}\textit{0.5525}& \cellcolor{LightGray!70}\underline{0.2415} & \cellcolor{LightGray!70}112.2518& \cellcolor{LightGray!70}\underline{3.7230}& \cellcolor{LightGray!70}\underline{55.7730} & \cellcolor{LightGray!70}\textbf{0.5175} & \cellcolor{LightGray!70}\textit{0.7159} \\
\bottomrule
\end{tabular}
\end{table*}

\begin{figure*}[!h]
\centering
\footnotesize
\includegraphics[width=1.0\textwidth]{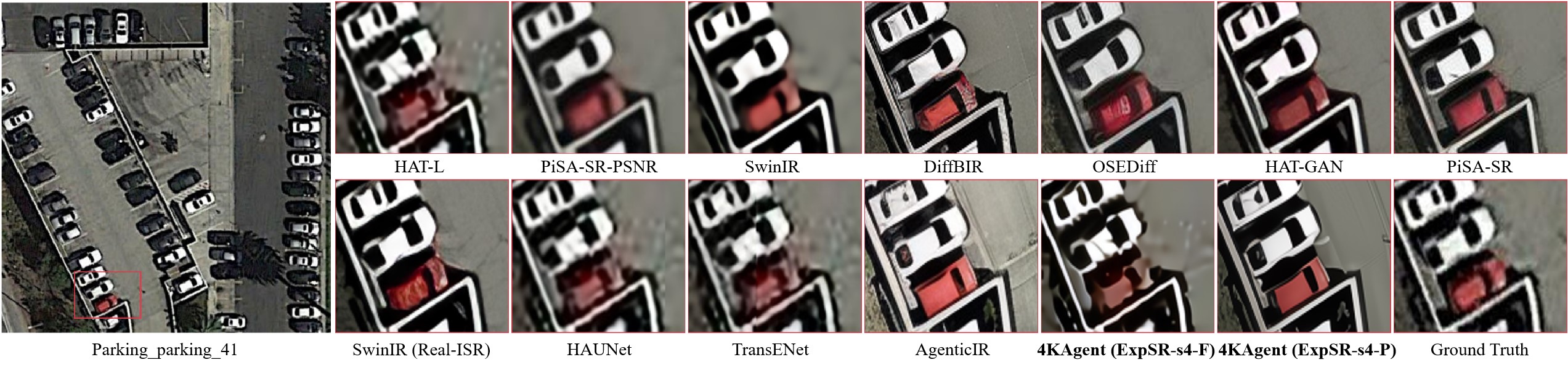}\\
\includegraphics[width=1.0\textwidth]{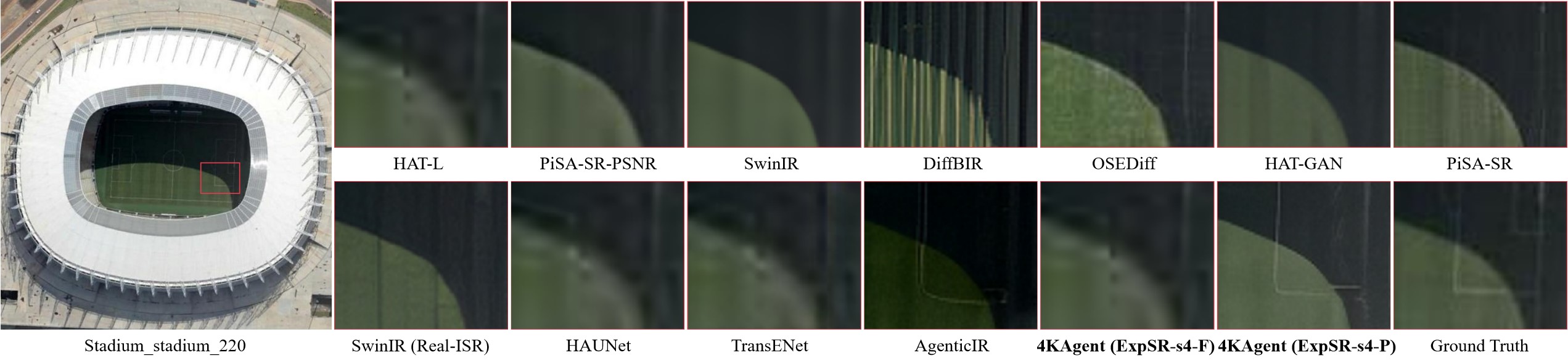}\\
\includegraphics[width=1.0\textwidth]{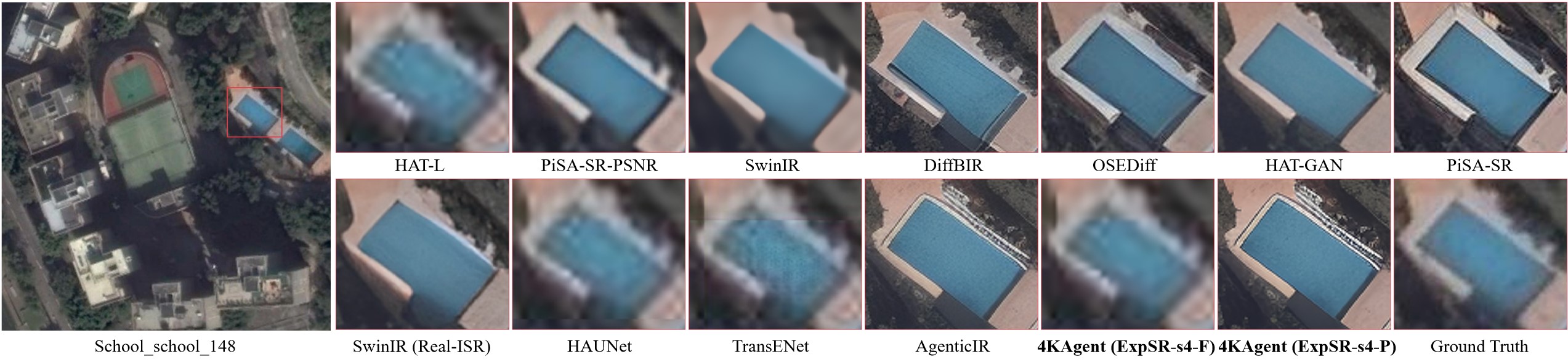}
\vspace{-5pt}
\caption{Visual comparison on AID dataset (160$\rightarrow$640).}
\label{fig:visual_aerial_aid_4x}
\end{figure*}

\begin{figure*}[!h]
\centering
\footnotesize
\includegraphics[width=1.0\textwidth]{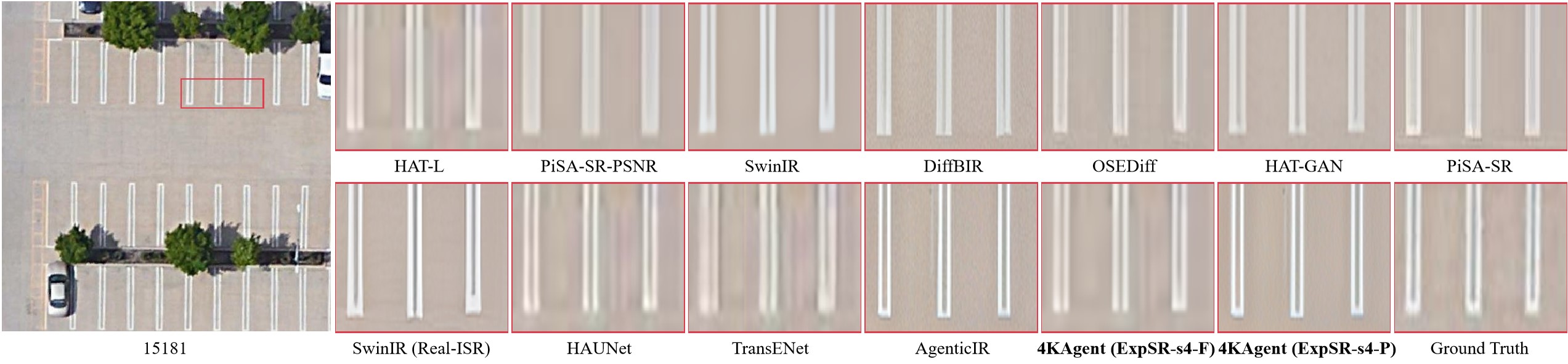} \\
\includegraphics[width=1.0\textwidth]{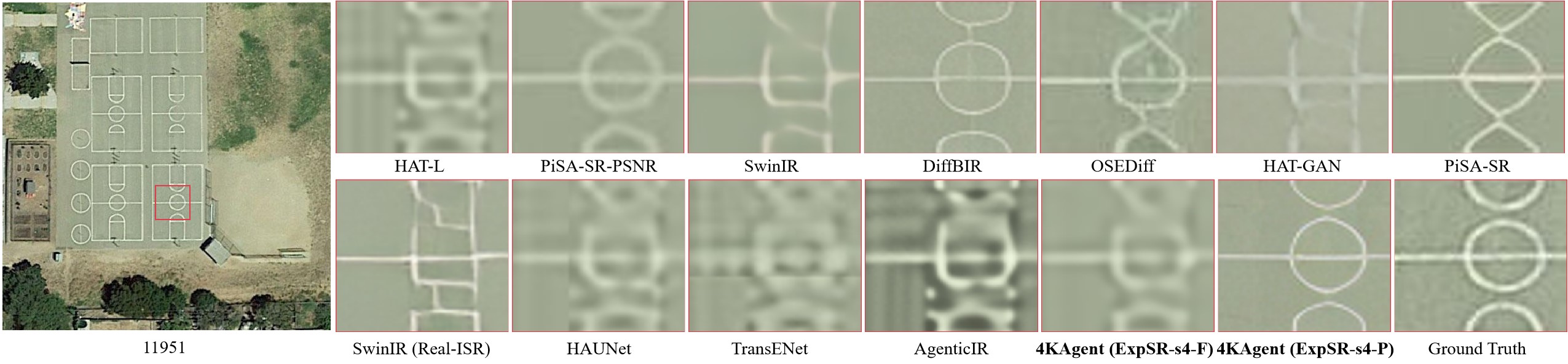} \\
\includegraphics[width=1.0\textwidth]{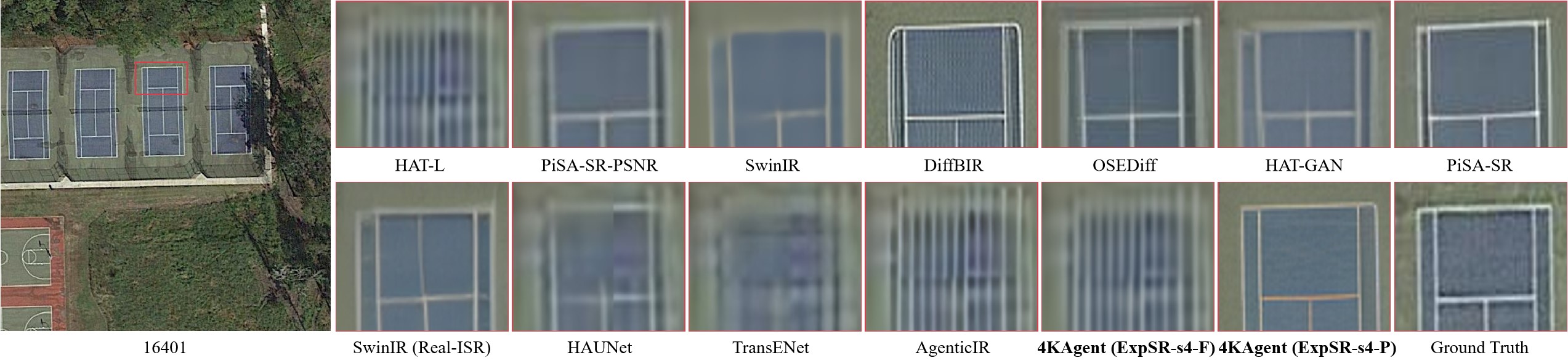}
\caption{Visual comparison on DIOR dataset (128\textrightarrow512).}
\label{fig:visual_aerial_dior_4x}
\end{figure*}

\begin{figure*}[!h]
\centering
\footnotesize
\includegraphics[width=1.0\textwidth]{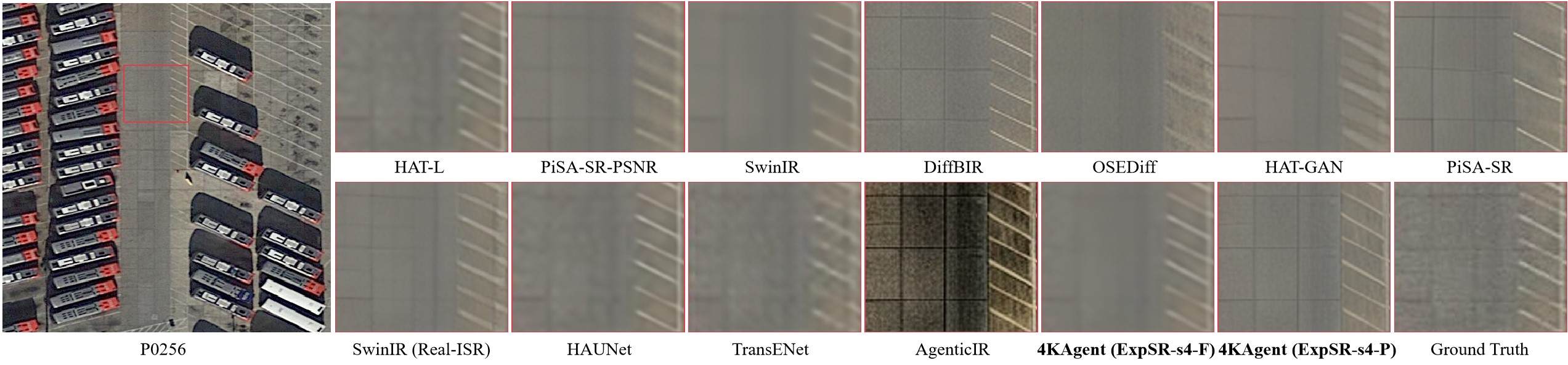} \\
\includegraphics[width=1.0\textwidth]{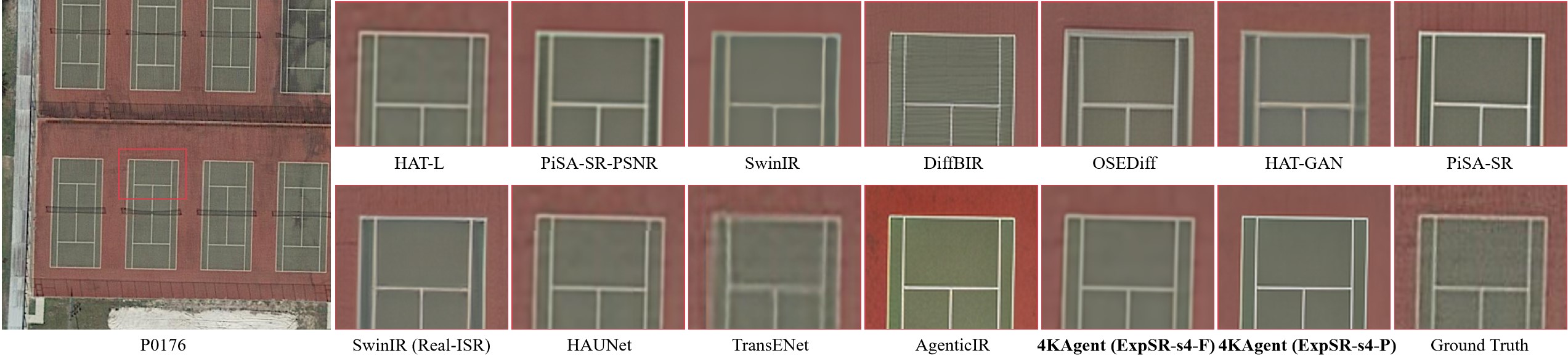} \\
\includegraphics[width=1.0\textwidth]{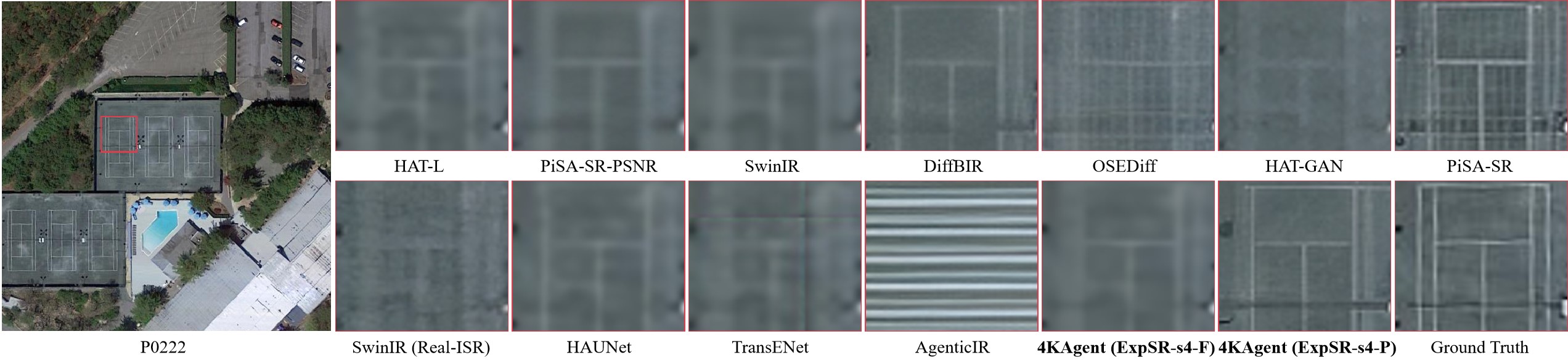}
\caption{Visual comparison on DOTA dataset (128$\rightarrow$512).
}
\label{fig:visual_aerial_dota_4x}
\end{figure*}

\begin{figure*}[!h]
\centering
\footnotesize
\includegraphics[width=1.0\textwidth]{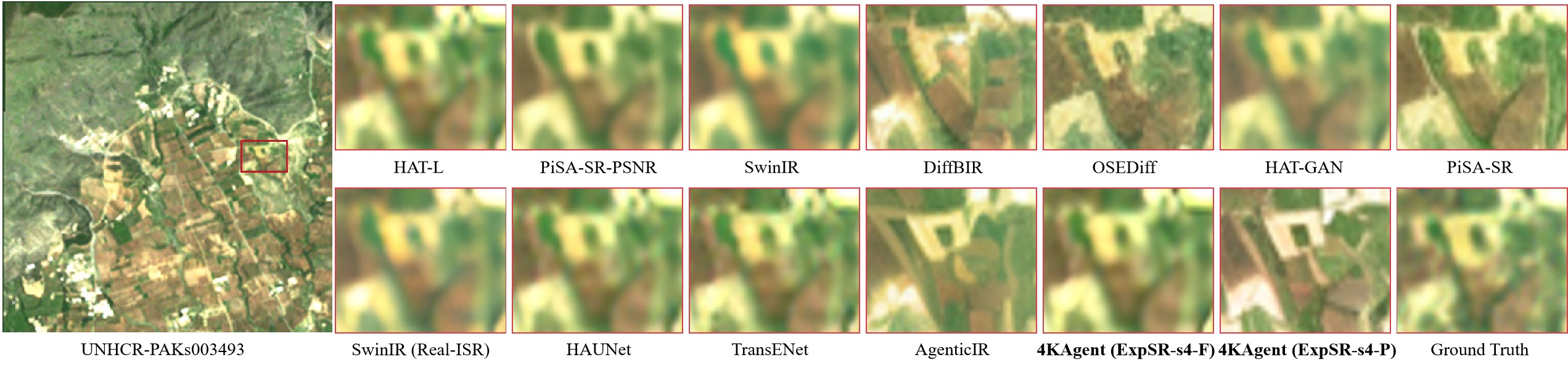} \\
\includegraphics[width=1.0\textwidth]{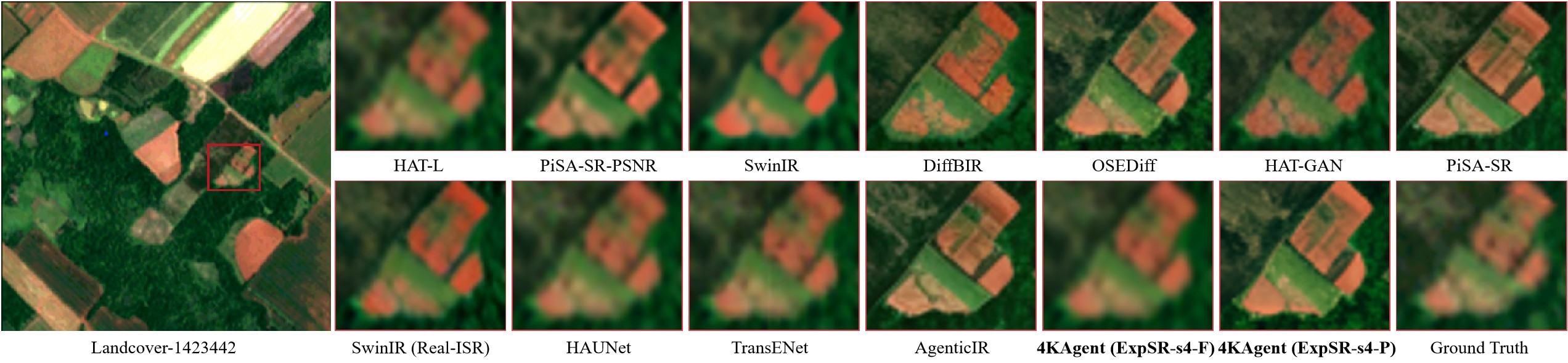} \\
\includegraphics[width=1.0\textwidth]{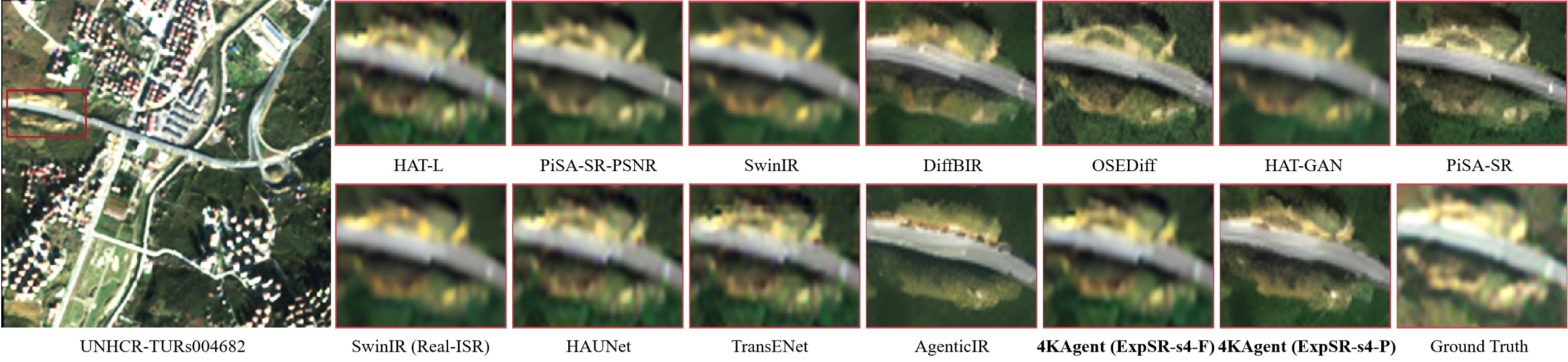}
\caption{Visual comparison on WorldStrat dataset (160$\rightarrow$640).
}
\label{fig:visual_aerial_worldstrat}
\end{figure*}

\begin{figure*}[!h]
\centering
\footnotesize
\includegraphics[width=1.0\textwidth]{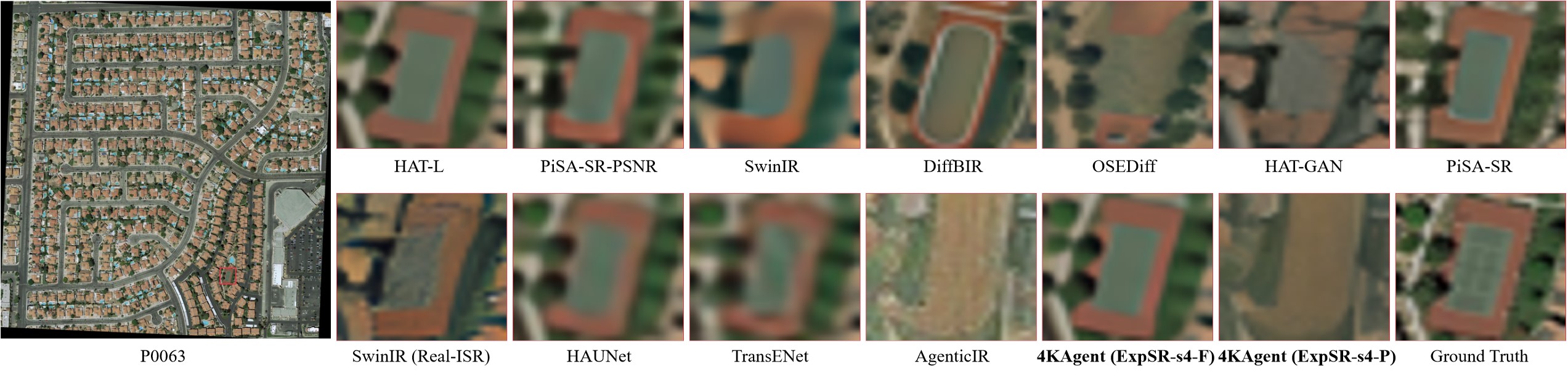} \\
\includegraphics[width=1.0\textwidth]{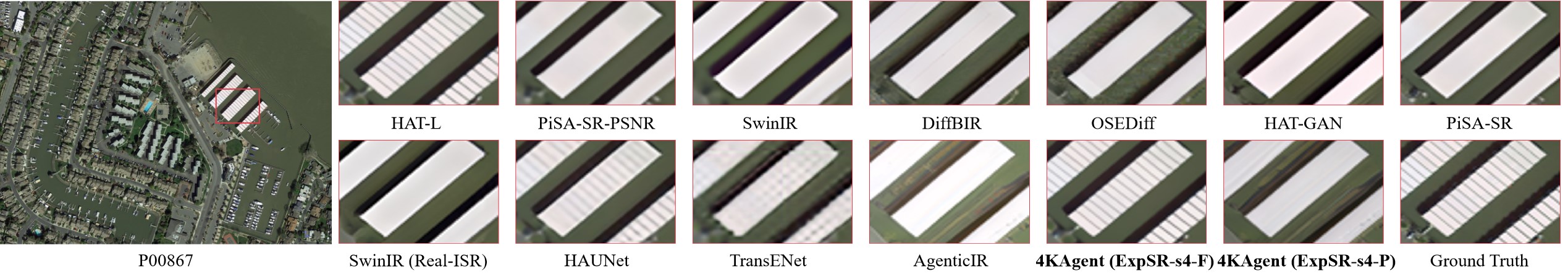} \\
\includegraphics[width=1.0\textwidth]{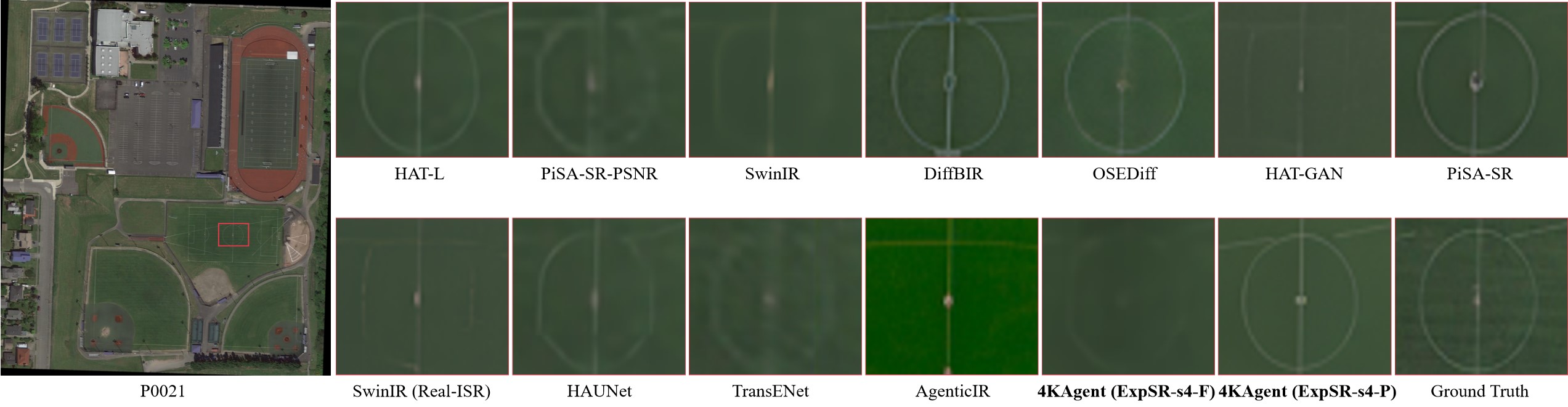} \\
\includegraphics[width=1.0\textwidth]{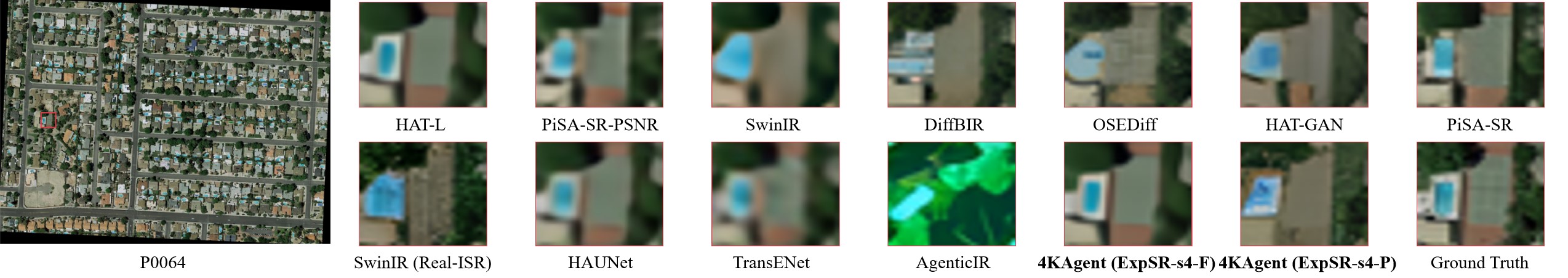} \\
\includegraphics[width=1.0\textwidth]{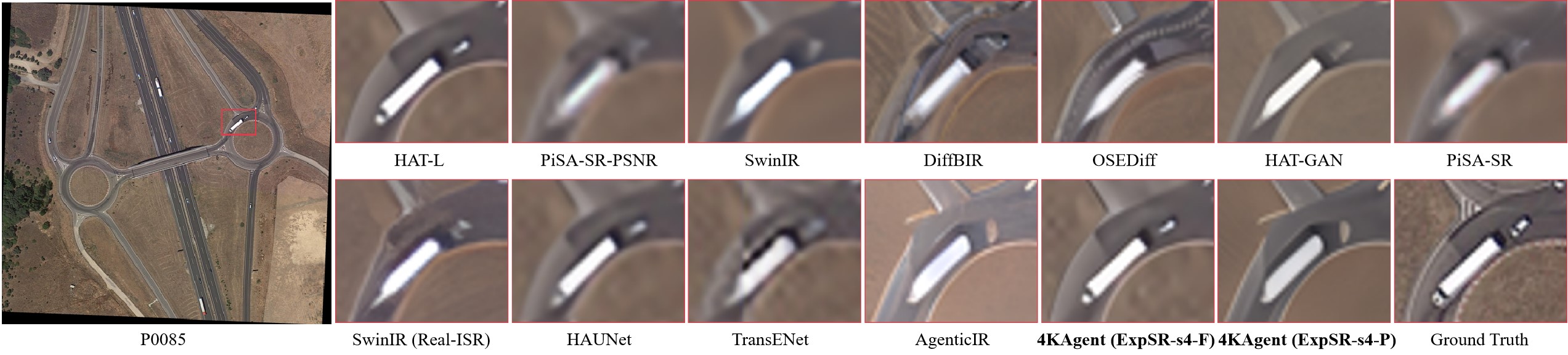}\\
\includegraphics[width=1.0\textwidth]{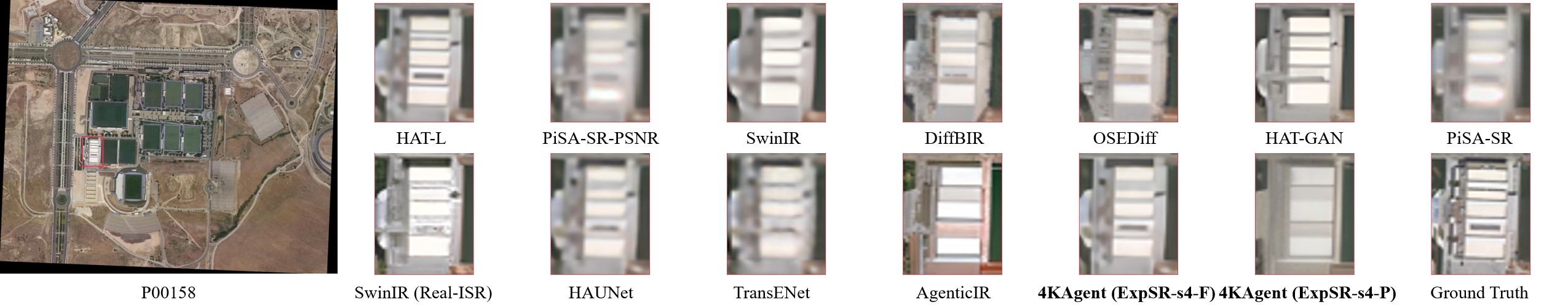}
\caption{Visual comparison on DOTA dataset (4K upscaling).}
\label{fig:visual_aerial_dota_16x}
\end{figure*}

\paragraph{Qualitative Comparison}
\cref{fig:visual_aerial_aid_4x,fig:visual_aerial_dior_4x,fig:visual_aerial_dota_4x,fig:visual_aerial_worldstrat,fig:visual_aerial_dota_16x} present a comprehensive visual comparison of all evaluated models across the tested datasets of 4$\times$ AID, 4$\times$ DIOR, 4$\times$ DOTA, 4$\times$ WorldStrat, and 16$\times$ DOTA. \textbf{Firstly}, 4KAgent with the perception preference consistently delivers superior perceptual quality on low-resolution SR datasets, as demonstrated in~\cref{fig:visual_aerial_aid_4x,fig:visual_aerial_dior_4x,fig:visual_aerial_dota_4x,fig:visual_aerial_worldstrat}. In contrast, 4KAgent with the fidelity preference excels on high-resolution 4K SR datasets, producing the most faithful reconstructions in~\cref{fig:visual_aerial_dota_16x}. \textbf{Secondly}, 4KAgent exhibits a clear advantage in reconstructing fine structures such as lines and patterns, as evident in~\cref{fig:visual_aerial_dior_4x,fig:visual_aerial_dota_4x}. \textbf{Finally}, in the challenging cross-sensor super-resolution scenario of WorldStrat, where LR and HR images originate from different sensors. 4KAgent with the perception preference still maintains promising visual performance, as shown in~\cref{fig:visual_aerial_worldstrat}. This demonstrates the robustness of 4KAgent across both resolution scales and sensor domains.

\paragraph{Discussions}
These results across fidelity and perception metrics, combined with qualitative visual comparisons, provide several key insights into the advantages of 4KAgent. \textbf{First}, the consistent top-tier performance of 4KAgent with fidelity-based profile across a wide range of datasets and scaling factors suggests that the agent's analytical pipeline and adaptive control provide more precise reconstruction than traditional feedforward models. Unlike conventional SR networks that are typically optimized for either low-level fidelity or high-level realism, 4KAgent’s architecture decouples these objectives through specialized profiles, allowing it to excel in both domains without compromise. \textbf{Second}, the perceptual strength of 4KAgent is reflected not only in numerical scores but also in its sharper textures, reduced artifacts, and more semantically coherent outputs in qualitative results, demonstrating the value of integrating agentic reasoning with perceptual priors, especially in real-world datasets like WorldStrat. \textbf{Finally}, the margin by which both 4KAgent variants outperform AgenticIR, demonstrating the superiority of 4KAgent in terms of agentic system. Our contributions in the design of specialized profiles, adaptive modulation, and perceptual alignment mechanisms are crucial to bridging the gap between task generality and SR specialization. 
Together, these findings indicate that agentic architectures, when properly aligned with SR objectives, enable scalable, generalizable, and controllable super-resolution across diverse remote sensing domains.

\clearpage

\subsection{Fluorescence Microscopic Image Super-Resolution}
\label{ssec:sci-sr-ii-micro}
Confocal fluorescence microscopy is one of the most accessible and widely used techniques for studying cellular and subcellular structures~\cite{hickey2021fluorescence,yang2021advancing}. It builds a sharp image by using either a single pinhole to scan point-by-point or an array of pinholes on a spinning disk to scan multiple points simultaneously to reject out-of-focus light, offering molecular specificity and 3D sectioning capabilities. However, it is constrained by diffraction\mbox{-}limited resolution down to 200 nm under visible light~\cite{thorley2014super}. Meanwhile, high\mbox{-}intensity illumination required for improved resolution leads to photobleaching and phototoxicity, limiting live\mbox{-}cell imaging duration and data throughput~\cite{tosheva2020between}. Deep learning\mbox{-}based single\mbox{-}image super\mbox{-}resolution (SISR) methods have shown great promise in recovering high\mbox{-}frequency details from lower\mbox{-}resolution inputs in biological microscopy, overcoming some limitations of hardware\mbox{-}based SR techniques~\cite{mannam2021deep}, despite the scarcity of large, publicly available fluorescence microscopy datasets. To extend the evaluation of our 4KAgent on this scientific application of different modalities, we conducted experiments on a representative dataset against baselines from major SISR families.

\paragraph{Settings.}
We evaluate 4KAgent on \emph{SR\mbox{-}CACO\mbox{-}2} benchmark dataset~\cite{belharbi2024sr}, which contains 2,200 unique images of the Caco-2 human epithelial cell line, labeled with three distinct fluorescent markers: Survivin (CELL0), E-cadherin / Tubulin (CELL1), and Histone H2B (CELL2), at $\times2$ (256$\rightarrow$512), $\times4$ (128$\rightarrow$512), and $\times8$ (64$\rightarrow$512) scales. To generate high-resolution images, each tile was scanned with a 1024$\times$1024 pixel resolution, and 8 scans were captured and then averaged together to reduce noise. Meanwhile, low-resolution images were captured directly by the microscope at three different scales without averaging. The full dataset contains 9,937 patches for each cell extracted from scanning confocal volumes, with tiles 9, 10, 14, 20 used as the test set. In our experiments, we randomly sampled 100 patches from each marker category in the test set at three super-resolution scales.

We evaluate 4KAgent with the \textbf{ExpSR-s2-F} profile, \textbf{ExpSR-s4-F} profile, and \textbf{ExpSR-s8-F} profile, considering the demands and requirements of the microscopy image super-resolution task. We benchmark 4KAgent against 15 representative SISR models, broadly spanning pre-upsampling, post-upsampling, iterative up-and-down sampling and progressive upsampling SR methods. Each model has been trained on the SR-CACO-2 training set before deployment. Encompassing a wide spectrum of upsampling strategies, this rigorous benchmark ensures a comprehensive comparison between our 4KAgent and other specialized and general-purpose SR methods, assessing 4KAgent's blind inference performance on the novel microscopy data domain.

\paragraph{Quantitative Comparison}
All quantitative results are in~\cref{tab:srcaco-test-perf-roi}. PSNR, SSIM, and NRMSE were selected as criteria. We noticed that the background of a cell constituted a significant proportion of an image, as was also reported in the original SR-CACO-2 benchmark. Because including the non-informative dark background in evaluation can lead to inflated and biased performance metrics, we adopted the masking strategy described in~\cite{belharbi2024sr} to define our Regions-of-Interest (ROIs) and calculated performance metrics based only on these areas. Across different scales and cell types, 4KAgent with \textbf{Fidelity} preference consistently achieves top performance in \textbf{pixel-level reconstruction} in ROI. The superior result on PSNR, SSIM, and NRMSE confirms 4KAgent's effectiveness in reconstructing fine fluorescence-labeled structures and low-level pixel fidelity, under various downsampling conditions. Furthermore, when compared with ENLCN, one of the most competitive methods, our 4KAgent with fidelity mode consistently exhibits a clear advantage in all numerical metrics, underscoring its ability to handle the blind super-resolution task for real-world microscopy data.
\renewcommand{\arraystretch}{1.1}
\begin{table*}[!t]
\centering
\caption{Performance comparison of evaluated models on the selected sr-caco-2 test set \textbf{on ROI only, \ie, cells}. The top three performances of each metric are marked in \textbf{bold}, \underline{underline}, and \textit{italic}.}
\label{tab:srcaco-test-perf-roi}
\setlength{\tabcolsep}{3pt}
\resizebox{\textwidth}{!}{
\small
% \begin{tabular}{lc|lc|*{3}{c|}gc*{3}{c|}gc*{3}{c|}g}
\begin{tabular}{lc|lc|
    *{3}{c|}cc
    *{3}{c|}cc
    *{3}{c|}c
}
\toprule
  && && \multicolumn{4}{c}{\textbf{PSNR${\uparrow}$}} & & \multicolumn{4}{c}{\textbf{NRMSE${\downarrow}$}} & & \multicolumn{4}{c}{\textbf{SSIM${\uparrow}$}}  \\
\textbf{SISR Methods} && Scale && \cellzero & \cellone & \celltwo & Mean && \cellzero & \cellone & \celltwo & Mean  && \cellzero & \cellone & \celltwo & Mean  \\ \midrule
\multirow{3}{*}{Bicubic }
&&\xtwo   && 34.93 &  32.61  &  30.24  &  32.59  
          && 0.0887 &  0.0681 & 0.0661 & 0.0743
          && 0.7899 &  0.7826 & 0.7038 & 0.7588  \\
&&\xfour  && 35.08 &  31.99  & 30.43  & 32.50 
          && 0.0793 & 0.0690 & 0.0641 & 0.0708
          && 0.8411 & 0.7998 & 0.7718 & 0.8042  \\
&&\xeight && 32.01 & 28.77 &  26.27 & 29.02    
          && 0.1311 & 0.1071 & 0.1240 & 0.1207
          &&  0.7280 & 0.6677 & 0.6808 & 0.6922
          \\ \hline
\textbf{Pre-upsampling SR} &   \multicolumn{16}{c}{} \\ \hline
\multirow{3}{*}{SRCNN~\cite{dong2014learning}} %{\small \emph{(eccv,2014)}} 
&&\xtwo   && 37.04 & 34.47 & 33.03 & 34.85    
          && 0.0610 & 0.0516 & 0.0471 & 0.0532   
          && 0.8733 & 0.8566 & 0.8283 & 0.8527 
          \\
&&\xfour  && 35.39 & 32.73 & 31.48 & 33.20 
          && 0.0704 & 0.0610 & 0.0563 & 0.0626   
          && 0.8707 & 0.8193 & 0.8104 & 0.8335 
          \\
&&\xeight  && 32.52 & 29.16 & 26.53 & 29.41
           && 0.1074 & 0.0924 & 0.1143 & 0.1047 && 0.8117 & 0.7219 & 0.7207 & \underline{0.7514}
           \\ \hline
\multirow{3}{*}{VDSR~\cite{kim2016accurate}} %{\small \emph{(cvpr,2016)}} 
&&\xtwo    && 37.50 & 34.29 & 33.00 & 34.93
           && 0.0602 & 0.0554 & 0.0479 & 0.0545
           && 0.8921 & 0.8608 & 0.8385 & \underline{0.8638}
           \\
&&\xfour   && 36.18 & 32.52 & 31.44 & 33.38
           && 0.0663 & 0.0638 & 0.0571 & 0.0624
           && 0.8777 & 0.8218 & 0.8180 & \underline{0.8392}
           \\
&&\xeight  && 32.03 & 28.80 & 26.42 & 29.08
           && 0.1307 & 0.1062 & 0.1237 & 0.1202
           && 0.7291 & 0.6712 & 0.6877 & 0.6960 \\ \hline
\multirow{3}{*}{DRRN~\cite{tai2017image}} %{\small \emph{(cvpr,2017)}} 
&&\xtwo    && 37.35 & 34.23 & 33.08 & 34.89
           && 0.0609 & 0.0555 & 0.0475 & 0.0546 && 0.8917 & 0.8597 & 0.8386 & \textit{0.8634}
           \\
&&\xfour   && 36.00 & 32.50 & 31.43 & 33.31
           && 0.0678 & 0.0637 & 0.0570 & 0.0628  && 0.8772 & 0.8216 & 0.8167 & 0.8385 \\
&&\xeight  && 31.92 & 28.31 & 26.39 & 28.88
           && 0.1310 & 0.1096 & 0.1230 & 0.1212
           && 0.7290 & 0.6583 & 0.6860 & 0.6911
           \\ \hline
\multirow{3}{*}{MemNet~\cite{tai2017memnet}} %{\small \emph{(iccv,2017)}} 
&&\xtwo    && 35.69 & 33.40 & 30.81 & 33.30
           && 0.0776 & 0.0557 & 0.0607 & 0.0647 && 0.8295 & 0.8280 & 0.7759 & 0.8111  \\
&&\xfour   && 34.61 & 32.48 & 30.26 & 32.45
           && 0.0808 & 0.0610 & 0.0654 & 0.0690  && 0.8465 & 0.8067 & 0.7651 & 0.8061
           \\
&&\xeight  && 32.00 & 28.76 & 26.52 & 29.09
           && 0.1272 & 0.0993 & 0.1183 & 0.1149
           && 0.7528 & 0.6972 & 0.7102 & 0.7201 \\ \hline
\textbf{Post-upsampling SR} &   \multicolumn{16}{c}{} \\ \hline
\multirow{3}{*}{NLSN~\cite{mei2021image}} %{\small \emph{(cvpr,2021)}} 
&&\xtwo    && 37.57 & 34.31 & 33.14 & 35.01
           && 0.0588 & 0.0527 & 0.0465 & 0.0527
           && 0.8911 & 0.8563 & 0.8375 & 0.8616 \\
&&\xfour   && 36.39 & 32.75 & 31.68 & \underline{33.61}
           && 0.0630 & 0.0600 & 0.0548 & \underline{0.0593} && 0.8754 & 0.8179 & 0.8131 & 0.8355
           \\
&&\xeight  && 32.56 & 29.13 & 26.30 & 29.33
           && 0.1147 & 0.0939 & 0.1205 & 0.1097 && 0.7909 & 0.7092 & 0.6989 & 0.7330
           \\ \hline
\multirow{3}{*}{DFCAN~\cite{qiao2021evaluation}} %{\small \emph{(nat. methods,2021)}} 
&&\xtwo    && 37.21 & 34.20 & 32.74 & 34.72
           && 0.0614 & 0.0561 & 0.0493 & 0.0556
           && 0.8899 & 0.8603 & 0.8375 & 0.8626
           \\
&&\xfour   && 35.92 & 32.49 & 31.29 & 33.23
           && 0.0684 & 0.0653 & 0.0582 & 0.0640
           && 0.8770 & 0.8223 & 0.8174 & \textit{0.8389}
           \\
&&\xeight  && 31.25 & 28.15 & 25.45 & 28.28
           && 0.1344 & 0.1079 & 0.1276 & 0.1233
           && 0.7447 & 0.6749 & 0.6841 & 0.7012
           \\ \hline
\multirow{3}{*}{SwinIR~\cite{liang2021swinir}} %{\small \emph{(iccvw,2021)}} 
&&\xtwo    && 24.55 & 34.48 & 33.08 & 30.71
           && 0.2349 & 0.0527 & 0.0473 & 0.1116
           && 0.3785 & 0.8626 & 0.8385 & 0.6932
           \\
&&\xfour   && 35.93 & 32.66 & 31.57 & 33.39
           && 0.0673 & 0.0618 & 0.0559 & 0.0617 && 0.8772 & 0.8198 & 0.8161 & 0.8377
           \\
&&\xeight  && 31.34 & 28.43 & 25.86 & 28.54
           && 0.1314 & 0.1035 & 0.1230 & 0.1193 && 0.7516 & 0.6838 & 0.6923 & 0.7092
           \\ \hline
\multirow{3}{*}{ENLCN~\cite{xia2022efficient}} %{\small \emph{(aaai,2022)}} 
&&\xtwo    && 37.59 & 34.41 & 33.15 & \underline{35.05}
           && 0.0574 & 0.0518 & 0.0462 & \underline{0.0518}
           && 0.8876 & 0.8569 & 0.8340 & 0.8595
           \\
&&\xfour   && 36.30 & 32.74 & 31.63 & 33.56 
           && 0.0638 & 0.0606 & 0.0553 & \textit{0.0599}
           && 0.8766 & 0.8196 & 0.8148 & 0.8370
           \\
&&\xeight  && 32.69 & 29.28 & 26.31 & \underline{29.43}
           && 0.1108 & 0.0921 & 0.1205 & 0.1078
           && 0.7998 & 0.7109 & 0.6984 & 0.7364
           \\ \hline
\multirow{3}{*}{GRL~\cite{li2023efficient}} %{\small \emph{(cvpr,2023)}} 
&&\xtwo    && 31.28 & 34.54 & 32.81 & 32.88
           && 0.1088 & 0.0522 & 0.0492 & 0.0701
           && 0.8043 & 0.8625 & 0.8337 & 0.8335
           \\
&&\xfour   && 35.76 & 32.81 & 31.48 & 33.35
           && 0.0678 & 0.0581 & 0.0565 & 0.0608
           && 0.8774 & 0.8133 & 0.8144 & 0.8350
           \\
&&\xeight  && 28.14 & 28.93 & 26.22 & 27.76
           && 0.1555 & 0.0953 & 0.1104 & 0.1204
           && 0.7296 & 0.7110 & 0.7197 & 0.7201
           \\ \hline
\multirow{3}{*}{ACT~\cite{yoo2023enriched}} %{\small \emph{(cvpr,2023)}} 
&&\xtwo    && 37.24 & 34.66 & 33.14 & \textit{35.01} 
           && 0.0619 & 0.0496 & 0.0459 & 0.0525
           && 0.8890 & 0.8604 & 0.8288 & 0.8594
           \\
&&\xfour   && 36.17 & 32.76 & 31.56 & 33.50 
           && 0.0652 & 0.0590 & 0.0554 & \textit{0.0599}  && 0.8761 & 0.8134 & 0.8065 & 0.8320
           \\
&&\xeight  && 32.74 & 29.13 & 26.39 & \textit{29.42}
           && 0.1064 & 0.0915 & 0.1152 & \underline{0.1044}
           && 0.8083 & 0.7128 & 0.7063 & 0.7425
           \\ \hline
\multirow{3}{*}{Omni-SR~\cite{wang2023omni}} %{\small \emph{(cvpr,2023)}} 
&&\xtwo    && 37.35 & 34.19 & 33.02 & 34.85
           && 0.0597 & 0.0548 & 0.0475 & 0.0540 && 0.8896 & 0.8562 & 0.8370 & 0.8609  \\
&&\xfour   && 35.86 & 32.53 & 31.49 & 33.29
           && 0.0680 & 0.0635 & 0.0563 & 0.0626
           && 0.8737 & 0.8165 & 0.8117 & 0.8340 \\
&&\xeight  && 30.44 & 28.21 & 25.32 & 27.99 
           && 0.1418 & 0.1075 & 0.1265 & 0.1253
           && 0.7231 & 0.6673 & 0.6808 & 0.6904
           \\ \hline
\textbf{Iterative up-and-down sampling SR} &   \multicolumn{16}{c}{} \\ \hline
\multirow{3}{*}{DBPN~\cite{haris2018deep}} %{\small \emph{(cvpr,2018)}} 
&&\xtwo    && 37.44 & 34.54 & 33.02 & 35.00
           && 0.0588 & 0.0512 & 0.0470 & \textit{0.0523} && 0.8872 & 0.8601 & 0.8339 & 0.8604
           \\
&&\xfour   && 36.22 & 32.85 & 31.64 & \textit{33.57}
           && 0.0638 & 0.0591 & 0.0548 & \underline{0.0593} && 0.8745 & 0.8216 & 0.8091 & 0.8351  \\
&&\xeight  && 32.39 & 28.89 & 26.36 & 29.21
           && 0.1103 & 0.0946 & 0.1157 & \textit{0.1069}
           && 0.8060 & 0.7084 & 0.7149 & \textit{0.7431}
           \\ \hline
\multirow{3}{*}{SRFBN~\cite{li2019feedback}} %{\small \emph{(cvpr,2019)}} 
&&\xtwo    && 36.02 & 33.49 & 31.38 & 33.63
           && 0.0767 & 0.0611 & 0.0576 & 0.0651
           && 0.8319 & 0.8205 & 0.7592 & 0.8038
           \\
&&\xfour   && 35.66 & 32.49 & 31.05 & 33.07
           && 0.0729 & 0.0636 & 0.0592 & 0.0653  && 0.8589 & 0.8131 & 0.7921 & 0.8214
           \\
&&\xeight  && 32.33 & 29.05 & 26.62 & 29.33
           && 0.1243 & 0.1019 & 0.1181 & 0.1148
           && 0.7553 & 0.6869 & 0.7081 & 0.7168  \\ \hline
\textbf{Progressive upsampling SR} &   \multicolumn{16}{c}{} \\ \hline
\multirow{3}{*}{ProSR~\cite{wang2018fully}} %{\small \emph{(cvprw,2018)}} 
&&\xtwo    && 36.92 & 34.66 & 32.80 & 34.79
           && 0.0621 & 0.0494 & 0.0485 & 0.0533
           && 0.8879 & 0.8577 & 0.8385 & 0.8614
           \\
&&\xfour   && 36.11 & 32.71 & 31.61 & 33.48
           && 0.0656 & 0.0614 & 0.0556 & 0.0609
           && 0.8771 & 0.8217 & 0.8164 & 0.8384
           \\
&&\xeight  && 32.13 & 29.43 & 26.36 & 29.31
           && 0.1268 & 0.0909 & 0.1224 & 0.1134  && 0.7504 & 0.7200 & 0.6919 & 0.7208
           \\ \hline
\multirow{3}{*}{MS-LapSRN~\cite{lai2018fast}} %{\small \emph{(tpami,2019)}} 
&&\xtwo    && 32.73 & 32.49 & 28.34 & 31.19
           && 0.1014 & 0.0593 & 0.0805 & 0.0804
           && 0.7957 & 0.8177 & 0.7735 & 0.7956
           \\
&&\xfour   && 30.91 & 31.36 & 30.69 & 30.99 
           && 0.1118 & 0.0672 & 0.0611 & 0.0801
           && 0.8124 & 0.7820 & 0.7858 & 0.7934
           \\
&&\xeight  && 30.67 & 27.64 & 24.68 & 27.66
           && 0.1206 & 0.1056 & 0.1305 & 0.1189
           && 0.7829 & 0.6902 & 0.6649 & 0.7127
           \\ \hline
\textbf{Agentic System} &   \multicolumn{16}{c}{} \\ \hline

\rowcolor{LightGray!70}
\multirow{3}{*}{}
    & \gcell{} & \gcell{\texttt{x2}} & \gcell{} & \gcell{39.92} & \gcell{36.95} & \gcell{33.93} & \gcell{\textbf{36.94}}
    & \gcell{} & \gcell{0.0508} & \gcell{0.0337} & \gcell{0.0426} & \gcell{\textbf{0.0424}}
    & \gcell{} & \gcell{0.9321} & \gcell{0.9105} & \gcell{0.8745} & \gcell{\textbf{0.9057}}
    \\
    \rowcolor{LightGray!70}{\textbf{4KAgent (ExpSR-sN-F)} (N $\in$ [\texttt{2,4,8}])}
    & \gcell{\cellcolor{LightGray!70}}  & \gcell{\texttt{x4}} & \gcell{} & \gcell{41.25} & \gcell{36.86} & \gcell{35.07} & \gcell{\textbf{37.73}}
    & \gcell{} & \gcell{0.0389} & \gcell{0.0318} & \gcell{0.0366} & \gcell{\textbf{0.0358}}
    & \gcell{} & \gcell{0.9555} & \gcell{0.9314} & \gcell{0.9089} & \gcell{\textbf{0.9319}}
    \\
    \rowcolor{LightGray!70}
    & \gcell{\cellcolor{LightGray!70}}  & \gcell{\texttt{x8}} & \gcell{} & \gcell{38.93} & \gcell{33.66} & \gcell{31.99} & \gcell{\textbf{34.86}}
    & \gcell{} & \gcell{0.0532} & \gcell{0.0483} & \gcell{0.0602} & \gcell{\textbf{0.0539}}
    & \gcell{} & \gcell{0.9378} & \gcell{0.9033} & \gcell{0.8929} & \gcell{\textbf{0.9113}}
    \\ \bottomrule

\end{tabular}
}
% \vspace{-1em}
\end{table*}

\begin{figure*}[!h]
\centering
\includegraphics[width=1.0\textwidth]{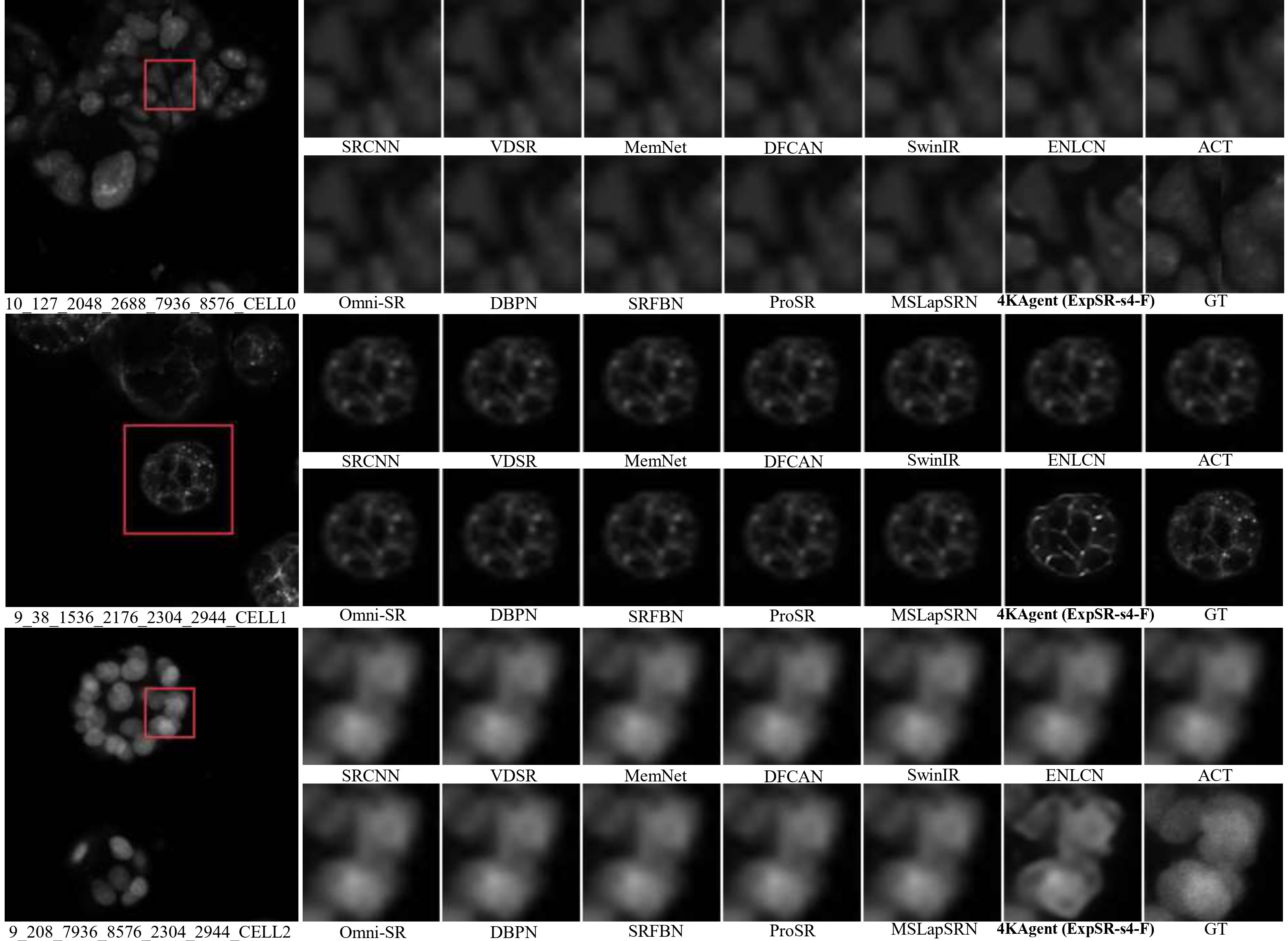}
\caption{Visualization of fluorescence microscopy image SR on SR-CACO-2 dataset (64$\rightarrow$512).
}
\label{fig:microscipy_result_8x}
\end{figure*}

\paragraph{Qualitative Comparison}
Representative qualitative results for the highly challenging 8$\times$ super-resolution task are shown in~\cref{fig:microscipy_result_8x}. At such a high magnification, where information loss is severe, the ability to reconstruct distinct biological structures for each of the three cellular markers becomes a critical test for any SISR method.

Our visual analysis reveals clear performance differences. For the inherently dim and sparse CELL0 Survivin marker, 4KAgent’s reconstruction is markedly clearer and closer to the ground truth. It successfully restores the faint midbody structure with higher fidelity than top-performing baselines like ENLCN and ACT, which struggle to resolve this signal from the background. This superior performance is also evident for CELL1, where 4KAgent delineates the membrane and cytoskeletal framework with sharp, continuous lines. In contrast, the outputs from most other methods appear noticeably blurry, failing to preserve the cell’s essential structural integrity. In the case of the bright nuclear marker CELL2, the diffuse nature of the chromatin structure means even the ground truth image itself lacks hard, well-defined edges. In this difficult context, 4KAgent reconstructs a complex, high-frequency textural pattern that is visually competitive with the other methods. While the intricate nature of the target makes absolute fidelity hard to judge, our method effectively generates a detailed result on par with other models, even under a zero-shot blind inference setting.

\paragraph{Discussions}
Our experiments show that 4KAgent delivers leading performance on the challenging SR-CACO-2 dataset, with quantitative metrics and qualitative results surpassing those of the evaluated specialist models. The superior performance of 4KAgent underscores its strong applicability to fluorescence microscopy SISR. First, it showcases strong zero-shot generalization, achieving highly competitive super-resolution performance on microscopy data, and could be further strengthened by adapting more domain-specific tools. Second, 4KAgent exhibits impressive cross-domain transferability, successfully adapting methods originally optimized for natural scenes to the distinct characteristics of fluorescence microscopy images. Third, the agent-based architecture enables the flexible and modular integration of existing models without requiring expensive retraining or model modification. Beyond immediate applications to super-resolution, the modularity and domain-agnostic nature of 4KAgent also suggest its broad potential for other real-world biomedical imaging domains where data scarcity or retraining costs are limiting factors.

\subsection{Pathology Image Super-Resolution}
\label{ssec:sci-sr-iii-patho}

Pathology images---particularly whole-slide images (WSIs) and their extracted patches---play a critical role in digital diagnostics and disease detection. Typically, glass slides containing tissue sections stained with hematoxylin and eosin are digitized using high-speed scanners at resolutions approaching $\sim$0.25 $\mu m$ per pixel, resulting in gigapixel-scale images characterized by distinct color profiles and high-frequency textures unique to cellular structures. However, the substantial costs and data storage requirements associated with ultra-high-resolution scanning have led many workflows to rely on computational upscaling from lower-resolution acquisitions. This task is challenging, as pathology images possess specialized characteristics that pose significant difficulties for conventional single-image super-resolution (SISR) methods originally optimized for natural scenes. To address this challenge, we evaluate our 4KAgent on the bcSR dataset~\cite{jia2023channelattention}, comparing it against several established techniques to assess its effectiveness in this specialized domain.

\paragraph{Settings.} 
Our evaluation is conducted on the bcSR benchmark dataset curated for pathology image super-resolution. The bcSR dataset was derived from the larger CAMELYON~\cite{Litjens2018CAMELYON} dataset, which contains whole-slide images (WSIs) of H\&E-stained breast cancer sentinel lymph node sections. To create bcSR, the authors first sampled representative 1024$\times$1024 patches from the original WSIs. Subsequently, a filtering process was applied to remove patches with large blank areas and to select for images with high color channel variance, ensuring the dataset was rich in informative and challenging tissue structures. The final bcSR dataset consists of 1,200 unique images, which are split into a 1,000-image training set and a 200-image test set.

Following the standard protocol established by the bcSR benchmark, the high-resolution ground truth images were downsampled using bicubic interpolation to generate the low-resolution inputs. We evaluated 4KAgent using the \textbf{ExpSR-s4-F} and \textbf{ExpSR-s8-F} profiles for the 4$\times$ and 8$\times$ tasks, respectively, prioritizing pixel fidelity which is critical for preserving fine diagnostic details. Performance was measured using the PSNR and SSIM metrics.

\paragraph{Quantitative Comparison}
Quantitative results for the pathology image super-resolution task are summarized in~\cref{tab:pathology-psnr-ssim}. Across both 4$\times$ and 8$\times$ upsampling tasks, our 4KAgent achieves the highest SSIM score among all evaluated methods. While CARN, a model specifically designed for this pathology dataset, attains a marginally higher Peak PSNR, 4KAgent's superior SSIM is more indicative of its ability to accurately preserve the complex tissue morphology and textures that are essential for pathological diagnosis, which is more critical for the reliability of features used for clinical assessment. This demonstrates the effectiveness of 4KAgent in recovering diagnostically relevant details from downsampled pathology patches, with performance comparable or superior to other specialized, fully-trained models.
\begin{table*}[!h]
\centering
% \scriptsize
\renewcommand{\arraystretch}{1.2}
\centering
\fontsize{6.5pt}{7.5pt}\selectfont
\setlength{\tabcolsep}{6pt}
\caption{Quantitative comparison on Pathology dataset. The top three performances of each metric are marked in \textbf{bold}, \underline{underline}, \textit{italic} respectively.}
\label{tab:pathology-psnr-ssim}
\begin{tabular}{lcc|cc}
\toprule
\textbf{Method} & \multicolumn{2}{c}{4$\times$} & \multicolumn{2}{c}{8$\times$} \\
\cmidrule(lr){2-3} \cmidrule(lr){4-5}
       & \textbf{PSNR$\uparrow$}    & \textbf{SSIM$\uparrow$}    & \textbf{PSNR$\uparrow$}    & \textbf{SSIM$\uparrow$}     \\
\midrule
Bicubic       & 27.019 & 0.6659 & 22.475 & 0.2776 \\
% \midrule
SRCNN~\cite{dong2014learning}
               & 27.475 & 0.7329 & 22.489 & 0.3624 \\
% \midrule
SRGAN~\cite{ledig2017photo}
               & 28.606 & 0.7719 & 23.729 & 0.5580 \\
% \midrule
EDSR~\cite{lim2017enhanced}    
               & 29.830 & 0.8058 & 24.366 & 0.5715 \\
% \midrule
RDN~\cite{zhang2018residual}          
               & \textit{29.913} & 0.8074 & 24.392 & 0.5711 \\
% \midrule
RCAN~\cite{zhang2018image}    
               & \underline{29.916} & \textit{0.8085} & \textit{24.404} & 0.5749 \\
% \midrule
SWD-Net~\cite{chen2020joint}    
               & 29.853 & 0.8000 & \underline{24.465} & \textit{0.5755} \\
% \midrule
CARN~\cite{jia2023channelattention}          
    & \textbf{29.964} & \underline{0.8408} & \textbf{24.479} & \underline{0.5763} \\
\midrule
\rowcolor{LightGray!70}
\textbf{4KAgent (ExpSR-sN-F)} (N $\in$ [\texttt{4,8}]) 
    & 29.746 & \textbf{0.8602} & 24.300 & \textbf{0.5826} \\
\bottomrule
\end{tabular}
\end{table*}

\begin{figure*}[!h]
\centering
\includegraphics[width=1.0\textwidth]{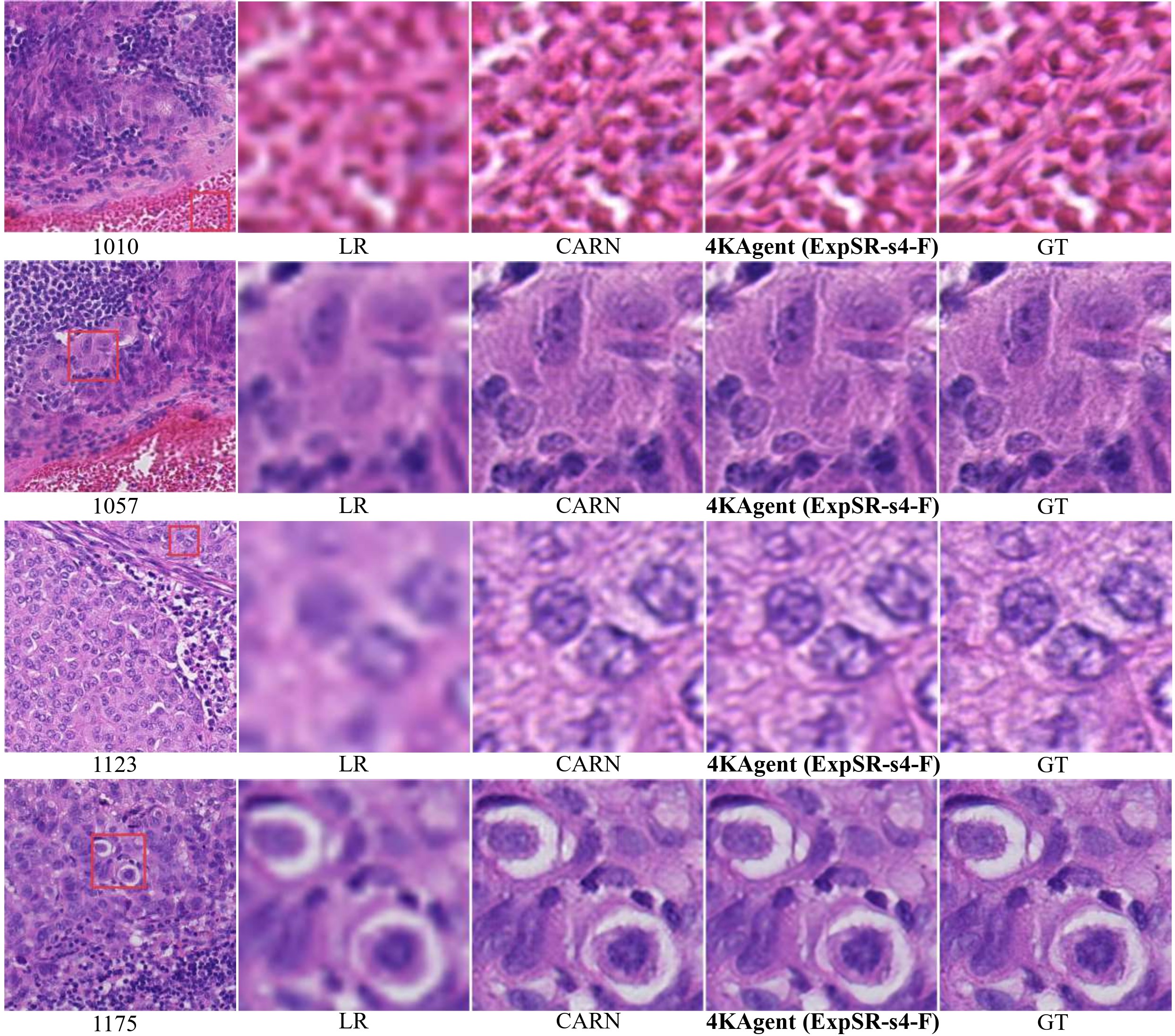}
\caption{Visual comparison of pathology image super-resolution on bcSR dataset (256$\rightarrow$1024).
}
\label{fig:pathology_result_4x}
\end{figure*}

\paragraph{Qualitative Comparison}
\cref{fig:pathology_result_4x} presents a qualitative comparison for 4$\times$ super-resolution on representative patches from the bcSR test set. While both methods significantly improve upon the heavily blurred low-resolution inputs, a closer inspection of the ROIs reveals that our training-free 4KAgent consistently produces results that are on par with, and often superior to, the fully-trained, domain-specific CARN model. This superior performance is particularly evident in challenging cases. In image 1010, 4KAgent successfully delineates individual cell boundaries and restores the heterogeneous texture of the tissue architecture. In contrast, CARN's output suffers from a loss of sharpness and definition, resulting in noisier and blurrier cell regions and poorly defined tissue architecture, while also introducing a slight color deviation. Similarly, in image 1175, 4KAgent accurately reconstructs the intricate internal structures, preserving the sharp outlines of the nuclei and cytoplasm. CARN's output, conversely, suffers from a loss of sharpness and inaccurate detail, while also exhibiting subtle grid-like artifacts. Across all examples, 4KAgent consistently generates nuclei with sharper boundaries and more distinct internal textures, along with clearer cell membranes, demonstrating a higher fidelity to the ground truth.

These visual improvements also directly correlate with 4KAgent's higher SSIM scores, confirming its enhanced ability to preserve the structural integrity of the tissue. The accurate recovery of such fine-grained morphological details is critical for potential downstream clinical applications. High-fidelity reconstructions like those from 4KAgent can enable more reliable automated analysis, such as precise nuclei segmentation for cell counting, classification of cellular atypia, and grading of cancerous tissue, thereby highlighting the potential value of our approach in digital pathology workflows.

\paragraph{Discussions}
The combined quantitative and qualitative results underscore the significant potential of 4KAgent for pathology image super-resolution. Although not leading in PSNR, 4KAgent's superior SSIM scores demonstrate a more accurate reconstruction of high-frequency textures and tissue morphology, which are paramount for pathological interpretation. Furthermore, because 4KAgent is not trained on a specific pathology dataset, it is less susceptible to overfitting to the characteristics of a single data source. This provides a significant advantage when performing SR on real-world pathology images acquired by different scanners and staining protocols. Additionally, 4KAgent's agentic framework allows for flexible expansion of its tool profile to better adapt to pathology imaging modalities. Its leading performance on the bcSR dataset validates the potential of this agentic approach as a robust and generalizable solution for biomedical imaging.

The ability to generate reconstructions with high structural fidelity has direct implications for critical downstream applications in computational pathology. By restoring sharper nuclei, clearer cell membranes, and more intelligible tissue architecture, 4KAgent provides a more reliable input for automated analysis pipelines. This can enhance the accuracy of tasks such as nuclei segmentation, cell counting, and the classification of cancerous tissue, ultimately making it a more robust and practically useful tool for AI-assisted biomedical diagnostics.

\subsection{Medical Image Super-Resolution: X-Ray, Ultrasound, and Fundoscopy}
\label{ssec:sci-sr-iiii-med}
In this chapter, we shift our focus to the super-resolution of clinical diagnostic imaging modalities, where the primary goal is to enhance anatomical and pathological details for improved diagnostic accuracy while minimizing patient burden, such as reducing exposure to ionizing radiation in X-ray imaging~\cite{umirzakova2024medical}. Although often grouped together, these modalities can be fundamentally diverse, operating on different physical principles: from X-rays utilizing ionizing radiation to ultrasound relying on acoustic waves. This diversity gives rise to unique image characteristics and modality-specific challenges, such as maintaining pathological invariance in chest X-rays or avoiding the generation of pseudo-structures in ultrasound images.

The prevailing approach in medical image SR has been the development of highly specialized models, each trained on specific datasets for a single modality~\cite{Yu2025Survey}. The rise of foundation models has led to more powerful specialist systems, such as those tailored for a single modality like CT or dermatology~\cite{pai2025vision, Kim2024MONET}, though a few pioneering works have begun to explore more universal solutions~\cite{lin2025healthgpt}. A significant drawback of such a specialized paradigm is poor generalization across different datasets and modalities, which creates a major bottleneck for practical clinical deployment and motivates our evaluation of 4KAgent's performance on these challenges. To this end, this chapter evaluates 4KAgent's performance across several distinct and clinically important modalities.

\paragraph{Settings.} 
We evaluate 4KAgent with the \textbf{ExpSR-s4-F} profile across three medical imaging modalities with distinct imaging principles: X-ray, ultrasound, and fundoscopy, benchmarking against their respective baselines. Benchmark datasets of each imaging modality are summarized as follows:

\begin{itemize}[leftmargin=*]
    \item \textbf{X-ray}. Chest X-ray 2017~\cite{kermany2018identifying} and Chest X-ray 14~\cite{wang2017chestx}. Specifically, Chest X-ray 2017 is a dataset of 5,856 pediatric images from Guangzhou Women and Children’s Medical Centre, split into 5,232 images for training and 624 images for testing. Chest X-ray 14 contains 112,120 frontal-view X-rays of 30,805 patients with 14 disease labels mined from radiology reports. Among them, 880 images additionally contain expert-annotated bounding boxes. Following the settings in \cite{xu2020low}, we evaluate on the Chest X-ray 2017 test set and the 880 annotated data in Chest X-ray 14. 
\end{itemize}

\begin{itemize}[leftmargin=*]
    \item \textbf{Ultrasound Image}. US-Case~\cite{US-CASE2025} and MMUS1K~\cite{Ultrasound}. The US-Case collection comprises over 7,000 sonographic images spanning organs such as the liver, heart, and mediastinum. Adopting the selection protocol from~\cite{Ultrasound}, we reused the subset of 111 images in the test set for benchmarking, excluding 11 scans whose small field of view limited their diagnostic value. MMUS1K features 1,023 anonymized multi-organ ultrasound scans, including bladder, gallbladder, thyroid, kidney, etc., sourced from Shanghai Tenth People’s Hospital. All images meet a minimum resolution of 448$\times$600 px and were cleansed of watermarks and blurring artifacts via LabelImg. The test set with label numbers from 0801 to 0900 was used for evaluation.
\end{itemize}

\begin{itemize}[leftmargin=*]
    \item \textbf{Fundoscopy Image}. DRIVE~\cite{staal2004ridge} consists of 40 color fundus images from a diabetic retinopathy screening program in the Netherlands collected by a Canon CR5 non-mydriatic 3CCD camera. The dataset is equally divided into 20 for training and 20 for testing. As all images are in 584$\times$565 resolution, following the setup in \cite{ahmad2022new}, the original high-resolution (HR) images were resized to 512$\times$512 and before being further utilized to generate LR pairs.
\end{itemize}

To consist with baseline methods for each dataset, the LR input images were generated by downsampling HR images via bicubic interpolation. For X-ray images, we use SSIM, FSIM~\cite{zhang2011fsim}, and MSIM~\cite{wang2003multiscale} as metrics, while PSNR and SSIM are used for Ultrasound and Fundoscopy images.

\begin{table*}[!h]
\centering
\scriptsize
\caption{Quantitative comparison on X-ray datasets. The top three performances of each metric are marked in \textbf{bold}, \underline{underline}, \textit{italic} respectively.}
\label{tab:X-ray-psnr-ssim}
\begin{tabular}{lccc|ccc}
\toprule
\textbf{Method} & \multicolumn{3}{c}{\textbf{Chest X-ray 2017}} & \multicolumn{3}{c}{\textbf{Chest X-ray 14}} \\
\cmidrule(lr){2-4} \cmidrule(lr){5-7}
       & \textbf{SSIM$\uparrow$}    & \textbf{FSIM$\uparrow$}    & \textbf{MSIM$\uparrow$ }   & \textbf{SSIM$\uparrow$}     & \textbf{FSIM$\uparrow$}    & \textbf{MSIM$\uparrow$}     \\
\midrule
Nearest-Neighbor &  0.637 & 0.672 & 0.668
        & 0.701 & 0.724 & 0.713 \\
\midrule
Interpolation~\cite{yang2014comparative}   
        & 0.615  & 0.663   & 0.644
        & 0.687   & 0.698   & 0.681   \\
\midrule
CTF~\cite{zhang2016coarse}    
        & 0.889   & 0.933   & \textit{0.954}
        & 0.917   & 0.955   & 0.943   \\
\midrule
ESPCN~\cite{shi2016real}  
        & 0.756   & 0.825   & 0.804
        & 0.795   & 0.822   & 0.815   \\
\midrule
FSRCNN~\cite{dong2016accelerating}     
        & \textit{0.897}   & 0.943   & 0.953
        & 0.917   & 0.959  & \textit{0.953}   \\
\midrule
LapSRN~\cite{lai2017deep}
        & 0.893   & 0.942   & \textit{0.954}
        & 0.915   & 0.956   & 0.949   \\
\midrule
SRGAN~\cite{ledig2017photo}
        & 0.821   & 0.896   & 0.868
        & 0.844   & 0.903   & 0.897   \\
\midrule
GAN-CIRCLE~\cite{you2019ct}
        & \textit{0.897}   & \textit{0.947}   & 0.923
        & \textit{0.919}   & \textit{0.969}   & 0.945   \\
\midrule
SNSRGAN~\cite{xu2020low}
        & \underline{0.911}   & \underline{0.981}   & \underline{0.983}
        & \underline{0.925}   & \underline{0.995}   & \underline{0.986}   \\
\midrule
\rowcolor{LightGray!70}
\textbf{4KAgent (ExpSR-s4-F)}    
        & \textbf{0.933}   & \textbf{0.996}   & \textbf{0.987}
        & \textbf{0.960}   & \textbf{0.999}   & \textbf{0.993}   \\
\bottomrule
  \end{tabular}
\end{table*}

\paragraph{Quantitative Comparison}
X-ray Quantitative results are summarized in~\cref{tab:X-ray-psnr-ssim}. Ultrasound Quantitative results are summarized in~\cref{tab:ultrasound-4x-restructured-corrected}. Fundoscopy Quantitative results are summarized in~\cref{table:DRIVE}, which collectively demonstrate 4KAgent's consistently superior performance across all three distinct medical imaging modalities. On the X-ray datasets, 4KAgent with Fidelity profile surpasses the specialized SNSRGAN~\cite{xu2020low} model across all structure-focused metrics. For Ultrasound imaging, it also achieves a significant performance leap, boosting the PSNR on the MMUS1K dataset by nearly 3 dB over the previous state-of-the-art, M2Trans~\cite{Ultrasound}. Similarly, on the DRIVE Fundoscopy dataset, 4KAgent again sets a new performance benchmark, improving the PSNR from 37.72 to 41.52 and the SSIM from 0.91 to 0.95. This consistent outperformance across modalities, from the need for pathological invariance in X-rays to the clarity of fine vessels in fundoscopy, highlights the effectiveness and robustness of 4KAgent for diverse medical SR tasks.

\begin{wraptable}[15]{r}{0.5\textwidth}
    \centering
    \fontsize{6.5pt}{9.5pt}\selectfont    \setlength{\tabcolsep}{2.5pt}
    \caption{Quantitative comparison on Ultrasound dataset. The top three performances of each metric are marked in \textbf{bold}, \underline{underline}, \textit{italic} respectively.}
    \label{tab:ultrasound-4x-restructured-corrected}
    \begin{tabular}{lcc|cc}
        \toprule
        \textbf{Method} & \multicolumn{2}{c}{\textbf{US-CASE}} & \multicolumn{2}{c}{\textbf{MMUS1K}} \\
        \cmidrule(lr){2-3} \cmidrule(lr){4-5}
        & \textbf{PSNR$\uparrow$} & \textbf{SSIM$\uparrow$} & \textbf{PSNR$\uparrow$} & \textbf{SSIM$\uparrow$} \\
        \midrule
        Bicubic & 28.90 & 0.7892 & 28.24 & 0.7817 \\
        \midrule
        EDSR~\cite{lim2017enhanced} & 30.82 & \textit{0.8497} & 30.04 & \textit{0.8326} \\
        \midrule
        SwinIR~\cite{liang2021swinir} & 28.50 & 0.7834 & 27.66 & 0.7758 \\
        \midrule
        ELAN~\cite{zhang2022efficient} & \textit{31.02} & 0.8464 & \textit{30.40} & 0.8309 \\
        \midrule
        ESRT~\cite{lu2022transformer} & 30.84 & 0.8374 & 30.25 & 0.8235 \\
        \midrule
        HAT~\cite{chen2023activating} & 28.72 & 0.7812 & 28.08 & 0.7582 \\
        \midrule
        M2Trans~\cite{Ultrasound} & \underline{31.32} & \underline{0.8516} & \underline{30.68} & \underline{0.8392} \\
        \midrule
        \rowcolor{LightGray!70}
        \textbf{4KAgent (ExpSR-s4-F)} & \textbf{33.27} & \textbf{0.8895} & \textbf{33.58} & \textbf{0.8678} \\
        \bottomrule
    \end{tabular}
\end{wraptable}

\begin{figure*}[!h]
\centering
\includegraphics[width=1.0\textwidth]{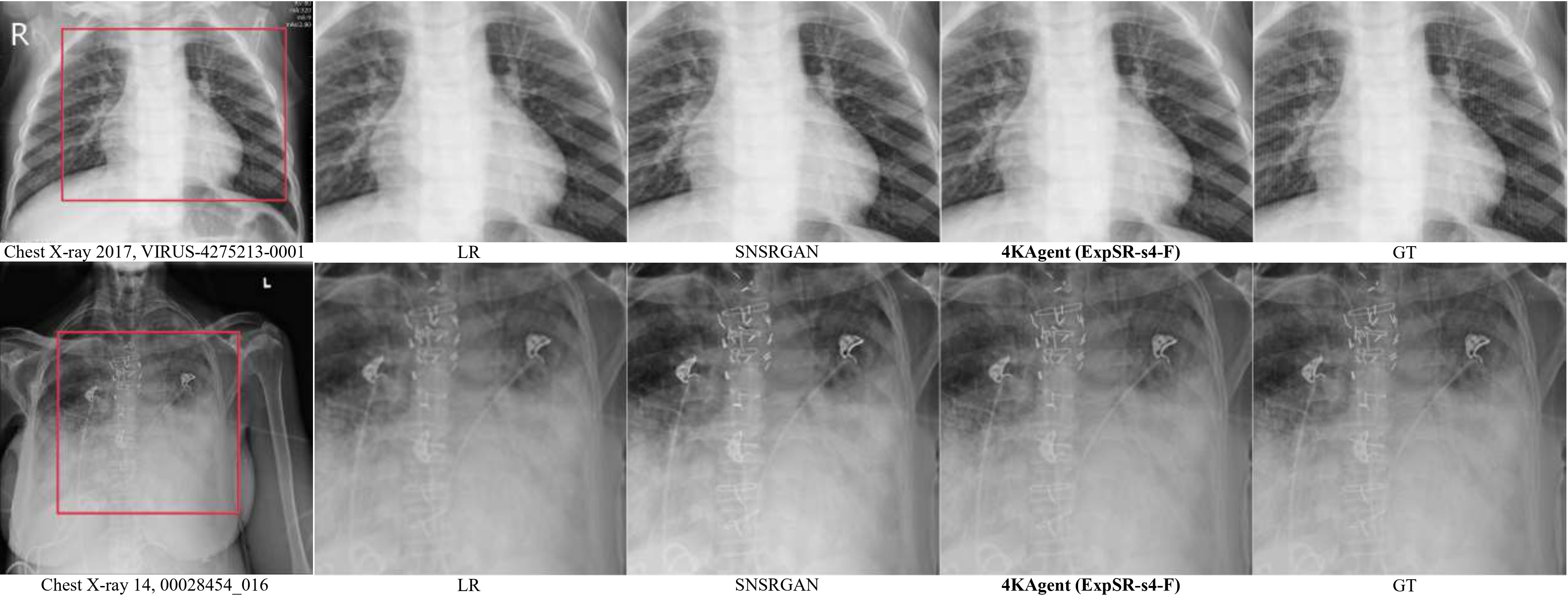}
\caption{Visual comparison of X-Ray image SR on Chest X-ray 2017 and Chest X-ray 14 dataset.
}
\label{fig:xray_result_4x}
\end{figure*}

\begin{figure*}[!h]
\centering
\includegraphics[width=1.0\textwidth]{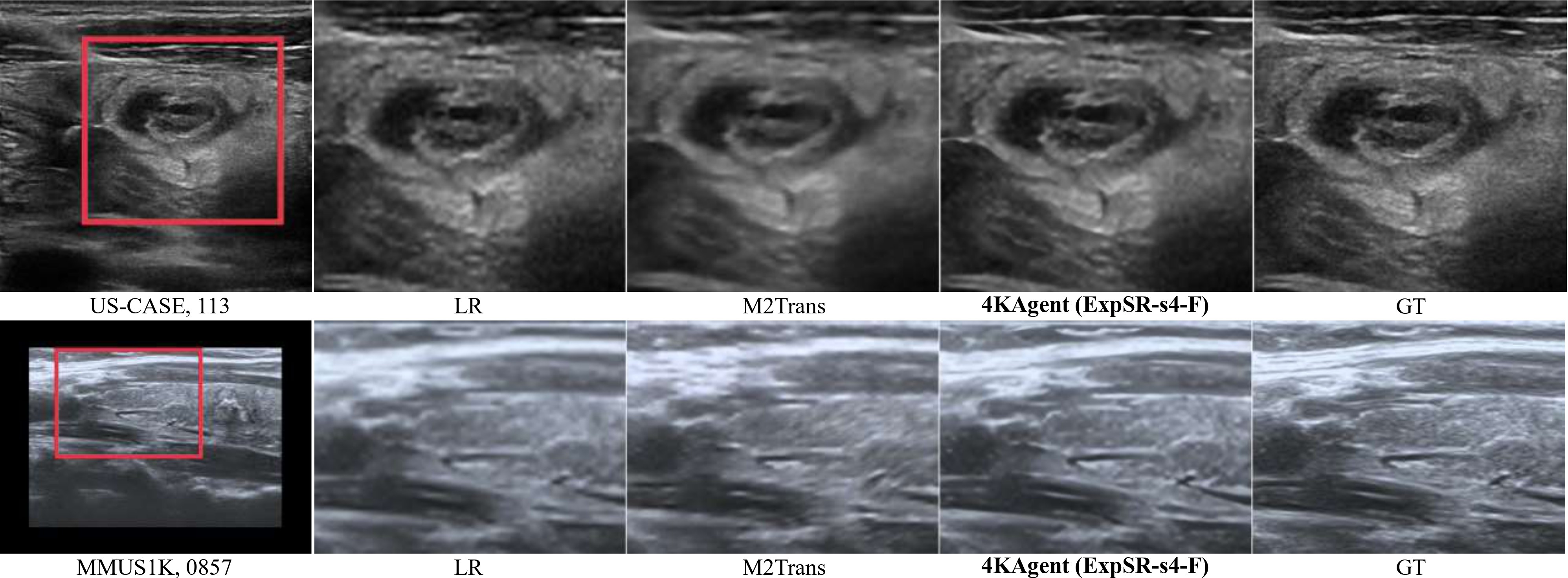}
\caption{Visual comparison of Ultrasound image SR on US-CASE and MMUS1K dataset.
}
\label{fig:ultrasound_result_4x}
\end{figure*}

\begin{figure*}[!h]
\centering
\includegraphics[width=1.0\textwidth]{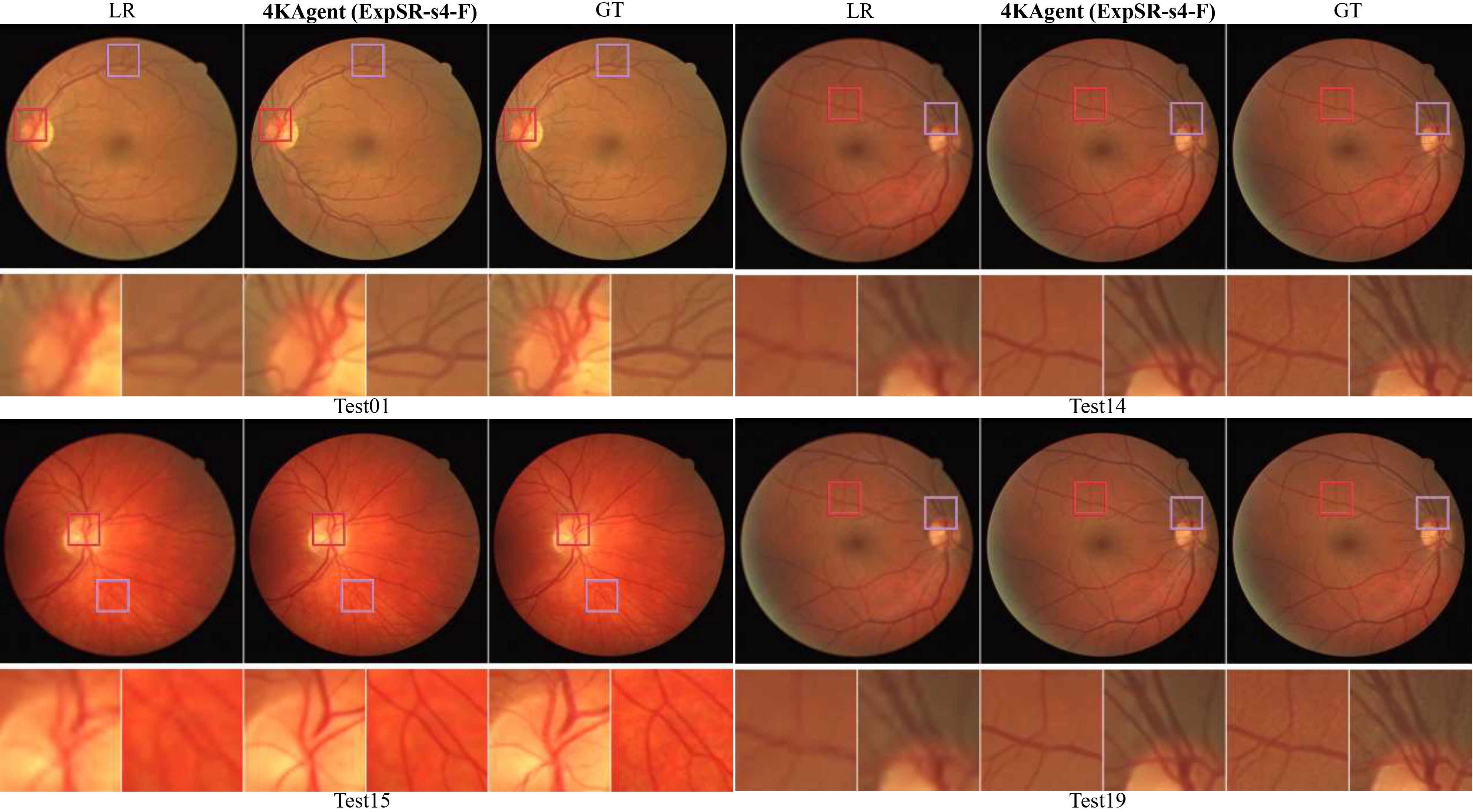}
\caption{Visual comparison of fundoscopy image SR on DRIVE dataset (128$\rightarrow$512).
}
\label{fig:fundoscopy_result_4x}
\end{figure*}

\paragraph{Qualitative Comparison}
Representative qualitative results for X-Ray SR, ultrasound SR, and fundoscopy SR are shown in~\cref{fig:xray_result_4x,fig:ultrasound_result_4x,fig:fundoscopy_result_4x}. The visual outcomes generally align with our quantitative findings and suggest the potential benefits of 4KAgent for clinical imaging applications.

\begin{wraptable}[12]{r}{0.5\textwidth}
\vspace{\intextsep}
    \centering
    \fontsize{6.5pt}{9.5pt}\selectfont    \setlength{\tabcolsep}{2.5pt}
    \caption{Quantitative comparison on Fundoscopy dataset. The top three performances of each metric are marked in \textbf{bold}, \underline{underline}, \textit{italic} respectively.}
    \label{table:DRIVE}
    \begin{tabular}{l|l|cc}
        \toprule
        {\textbf{Dataset}} & {\textbf{Method}} &   \textbf{PSNR$\uparrow$}  &  \textbf{SSIM$\uparrow$}        \\
        \midrule
        \multirow{4}{*}{DRIVE}
        & Bicubic & 25.20 &0.86 \\
        & SRGAN~\cite{ledig2017photo} & \textit{34.22}&\textit{0.88}\\
        & Ahmad \emph{et al.}~\cite{ahmad2022new} & \underline{37.72}&\underline{0.91}  \\
        & \cellcolor{LightGray!70}\textbf{4KAgent (ExpSR-s4-F)} & \cellcolor{LightGray!70}\textbf{41.52} & \cellcolor{LightGray!70}\textbf{0.95} \\
        \bottomrule
    \end{tabular}
\end{wraptable}

For X-ray super-resolution, where maintaining pathological invariance is important, 4KAgent produces reconstructions with improved delineation of lung parenchyma and clearer visibility of rib cage contours. It appears to achieve this clarity while reducing some of the over-smoothing artifacts occasionally seen in other SR methods, thus helping to preserve diagnostic details that are crucial for identifying pulmonary abnormalities with greater confidence.

In the ultrasound comparisons, 4KAgent shows a notable advantage. On the US-CASE example, it restores clearer tissue boundaries and more internal detail compared to the blurrier reconstruction from the M2Trans baseline. Similarly, for the MMUS1K image, 4KAgent appears to reduce speckle noise while enhancing anatomical definition, whereas the baseline result is affected by some noise and artifacts. In both cases, 4KAgent generates echogenic patterns that more closely resemble the ground truth, improving overall image fidelity.

The fundoscopy results demonstrate 4KAgent's effectiveness in restoring details from degraded inputs. Compared to the LR image, 4KAgent's reconstruction of the retinal vascular network shows a clear improvement. The method produces sharper and more continuous vessels, resolving many of the fine micro-vessels and bifurcation points that are obscured in the LR version. The resulting image more closely resembles the HR ground truth, suggesting its potential to aid in retinopathy screening from lower-resolution captures without sacrificing critical diagnostic details.


\paragraph{Discussions}
From both quantitative and qualitative perspectives, our evaluation suggests that 4KAgent is a capable system for cross-domain super-resolution across diverse clinical imaging modalities, showing competitive performance on X-ray, ultrasound, and fundoscopy datasets. This result is notable, as the prevailing approach often involves developing specialized models for each modality, a paradigm that can be limited by poor generalization across different scanners and protocols. By not relying on domain-specific training, 4KAgent’s agentic framework offers a flexible alternative, adaptively deploying its tools to address the unique challenges of each image, from enhancing the clarity of lung markings in chest radiographs to defining subtle echogenic interfaces in ultrasound and resolving fine vascular networks in fundoscopy.

The ability to generate reconstructions with improved structural detail may have implications for downstream applications. For example, improved sharpness in retinal vessels could aid in retinopathy screening; clearer ultrasound images help with tissue boundary delineation for segmentation; and more detailed X-rays could enhance the visibility of subtle pulmonary abnormalities. Ultimately, the performance of our agentic approach indicates its significant potential for robust deployment across more real-world clinical workflows and imaging modalities, driven by its adaptability and inherent extensibility for incorporating more specialized medical profiles.

\section{Ablation Studies}
\label{sec:ablation}

In this section, we conduct ablation studies on core components in 4KAgent system: (1) Q-MoE policy and (2) Face restoration pipeline. Then, we perform running time analysis of 4KAgent.

\textbf{Q-MoE policy.} To assess the contribution of our Q-MoE mechanism during execution and reflection, we perform an ablation study in which Q-MoE is replaced by the DFS strategy from AgenticIR \cite{zhu2024intelligent}, denoting this variant as \textbf{4KAgent (DFS)}. Experiments are conducted on the MiO-100 Group C dataset under the multiple-degradation image restoration setting.

\begin{figure*}[!h]
\centering
\includegraphics[width=\textwidth]{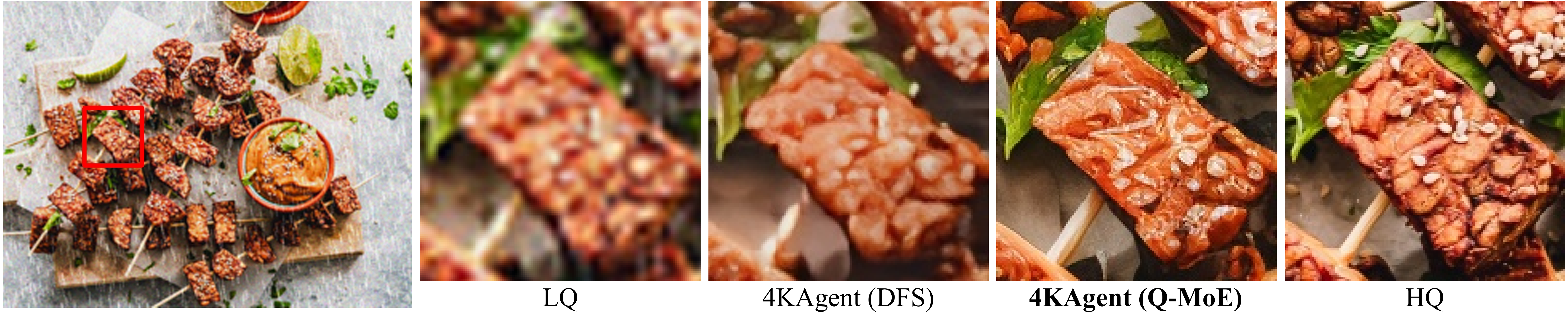}
\caption{Visual comparisons for ablation study on Q-MoE.}
\vspace{-3pt}
\label{fig:ablation_compare_moe}
\end{figure*}

As shown in \cref{table:ablation_moe}, integrating Q-MoE leads to substantial improvements in perceptual quality. Specifically, metrics such as LPIPS, MANIQA, CLIPIQA, and MUSIQ exhibit significant gains, with minimal impact on fidelity metrics like PSNR and SSIM. Furthermore, the visual comparisons presented in \cref{fig:ablation_compare_moe} provide additional evidence, showing that 4KAgent equipped with Q-MoE generates noticeably sharper and more realistic details compared to the DFS-based variant.
\begin{table*}[!h]
\renewcommand{\arraystretch}{1.2}
\centering
\fontsize{6.5pt}{7.5pt}\selectfont
\setlength{\tabcolsep}{2.5pt}
\caption{Ablation study on Q-MoE policy. The better performance is marked in \textbf{bold}.}
\label{table:ablation_moe}
\begin{tabular}{l|l|cccccc}
\toprule
{\textbf{Dataset}} & {\textbf{Method}}   & \textbf{PSNR$\uparrow$}&  \textbf{SSIM$\uparrow$} &  \textbf{LPIPS$\downarrow$}   &  \textbf{MANIQA$\uparrow$}  &  \textbf{CLIPIQA$\uparrow$}  &  \textbf{MUSIQ$\uparrow$}   \\ 
\midrule
\multirow{2}{*}{MiO100 - Group C}
& 4KAgent (DFS) & \textbf{19.81} & \textbf{0.5785} & 0.4381 & 0.3286 & 0.4854 & 54.03  \\   
& 4KAgent (\textbf{Q-MoE}) & 19.77  & 0.5629  & \textbf{0.4271}  & \textbf{0.3545} & \textbf{0.5233}  & \textbf{55.56}   \\ 
\bottomrule
\end{tabular}
\end{table*}
\vspace{-2pt}

\textbf{Face restoration pipeline.} To evaluate the impact of our Face Restoration Pipeline, we conduct an ablation study on the WebPhoto-test dataset using three profiles: \textbf{ExpSR-s4-P}, \textbf{ExpSRFR-s4-P}, and \textbf{GenSRFR-s4-P}. Experiment result and the difference among these profiles are shown in~\cref{table:ablation_face}. 

Enabling the face restoration module (i.e., switching profile from \textbf{ExpSR-s4-P} to \textbf{ExpSRFR-s4-P} and \textbf{GenSRFR-s4-P}) yields higher face IQA scores (CLIB-FIQA and DSL-FIQA). Moreover, when setting the `Restore Option' to `None' rather than `super-resolution', we observe further improvements across both generic image IQA metrics (NIQE and MUSIQ) and face IQA metrics.
\begin{table*}[!h]
\renewcommand{\arraystretch}{1.2}
\centering
\fontsize{6.5pt}{7.5pt}\selectfont
\setlength{\tabcolsep}{2.5pt}
\caption{Ablation study on face restoration pipeline on WebPhoto-Test dataset. The best and second-best performances are marked in \textbf{bold} and \underline{underline} respectively.}
\label{table:ablation_face}
\begin{tabular}{l|cc|cccccc}
\toprule
{\textbf{Method}} & \textbf{Restore Option} &  \textbf{Face Restore}  &   \textbf{NIQE$\downarrow$}  &  \textbf{CLIPIQA$\uparrow$} & \textbf{MUSIQ$\uparrow$}  &  \textbf{MANIQA$\uparrow$}  &  \textbf{CLIB-FIQA$\uparrow$}  & \textbf{DSL-FIQA$\uparrow$}     \\
\midrule
4KAgent (ExpSR-s4-P) & super-resolution & False  & 5.11  & \textbf{0.7119} & \underline{73.62} & \textbf{0.6601} & 0.6415 & 0.7194  \\
4KAgent (ExpSRFR-s4-P) & super-resolution & True & \underline{4.53}  & 0.6600  & 72.89 & 0.6405 & \underline{0.6602} & \underline{0.7237} \\
4KAgent (GenSRFR-s4-P) & None & True & \textbf{4.15}  & \underline{0.7077} & \textbf{75.92} & \underline{0.6576} & \textbf{0.6671} & \textbf{0.7683} \\
\bottomrule
\end{tabular}
\end{table*}

\begin{figure*}[!h]
\centering
\includegraphics[width=\textwidth]{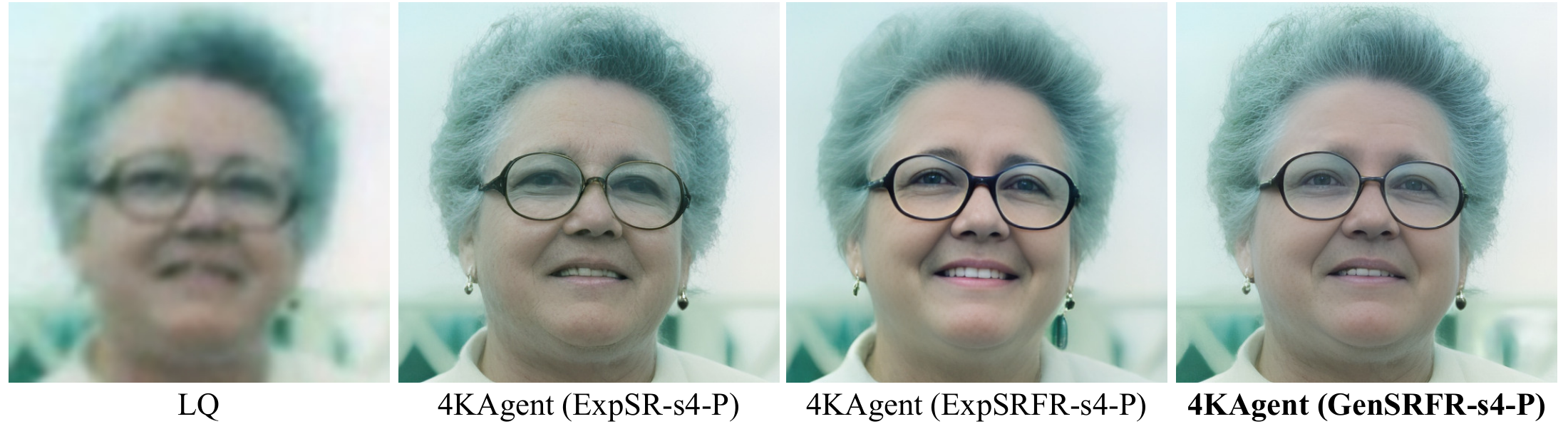}
\caption{Visual comparisons for ablation study on face restoration pipeline.}
\vspace{-5pt}
\label{fig:ablation_compare_face}
\end{figure*}

Visual comparisons are shown in~\cref{fig:ablation_compare_face}. 4KAgent with \textbf{GenSRFR-s4-P} profile produces the finest facial details (hair details, harmony of facial area and background area). This trend indicates that WebPhoto-test images suffer from complex, mixed degradations and 4KAgent benefits from integrating multiple restoration tasks with Q-MoE driven selection to achieve superior visual quality.

\textbf{Running Time Analysis.} The inference time of 4KAgent varies depending on the selected profile, the quality of the input image, and the length of the restoration plan. In this section, we analyze the inference time of 4KAgent using NVIDIA RTX 4090 GPUs. Specifically, we report the fastest and slowest cases observed in our experiments. The fastest case involves super-resolving images ($\times4$) from the B100 dataset using the \textbf{ExpSR-s4-F} profile. The slowest case corresponds to jointly restoring and upscaling low-quality images from the DIV4K-50 dataset to 4K resolution under the \textbf{Gen4K-P} profile. The inference times for these two cases are summarized in~\cref{table:4kagent_infer_time}.
\vspace{-3pt}
\begin{table}[!h]
    \centering
    \caption{Inference time of 4KAgent (fastest and slowest cases on our experiments).}
    \vspace{6pt}
    \fontsize{6.5pt}{8.5pt}\selectfont
    \setlength{\tabcolsep}{4.0pt}
    \label{table:4kagent_infer_time}
    \begin{tabular}{l l l l c c}
        \toprule
        \textbf{Profile Nickname} & \textbf{Task} & \textbf{Resolution} & \textbf{Benchmark} & \textbf{Length of Plan} & \textbf{Inference Time (s)}  \\
        \midrule
        ExpSR-s4-F & Super-resolution ($4\times$) & $120 \times 80 \rightarrow 480 \times 320$ & B100 & $1.0 \pm 0.0$ & $50.96 \pm 2.01$ \\
        \midrule
        Gen4K-P & Joint restoration + 4K Upscaling & $256 \times 256 \rightarrow 4096 \times 4096$ & DIV4K-50 & $3.4 \pm 0.6$ & $1551.76 \pm 230.73$ \\
        \bottomrule
    \end{tabular}
\end{table}
\vspace{-6pt}

As 4KAgent currently executes its tools sequentially, there is substantial potential for acceleration, for example, by running independent restoration tools in parallel at each step.

\section{Applications and Broader Impacts}

\subsection{Applications}

\paragraph{High-resolution On-Demand Media Streaming}
\namenocolor\ offers significant potential for enabling network operators, such as YouTube, Netflix, Instagram, Amazon Prime Video, TikTok, Kwai, Snap, Twitch, to name a few, to deliver 4K-quality video services from much lower-bitrate streams.
For example, edge-based SR can upscale a 1K stream to 4K at the user's device~\cite{zhan2021achieving}, allowing providers to store and transmit mostly lower-resolution content (\eg, 1K) which can then be upscaled to 4K quality on the end-user's device using edge-based processing.
This approach dramatically cuts storage and bandwidth costs (and even energy use) compared to naïvely streaming native 4K~\cite{li2023towards}.
 Technologies like NVIDIA's Deep Learning Super Sampling (DLSS)~\cite{dlss} demonstrate the feasibility and usability of real-time super-resolution on GPU chips.
Integrating such real-time upscaling into adaptive streaming protocols could also improve user experience by minimizing disruptive quality shifts often associated with variable network conditions, ensuring viewers consistently receive high-resolution playback on capable displays.

\paragraph{Video Conferencing and Telepresence}
Network bandwidth constraints and limitations inherent in typical webcams or smartphone cameras often necessitate transmitting video streams at resolutions lower than 4K. 
Implementing SR algorithms, such as 4KAgent, on the receiver's end can effectively upscale these lower-resolution feeds. This process restores fine-grained details in facial features or gestures that might otherwise be lost, thereby enhancing the perceived visual quality and potentially aiding communication cues like lip-reading or the interpretation of subtle expressions~\cite{naderi2025icme, xiao2023online, li2025ntire}. Consequently, even devices with modest camera capabilities can deliver an experience approximating 4K quality to the viewer, without requiring increased upload bandwidth from the sender. This democratization of high-resolution video conferencing can improve remote collaboration, making it more accessible and effective for users constrained by network limitations or hardware capabilities.

\paragraph{Surveillance and Security}

Image SR technologies like 4KAgent (with fidelity-based profile) offer significant value in enhancing footage from law enforcement operations, particularly from body-worn cameras and dashcams. 
These devices often capture video at resolutions like 720p or 1080p with wide fields of view, resulting in low-detail imagery, especially in challenging conditions such as low light~\cite{li2024light}. 
Faces or license plates captured at a distance may span only a few dozen pixels, far below recommended thresholds for identification (\eg,  $\sim$90$\times$60 pixels per face for courtroom evidence~\cite{vqips-report}).
 The quality is often further compromised by heavy compression and sensor limitations, introducing noise and motion blur. Modern SR approaches, particularly ``blind'' methods that model complex real-world degradations, can effectively mitigate these issues and restore detail in practical bodycam footage. 
By enhancing critical regions (faces, license plates) in police videos, SR can improve both human and automated identification, while preserving the veracity required for judicial use.

Similarly, public surveillance systems, including city-wide CCTV networks, border security cameras, and transit hub monitoring, face comparable challenges related to resolution and image quality. 
Fixed cameras covering wide areas often render persons or objects of interest with very low pixel counts, with quality impacted by distance, illumination, camera motion, and aggressive compression techniques employed to manage bandwidth and storage~\cite{nascimento2023super}. 
SR provides a means to enhance detail retroactively without costly hardware upgrades. 
Field studies have also reported the effectiveness of SR. For example, a National Institute of Justice study~\cite{abiantun2019ssr2} showed that multi-resolution SR could reconstruct identifiable features from extremely low-resolution facial images comparable to those from real-world security cameras. 
Overall, SR can act as a force multiplier for legacy surveillance infrastructure, enhancing situational awareness and forensic capabilities. However, the enhanced capability for identification also raises potential privacy concerns, which will be discussed in~\cref{ssec:limitations}.

\paragraph{Gaming and Entertainment}

SR techniques are extensively utilized in the entertainment sector to enhance visual quality while sustaining high frame rates, particularly in demanding gaming, VR, and AR applications.
A prominent example is NVIDIA's DLSS, a suite of AI-powered neural rendering techniques that upscale lower-resolution frames to higher target resolutions, as high as 4K.
DLSS can significantly improve performance, often more than doubling GPU throughput and leading to substantial frame rate increases—for instance, one report indicated a boost of up to approximately 360\% (\eg, from 8 to 36.8 fps on an RTX 2060 at 4K). Successive iterations like DLSS 3 with Frame Generation and DLSS 3.5 with Ray Reconstruction have introduced further advancements by using AI to generate additional frames or improve ray-traced effects.

\paragraph{VR, AR, and XR}
This need for efficient, high-quality rendering extends critically to the domain of spatial intelligence and computing, as seen in advanced devices like the Apple Vision Pro, which aims to deliver experiences with more pixels than a 4K TV per eye. While such platforms boast high native display resolutions, SR techniques could play a crucial role in rendering complex mixed-reality scenes or high-fidelity passthrough video efficiently, maintaining visual clarity without overwhelming the processing capabilities. Similarly, as smart glasses like the Ray-Ban Meta Wayfarer evolve and potentially incorporate more advanced display capabilities for augmented reality overlays, SR will be key to delivering crisp digital information without excessive battery drain. Broader XR initiatives, such as Google's development of Android XR, also stand to benefit from robust SR solutions to enable a diverse ecosystem of devices to achieve compelling visual experiences. For all these platforms, from gaming consoles to sophisticated XR headsets and smart glasses, the ability of SR systems like 4KAgent to adaptively enhance visual quality from various inputs will be paramount in balancing immersive, high-resolution experiences with practical performance and power constraints.

\paragraph{AI-Generated Content (AIGC) Production Industry}

Photographers, digital artists, and filmmakers increasingly leverage SR tools to enlarge, restore, and enhance the quality of both conventional (\eg, old photos, archival digital footage) and even AI-generated images and video footage. 
We have demonstrated in \cref{ssec:aigc-sr} that 4KAgent is capable of synthesizing high-fideltiy details in generated media, coinciding with a recent trend that generates a high-resolution advertisement for KFC\footnote{\url{https://x.com/Wesley_Kibande/status/1908091178723029193}}, leveraging outputs from generative video models such as Google's Veo~\cite{veo2}, Luma AI's Dream Machine~\cite{Dream-Machine}, and OpenAI's Sora~\cite{SORA}, further enhanced using  Topaz Labs Video Upscaler to achieve higher resolutions (\eg, 4K) and professional quality suitable for broader use. 
SR techniques are crucial for bridging this gap towards generating ultra-high-resolution content, enabling creators to enhance these AI-generated visuals. For instance, images generated for concept art, marketing materials, or virtual environments can be significantly improved in detail and clarity through SR, making them suitable for 4K displays or large-format printing. Similarly, generative video content, which might be created at lower resolutions to manage computational costs, can be upscaled using specialized tools like Topaz Video AI to achieve crisper, higher-resolution results (\eg, 4K) ready for distribution or integration into larger productions. State-of-the-art SR methods, including GAN-based approaches, can synthesize photorealistic details, effectively transforming AIGC outputs into polished, professional-grade assets. The ability of robust and adaptive SR solutions like 4KAgent to handle the diverse and sometimes unpredictable nature of AIGC makes them particularly valuable for ensuring that AI-driven creative endeavors can meet high-quality benchmarks.

\paragraph{Scientific Imaging}
 High-resolution imagery is crucial across numerous scientific disciplines, particularly when native sensor capabilities are constrained. In remote sensing, deep super-resolution (SR) methods significantly enhance spatial details of satellite imagery, facilitating accurate land-use classification and environmental monitoring~\cite{nguyen2021selfgs, shermeyer2019effects}. For instance, self-supervised SR techniques trained on sequences of satellite images yield sharper and less noisy results compared to raw captures, substantially improving downstream geospatial analysis~\cite{nguyen2021selfgs}. Microscopy and biomedical imaging similarly benefit from SR, particularly through novel quantum imaging techniques. Recent advancements by Zhang et al. and He et al. leverage quantum entanglement to achieve unprecedented imaging resolution, demonstrating quantum microscopy at the Heisenberg limit and significantly enhancing cellular and sub-cellular visualization~\cite{zhang2024quantumgs, he2023quantumgs}. Additionally, computational SR methods applied to microscopy, like content-aware restoration for fluorescence images~\cite{weigert2018contentgs}, complement these quantum techniques by computationally reconstructing detailed 3D biological structures from limited optical inputs. Thus, versatile and advanced SR frameworks such as 4KAgent, coupled with emerging quantum imaging methods, can revolutionize scientific research by providing richer, more precise imagery across multiple imaging modalities.

\paragraph{Medical Image Applications} In medical imaging, SR facilitates detailed diagnostics by transforming low-dose or rapidly acquired imaging scans into high-fidelity medical images. Techniques employing deep learning-based SR on modalities such as Magnetic Resonance Imaging (MRI) and Computed Tomography (CT) have shown promise in generating accurate, high-resolution images from suboptimal inputs, thus reducing patient radiation exposure and acquisition time without sacrificing diagnostic quality~\cite{chen2018braings, zhao2019channelgs}. For instance, generative adversarial networks (GANs) and transformer-based SR approaches like TransMRI demonstrate substantial improvements in enhancing anatomical details critical for diagnostic accuracy~\cite{feng2021taskgs}. Consequently, methods like 4KAgent, which provide universal super-resolution capabilities, can significantly impact clinical diagnostics by offering highly detailed and diagnostically reliable imagery from resource-efficient imaging procedures.

\paragraph{Embodied AI and Robotics} Embodied AI systems, including robotics platforms, leverage SR to enhance visual perception, critical for tasks such as navigation, object manipulation, and human-robot interaction. Robotic visual systems frequently face limitations in sensor resolution and onboard processing capacity, challenges that SR methods can effectively address. Recent studies indicate that integrating SR into robotic vision pipelines notably improves object detection and localization, particularly for distant or small-scale objects critical in dynamic environments~\cite{martinez2021super}. Furthermore, real-time lightweight SR models tailored for robotic platforms have been developed, improving perception accuracy and enabling robots to perform complex tasks efficiently, such as precise grasping and navigation through cluttered or visually challenging scenarios~\cite{angarano2023generative,islam2020simultaneous}. Consequently, robust SR algorithms substantially advance robotic autonomy and operational effectiveness.

\paragraph{Autonomous Vehicles, Drones, and Intelligent Transportation Systems (ITS)} Autonomous vehicles, aerial drones, and intelligent transportation systems increasingly depend on high-quality visual inputs for safe and efficient operation. SR technologies significantly boost the capability of onboard cameras to discern critical details from lower-resolution captures under real-world constraints. For instance, SR-enhanced imagery improves the detection accuracy of pedestrian, vehicle, and traffic sign recognition systems, crucial for autonomous navigation and safety-critical decision-making~\cite{telcceken2024new, musunuri2022srodnetgs,liu2024auto}. Research demonstrates that training object detection models with super-resolved images significantly enhances their effectiveness in challenging scenarios, including poor visibility or long-range object detection from drones or vehicle-mounted cameras~\cite{musunuri2022srodnetgs}. Furthermore, ITS can employ SR-enhanced camera networks for robust and precise traffic monitoring and anomaly detection, improving urban mobility management and public safety.

\paragraph{GIScience and GeoAI} Geographic Information Science (GIScience) heavily utilizes spatially explicit, high-resolution data for mapping, analysis, and policy-making. Deep SR has recently emerged as a powerful method for enhancing geospatial imagery resolution, substantially improving the interpretability and accuracy of satellite-derived GIS datasets. For example, SR methods significantly improve the extraction accuracy of buildings, roads, vegetation, and other geographic features from medium- to low-resolution imagery, providing a cost-effective alternative to deploying expensive high-resolution satellite sensors~\cite{shermeyer2019effects, pouliot2018landsatgs}. Additionally, high-quality SR outputs are vital for applications such as precision agriculture and disaster response planning, demonstrating the broad utility of universal SR frameworks like 4KAgent within GIScience research and applications.

\subsection{Broader Impacts}

\paragraph{Economic Impacts.}
SR technologies, exemplified by 4KAgent, drive significant economic advantages by enhancing operational efficiency, creating new markets, and supporting environmental sustainability. By reducing bandwidth, storage, and infrastructure costs associated with high-resolution content delivery, SR solutions enable economical, high-quality media dissemination over constrained networks, benefiting digital platforms and smaller businesses~\cite{li2021comisr,zhang2021video}. Companies such as BytePlus and Maxar Intelligence have successfully leveraged SR technologies to open new markets in healthcare diagnostics, geospatial intelligence, and media restoration~\cite{byteplus2025,maxar2025}. Additionally, by minimizing data storage and transmission demands, SR contributes meaningfully to the environmental goals of reduced energy consumption and lower carbon emissions~\cite{lee2019mobisr}.

\paragraph{Accessibility.}
SR substantially promotes digital equity by enabling high-quality visual content access for users in regions with limited bandwidth or resource constraints without necessitating advanced infrastructure or expensive devices~\cite{peroni2024end}. Particularly in education and healthcare, SR supports remote learning and telemedicine by delivering clearer instructional and diagnostic imagery, significantly benefiting underserved communities~\cite{byteplus2025-healthcare}. Moreover, SR-integrated assistive technologies offer improved accessibility for individuals with visual impairments by enhancing image clarity and text readability, facilitating greater inclusion and interaction within the digital sphere~\cite{lighthouseguild2025,w3c2025}.

\paragraph{Vertical Impacts to Industry.}
Demand for real-time, high-quality SR capabilities stimulates technological progress across various industries, including media, robotics, autonomous systems, and scientific visualization. SR advancement drives innovation in specialized neural hardware, edge computing solutions, and embedded AI, spurring the development of powerful and efficient Neural Processing Units (NPUs) in consumer devices~\cite{chang2018energy,lee2020srnpu}. Additionally, SR techniques significantly enhance autonomous vehicle, drone, and robotic platform capabilities by improving object detection accuracy, scene understanding, and decision-making reliability, particularly in challenging operational environments~\cite{musunuri2022srodnetgs,shermeyer2019effects}. These impacts extend to scientific instrumentation, such as microscopy and geospatial imaging, where SR enables unprecedented detail and precision~\cite{nguyen2021selfgs,weigert2018contentgs}.

\subsection{Limitations and Potential Negative Societal Impacts}
\label{ssec:limitations}
\paragraph{Efficiency and Computational Cost.}
SR methods, particularly deep learning-based and agentic frameworks like 4KAgent, often require substantial computational resources for training and inference. High-resolution SR models usually rely on resource-intensive GPU or TPU clusters, imposing significant energy consumption~\cite{song2021addersr}. Even optimized inference can become computationally burdensome at the edge, potentially limiting deployment on low-powered or mobile devices unless significant model compression and acceleration techniques are applied~\cite{lee2020srnpu,zhang2021edge}. Balancing performance and efficiency remains a critical open challenge, particularly for real-time applications in resource-constrained environments.

\paragraph{Bias, Fairness, and Model Drift.}
Data-driven SR models inherit biases from their training datasets, potentially leading to uneven quality across different image categories, demographics, or scenarios~\cite{gebru2021datasheetsgs,mehrabi2021surveygs}. Such biases might systematically disadvantage specific groups, for instance, by inadequately resolving images related to underrepresented populations or environments. Model drift over time—where the model gradually becomes less accurate due to changing real-world distributions—also poses a serious issue for practical deployment, requiring continuous monitoring and recalibration to ensure fairness and reliability~\cite{gama2014surveygs,lu2018learning}.

\paragraph{Ethical Issues and Privacy.}
Enhanced imaging capabilities enabled by SR, particularly in surveillance contexts, can amplify privacy risks. The ability to recover detailed features such as faces or license plates from previously anonymized or low-resolution imagery might lead to unauthorized or unethical identification of individuals~\cite{hartzog2018privacy}. This capability necessitates clear regulatory guidelines and ethical oversight to avoid misuse~\cite{li2025rethinking}. Concerns are especially pronounced in contexts of law enforcement, border surveillance, and public monitoring, where SR technologies must be carefully governed to prevent potential violations of civil liberties and personal privacy~\cite{crawford2021atlas}.

\paragraph{Failure Modes in High-stake Settings.}
The adoption of SR techniques in high-stakes environments, including medical diagnostics, autonomous vehicles, and security monitoring, introduces risks associated with model hallucinations or misleading image reconstructions~\cite{cohen2018distributiongs}. SR models, particularly generative approaches, may produce plausible yet incorrect details absent from the original low-resolution inputs, potentially leading to erroneous interpretations or decisions~\cite{antun2020instabilitiesgs}. In clinical settings, for example, SR-generated artifacts could result in incorrect diagnoses or overlooked medical conditions, underscoring the importance of rigorous validation and transparency regarding model uncertainty and reliability~\cite{kelly2019keygs}.

\section{Related Works}
\label{related_works}

\subsection{Image Super-Resolution}
Deep learning has significantly advanced the field of single-image Super-Resolution (SR). The seminal work, SRCNN~\cite{dong2014learning}, introduced a convolutional net for SR, with a primary focus on optimizing the Mean Squared Error between the super-resolved and high-resolution images. Following this, numerous studies have enhanced reconstruction accuracy by improving network architectures, including residual and dense connections~\cite{kim2016accurate, lim2017enhanced, zhang2018residual}, attention mechanisms~\cite{chen2020learning, dai2019second, zhang2018image}, and multi-scale networks~\cite{gao2019multi, li2018multi}. While these methods perform well in modeling the posterior distribution of the training data, they inevitably suffer from the issue of overly smooth visual results~\cite{ledig2017photo, menon2020pulse, xiao2021balanced}. In recent years, significant efforts have been made to develop generative model-based SISR techniques that produce more visually appealing results. These include autoregressive models~\cite{dahl2017pixel, menick2018generating, van2016conditional}, GAN-based models~\cite{ledig2017photo, wang2021real, zhang2021designing, chen2022real, liang2022efficient, li2024sed}, and diffusion-based models~\cite{wang2024exploiting, yang2024pixel, lin2024diffbir, wu2024seesr, saharia2022image, wang2024sinsr, wu2024one}. SRGAN~\cite{ledig2017photo}, as a pioneering GAN-based SR model, assumes image degradation through bicubic downsampling and generates photo-realistic images. BSRGAN~\cite{zhang2021designing} and Real-ESRGAN~\cite{wang2021real} achieve promising real-world SR results by using randomly shuffled degradation and higher-order degradation. SwinIR~\cite{liang2021swinir} replaces the CNN-based generator network with visual transformers, leading to more stable training and more realistic textures. Additionally, SeD~\cite{li2024sed} introduces a semantic-aware discriminator to capture fine-grained distributions by incorporating image semantics as a condition. Recent diffusion-based models have focused on fine-tuning the Stable Diffusion model for reconstructing high-quality images, using low-quality images as control signals. Notably, StableSR~\cite{wang2024exploiting} fine-tunes a time-aware encoder and employs feature warping to balance fidelity and perceptual quality. SeeSR~\cite{wu2024seesr} introduces degradation-robust, tag-style text prompts to enhance the semantic awareness of the Real-ISR model. Furthermore, recent studies on diffusion-based models, such as SinSR~\cite{wang2024sinsr}, OSEDiff~\cite{wu2024one}, PiSA-SR~\cite{sun2024pixel}, and GuideSR~\cite{arora2025guidesr} achieve one-step image super-resolution.

\subsection{Image Restoration}
Recent advances in deep learning have led to remarkable progress in blind image restoration tasks, including denoising, deblurring, deraining, dehazing, and removal of JPEG compression artifacts. Early works such as ARCNN~\cite{dong2015compression} demonstrated the potential of compact convolutional neural networks, particularly in the context of image denoising. Since then, a broad range of sophisticated network architectures and training strategies have been developed to further enhance restoration performance. These include the use of residual blocks~\cite{zhang2021plug,liu2019dual,zhang2017beyond}, attention mechanisms~\cite{yu2019free,chen2022simple,tu2022maxim,gu2019self,zamir2020learning}, and Transformer-based designs~\cite{valanarasu2022transweather,zhu2024mwformer,zamir2022restormer,wang2022uformer}; as well as generative paradigms such as GANs~\cite{gulrajani2017improved,chen2019low,bau2020semantic,gu2020image,hussein2020image,menon2020pulse,pan2021exploiting,liu2025frequency} and diffusion models~\cite{lin2024diffbir,xia2023diffir,jiang2024autodir,wang2022zero,kawar2022denoising,fei2023generative}.
Notably, general-purpose restoration models like Uformer~\cite{wang2022uformer}, MAXIM~\cite{tu2022maxim}, Restormer~\cite{zamir2022restormer}, and NAFNet~\cite{chen2022simple} have demonstrated strong performance across diverse restoration tasks, often trained independently for each specific degradation type. However, such single-degradation methods often struggle in real-world scenarios where multiple types of degradations co-exist. This limitation has sparked growing interest in the emerging field of All-in-One image restoration, which aims to build unified models capable of handling a wide range of degradations with a single network~\cite{li2022all,ma2023prores,park2023all,yao2024neural,zhang2023all,valanarasu2022transweather}.
For instance, AirNet~\cite{li2022all} introduces a degradation classifier trained via contrastive learning to guide restoration, while ADMS~\cite{park2023all} employs a multi-type degradation classifier to dynamically select Adaptive Discriminant filters, enabling degradation-specific parameter modulation within the restoration network.

\subsection{LLM Agents}

Advancements in LLM-based frameworks have enabled more structured reasoning and agent designs, particularly for complex multimodal tasks. Initial efforts emphasized improving reasoning capabilities through refined prompting strategies and modular architecture. Chain-of-Thought (CoT) prompting~\cite{wei2022chain} introduced stepwise reasoning, facilitating decomposition and interpretability across diverse tasks. ReAct~\cite{yao2023react} combined reasoning with tool interaction by interleaving thought traces and external actions, supporting more adaptive behavior. Extending this direction, CoALA~\cite{sumers2023cognitive} formalized components such as memory, reasoning, and control within a cognitive architecture, offering a modular design space for building general-purpose language agents. These developments established a basis for domain-specific agent systems with integrated reasoning pipelines.
Building on these foundations, application-driven LLM agents have been developed to incorporate tool use and dynamic decision-making within specialized domains. In vision tasks, MMCTAgent~\cite{kumar2024mmctagent}, VideoAgent~\cite{wang2024videoagent}, and ReAgent-V~\cite{zhou2025reagent} implement planning and evaluation pipelines for image and video analysis, incorporating external modules for retrieval and verification.
In the medical domain, agents such as MedCoT~\cite{liu2024medcot}, CLINICR~\cite{nachane2024few}, and MMedAgent-RL~\cite{xia2025mmedagent} employ hierarchical reasoning frameworks to address clinical questions, integrating structured logic and domain-specific knowledge to enhance interpretability and decision quality. 

Similarly, LLM-based agents have also emerged as a promising paradigm for tackling complex image restoration tasks involving multiple degradations. RestoreAgent~\cite{chen2024restoreagent} pioneered the use of MLLMs for autonomous task identification, model selection, and execution planning. AgenticIR~\cite{zhu2024intelligent} introduced a five-stage human-inspired workflow—Perception, Scheduling, Execution, Reflection, and Rescheduling—augmented with self-exploration to build IR-specific experience. MAIR~\cite{jiang2025multi} advanced this by employing a multi-agent system guided by real-world degradation priors, improving both efficiency and scalability. HybridAgent~\cite{li2025hybrid} proposed a hybrid interaction scheme with fast and slow agents, along with a mixed-distortion removal strategy to mitigate error propagation. Q-Agent~\cite{zhou2025q} further introduces a quality-driven chain-of-thought framework, leveraging no-reference IQA metrics to guide greedy restoration without costly rollbacks. These works demonstrate the growing potential of combining general-purpose language intelligence with visual tools for robust, adaptive image restoration.
More recently, JarvisIR~\cite{lin2025jarvisir} and JarvisArt~\cite{lin2025jarvisart} leveraged intelligent agent workflows to perform task-oriented image restoration and creative photo retouching.

\section{Concluding Remarks}

In this paper, we introduced 4KAgent, a versatile agentic image super-resolution generalist model designed to universally upscale images of diverse types and degradation levels to 4K resolution. By leveraging advanced multi-expert integration, adaptive decision-making, and specialized tools for perception and fidelity optimization, 4KAgent significantly enhances restoration quality across various challenging domains, including severely degraded images, natural scenes, portraits, AI-generated content, and specialized scientific modalities such as remote sensing, microscopy, and medical imaging.
Our extensive evaluations on standard benchmarks and specialized datasets confirm that 4KAgent consistently outperforms existing state-of-the-art approaches, especially in complex scenarios where conventional super-resolution methods fall short. This robust performance, achieved without domain-specific retraining, demonstrates the model’s unique generalizability and practical utility for generic deployment in both consumer, commercial, and scientific applications.

\paragraph{Future Work}
Looking ahead, we have identified several promising directions that can further enhance the capabilities and applicability of 4KAgent, enabling broader use cases.
First, we will optimize the efficiency of 4KAgent by designing more accurate distortion-perception models and refining the execution-reflection-rollback mechanism to achieve faster and higher-quality image restoration. Second, we will prioritize enhancing the safety and robustness of 4KAgent to mitigate risks such as privacy breaches and the generation of harmful imagery. Lastly, we will continuously expand the toolbox of 4KAgent by integrating additional domain-specific restoration methods and developing targeted restoration profiles, thus significantly improving performance and user experience across specialized imaging applications.

\clearpage

\bibliographystyle{plain}  %plainnat,abbrvnat,unsrtnat
\small
\bibliography{Reference} 
\normalsize

%%%%%%%%%%%%%%%%%%%%%%%%%%%%%%%%%%%%%%%%%%%%%%%%%%%%%%%%%%%%

% \newpage
% \appendix
% \input{Chapters/Appendix}

%%%%%%%%%%%%%%%%%%%%%%%%%%%%%%%%%%%%%%%%%%%%%%%%%%%%%%%%%%%%

\end{document}